\documentclass[twoside,11pt]{article}

\usepackage[preprint]{jmlr2e}

\usepackage{amsfonts}
\usepackage{amsmath}
\usepackage{tikz}
\usetikzlibrary{calc}
\usepackage{pgfplots}
\usepgfplotslibrary{patchplots}
\usetikzlibrary{patterns, positioning, arrows, chains, shapes.geometric, fit, shapes, arrows.meta, calc, backgrounds}
\pgfplotsset{compat=1.15}
\usepackage{dirtytalk}
\usepackage{subcaption}
\usepackage{tabularx}
\usepackage{booktabs}
\usepackage[labelfont=bf,format=plain,justification=raggedright,singlelinecheck=false]{caption}
\usepackage{hyperref}
\usepackage{url}
\usepackage{ntheorem}
\usepackage[capitalise]{cleveref}

\makeatletter
\def\cleartheorem#1{\expandafter\let\csname#1\endcsname\relax
    \expandafter\let\csname c@#1\endcsname\relax
}
\makeatother
\cleartheorem{example}
\cleartheorem{theorem}
\cleartheorem{lemma}
\cleartheorem{proposition}
\cleartheorem{remark}
\cleartheorem{corollary}
\cleartheorem{definition}
\cleartheorem{conjecture}
\cleartheorem{axiom}

\newtheorem{theorem}{Theorem}
\newtheorem{lemma}[theorem]{Lemma} 
\newtheorem{proposition}[theorem]{Proposition} 
\newtheorem{remark}[theorem]{Remark}

\newtheorem{definition}[theorem]{Definition}

\pgfmathdeclarefunction{gauss}{3}{%
  \pgfmathparse{1/(#3*sqrt(2*pi))*exp(-((#1-#2)^2)/(2*#3^2))}%
}
\def\R{\mathbb R }
\def\N{\mathbb N}
\def\P{\mathbb P}
\def\E{\mathbb E}
\def\L{\mathbb L}
\def\Q{\mathcal Q}
\def\F{\mathcal F}
\def\I{\mathcal I}

\def\K{\mathcal K}
\def\X{\mathcal X}
\def\Y{\mathcal Y}
\def\Z{\mathcal Z}
\def\V{\mathcal V}
\def\U{\mathcal U}
\def\B{\mathcal B}
\def\M{\mathcal M}
\def\A{\mathcal A}
\def\C{\mathcal C}
\def\G{\mathcal G}
\def\T{\mathcal T}
\def\H{\mathcal H}
\def\S{\mathcal S}
\def\CGM{\mathfrak{C}}
\def\SCM{\mathfrak{M}}
\def\CR{\mathfrak{A}}
\def\Norm{{\mathcal{N}(0,1)}}
\newcommand{\Prob}[1]{\mathcal{P}(#1)}
\newcommand{\Unif}[1]{\operatorname{Unif}(#1)}
\newcommand{\Joint}[3]{\Q_{\text{joint}}\left(#1,#2,#3 \right)}
\newcommand{\Jointc}[3]{\Q^*_{\text{joint}}\left(#1,#2,#3 \right)}
\newcommand{\law}[1]{\mathbb{L}\left(#1\right)}

\newcommand{\Norma}[1]{\Q_{\Norm}\left(#1\right)}
\newcommand{\Normac}[1]{\Q^*_{\Norm}\left(#1\right)}
\newcommand{\Lap}{\operatorname{Lap}}

\def\Id{\operatorname{Id}}
\def\MMD{\operatorname{MMD}}
\def\OT{\operatorname{OT}}
\def\Loss{\operatorname{L}}
\newcommand{\rmd}{\mathrm{d}}
\newcommand{\ttx}{\mathtt{x}}
\newcommand{\tty}{\mathtt{y}}

\newcommand{\abs}[1]{{\left\lvert #1 \right\rvert}}
\newcommand{\norm}[1]{{\left\lVert #1 \right\rVert}}

\newcommand{\pa}{\operatorname{pa}}
\newcommand{\desc}{\operatorname{desc}}
\newcommand{\ndesc}{\overline{\operatorname{desc}}}
\newcommand{\an}{\operatorname{an}}
\newcommand{\nan}{\overline{\operatorname{an}}}
\newcommand{\doint}{\operatorname{do}}

\newcommand{\Obs}[1]{\operatorname{Obs}_{\mathfrak{#1}}}
\newcommand{\Int}[1]{\operatorname{Int}_{\mathfrak{#1}}}
\newcommand{\Ctf}[1]{\operatorname{Ctf}_{\mathfrak{#1}}}

\newcommand\independent{\protect\mathpalette{\protect\independenT}{\perp}}
\def\independenT#1#2{\mathrel{\rlap{$#1#2$}\mkern2mu{#1#2}}}

\usepackage{lastpage}
\jmlrheading{00}{2025}{1-\pageref{LastPage}}{X/XX; Revised X/XX}{X/XX}{21-0000}{Lucas De Lara}

\ShortHeadings{Canonical Representations of Markovian SCMs}{De Lara}
\firstpageno{1}

\begin{document}

\title{Canonical Representations of Markovian Structural Causal Models: A Framework for Counterfactual Reasoning}

\author{\name Lucas De Lara \email lucas.de-lara@univ-lorraine.fr \\
       \addr Institut \'Elie Cartan de Lorraine\\
       Université de Lorraine\\
       Nancy, France}

\editor{}

\maketitle

\begin{abstract}
Counterfactual reasoning aims at answering contrary-to-fact questions like “Would have Alice recovered had she taken aspirin?” and corresponds to the most fine-grained layer of causation. Critically, while many counterfactual statements cannot be falsified—even by randomized experiments—they underpin fundamental concepts like individual-wise fairness. Therefore, providing models to formalize and implement counterfactual beliefs remains a fundamental scientific problem. In the Markovian setting of Pearl’s causal framework, we propose an alternative approach to structural causal models to represent counterfactuals compatible with a given causal graphical model. More precisely, we introduce counterfactual models, also called canonical representations of structural causal models. They enable analysts to choose a counterfactual assumption via random-process probability distributions with preassigned marginals and characterize the counterfactual equivalence class of structural causal models. Using these representations, we present a normalization procedure to disentangle the (arbitrary and unfalsifiable) counterfactual choice from the (typically testable) interventional constraints. In contrast to structural causal models, this allows to implement many counterfactual assumptions while preserving interventional knowledge, and does not require any estimation step at the individual-counterfactual layer: only to make a choice. Finally, we illustrate the specific role of counterfactuals in causality and the benefits of our approach on theoretical and numerical examples.
\end{abstract}


\section{Introduction}

\emph{Pearl's causality ladder} distinguishes three levels of queries of increasing strength that causal reasoning seeks to answer: (1) \emph{observational}, (2) \emph{interventional}, and (3) \emph{counterfactual} \citep{pearl2018book}. For illustration, consider a medical context where an analyst aims to understand the link between taking aspirin and recovering from headache. The first level focuses on predictions from observations, from \emph{seeing}. It allows to answer questions such as \say{What is the recovery rate among people taking aspirin?}. The second level addresses predictions from actions, from \emph{doing}. It enables one to answer questions like \say{What percentage of patients would recover if we give them aspirin?}. This corresponds to the result of a randomized controlled trial, where (in contrast to the observational rung) patients do not freely choose whether they take aspirin: an agent blindly assigns the treatment status. The third level tackles predictions from contrary-to-fact events, from \emph{imagining}. It notably permits one to answer questions of the form \say{Had Alice taken aspirin (assuming she did not), would have she recovered?}.

\emph{Causal inference} refers to the task of answering queries from the second or third levels, using the lower levels along with additional assumptions, typically encoded as mathematical models. The interventional and counterfactual rungs differ fundamentally at two regards in causal inference: the types of conclusions they allow to make and the possible inference methods to reach these conclusions. First, interventional questions address only \emph{general causes} (does the treatment work?) whereas many counterfactual questions generally deal with \emph{singular causes}\footnote{Singular causation is also often referred to as \emph{individual causation} or \emph{actual causation}.} (does the treatment work for Alice?) \citep[Chapter 7]{pearl2009causality}. The treatment could perfectly work in average in the whole population while degrading Alice's health. Second, while various interventional statements can be tested through fully randomized experiments, most counterfactual statements cannot be empirically verified. Verification of the example statement would require to also observe Alice's outcome in the alternative reality where she took the treatment all other things being kept equal.

The fact that some counterfactual statements cannot be falsified notably led \cite{dawid2000causal} to qualify them as \say{metaphysical} and to advocate restricting causal analysis to the interventional rung. Nevertheless, counterfactuals have always played a crucial role in common language (to express our beliefs on causation) and in scientific modeling (precisely to ask metaphysical questions) \citep{pearl2000causal}. Furthermore, as reminded by \cite{nasr2023counterfactual}, they serve to define fundamental concepts like harm \citep{richens2022counterfactual,mueller2023personalized,sarvet2025perspectives}, credit \citep{mesnard2021counterfactual}, and fairness \citep{kusner2017counterfactual}, at the basis of critical applications notably in justice (see also \citep[Section 4.4]{pearl2016causal}). In particular, by allowing to articulate \emph{singular} causes, counterfactuals can define notions of algorithmic fairness at the \emph{individual} level \citep{kusner2017counterfactual}: the only legally-grounded level in the French law. Therefore, despite their unfalsifiability, designing intelligible models to formally represent counterfactual conceptions remains an essential scientific problem with practical consequences. Addressing this problem is precisely the goal of this article.

Counterfactual assumptions that cannot be tested rest on an arbitrary choice. Supposing that had Alice received aspirin her level of pain would have been given, for instance, by rank preservation between the control and treated groups can be nothing more than choice. This is why analysts working at the level of singular causation need a mathematical framework to formalize such choices on top of given assumptions describing general causation. \emph{Structural causal models} \citep{pearl2009causality}, by fully describing the latent rules governing the data-generating process, enable one to answer queries from the whole causality ladder \citep{pearl2018book,bareinboim2022pearl}. However, we argue that structural causal models in their classical form---based on structural equations and exogenous distributions over a directed graph---raise some issues to sensibly encode the counterfactual level, and more specifically singular causes. Notably, the structural representation can be hard to unpack into informal, intelligible counterfactual knowledge. Conversely, it may feel unclear how to translate counterfactual beliefs into the structural formalism. Moreover, the structural framework is inconvenient to modify counterfactual assumptions without changing the observational and interventional levels.

To solve this problem, we propose a class of models that are equivalent to structural causal models, in the sense that they characterize the same causality ladders, while properly separating layers of causation (notably general and singular causes). Crucially, these models do not rely on standard structural equations to represent counterfactual conceptions. We call them \emph{counterfactual models} or \emph{canonical representations of structural causal models}. They build upon the fact that a singular counterfactual quantity in a structural causal model is mathematically derived from a joint probability distribution between marginals representing general causes. Then, we introduce \emph{normalizations of counterfactual models}, which allow to specify the intrinsic cross-dependencies of counterfactual joint probability distributions independently of the marginals. This approach enables analysts to transparently stipulate their singular-level assumptions without altering the general-level ones.  

This work has theoretical and practical interests. It provides a natural framework to derive a counterfactual conception from a causal model and reciprocally to integrate a counterfactual conception into a causal model. Furthermore, it enables analysts aiming to learn a causal model to disentangle the objective of fitting the given general causes (encoded by a graphical model) from the choice of the singular causes. Note that our approach builds upon a \emph{given} causal graphical model; it does not address the estimation or design of such a model. In an introductory example below, we illustrate the limitations of structural causal models and the principles of the proposed solution. We emphasize that for general purposes, counterfactual models are not necessarily better than structural causal models. They simply offer an alternative---yet equivalent---perspective to counterfactual reasoning, that has several advantages.

Overall, this article aims at enriching the understanding and modeling of counterfactuals in causality. It doing so, we expect to clarify fundamental differences within the causality research, based on the concerned layer of causation and the way causal models are employed. Further, we hope to bridge the gap between conceptual counterfactual notions and their practicality, by providing a clearer and more convenient framework than structural causal models to test, discuss, and implement any counterfactuals compatible with a same causal graphical model. Developing this approach also led us to study random processes and distributional regression in theory and in practice.

\subsection{Outline of the paper}

After an introductory example illustrating the challenges of reasoning counterfactually with structural causal models and the functioning of counterfactual models (\cref{sec:intro_example}), the rest of the paper is organized as follows:
\begin{itemize}
    \item \cref{sec:notation} presents the \emph{basic notation} and essential background on probability;
    \item \cref{sec:setup} introduces \emph{Pearl's causal framework}, notably causal graphical models and structural causal models;
    \item \cref{sec:ctf_models} is the main section: it defines \emph{counterfactual models}, proves their equivalence to structural causal models, and explains how to specify them in practice via normalizations;
    \item \cref{sec:regression} studies a generic \emph{distributional regression} problem with monotonicity constraints and details how it can notably be applied to implement normalizations of counterfactual models;
    \item \cref{sec:exp} contains \emph{numerical experiments};
    \item \cref{sec:comparison} discusses the similarities and differences of our approach to \emph{related works}.
\end{itemize}
We defer the proofs of the theoretical results to \cref{sec:proofs}. \Cref{sec:monotonic_networks} and \cref{sec:quantile_regression} propose extra background and experiments.  

\subsection{Motivating example}\label{sec:intro_example}

The goal of this example is three-fold. First, to illustrate the specific place of counterfactual reasoning, notably singular causation, in causal analysis. Second, to highlight issues that arise when modeling counterfactual assumptions through structural causal models. Third, to introduce an alternative way of intelligibly encoding counterfactual assumptions. This presents all the basics ideas of the paper. For simplicity, we use on-the-fly mathematical notation and informal definitions of causal models throughout this example. We refer to \cref{sec:notation} and \cref{sec:setup} for a complete specification.

\subsubsection{Illustrating the causal hierarchy}

Let us illustrate the causal hierarchy on a concrete toy example. We consider a fictitious medical study where the variables \texttt{t} and \texttt{y} respectively represent a \emph{medical dose} and a \emph{health outcome}. The possible observations follow the probability distribution $P_{\texttt{t},\texttt{y}}$ on $\R^2$ given by $P_{\texttt{t}} := \operatorname{Unif}([0,10])$ and $P_{\texttt{y} \mid \texttt{t}} (\bullet | t) := \mathcal{N}(m(t),1)$ where $m(t):= 10 \sin(\frac{\pi t}{14})$ for $t \in [0,10]$. The knowledge of $P_{\texttt{t},\texttt{y}}$ alone corresponds to the observational level: by inferring features of $P_{\texttt{t},\texttt{y}}$, an analyst can estimate statistical associations between \texttt{t} and \texttt{y} but not necessarily causal dependencies.

\begin{figure}[tb]
    \centering
    \begin{subfigure}[b]{0.47\textwidth}
        \centering
        \includegraphics[width=\linewidth]{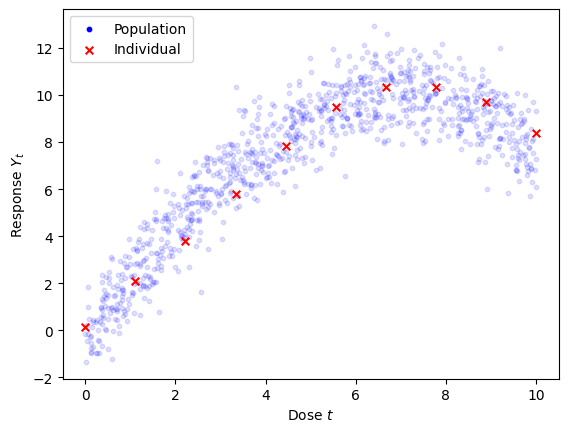}
        \caption{Population response / Individual response}
        \label{fig:pop_indiv}
     \end{subfigure}
     \hfill
     \begin{subfigure}[b]{0.47\textwidth}
        \includegraphics[width=\linewidth]{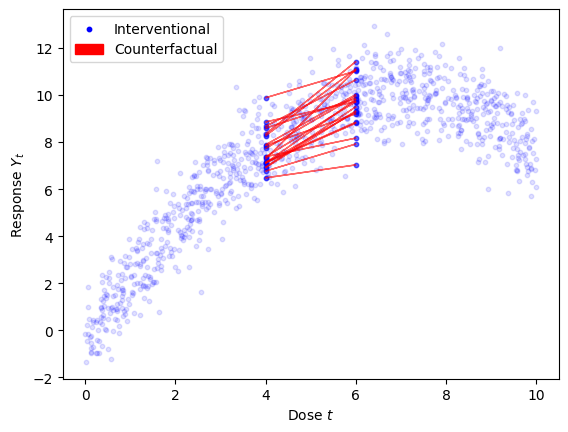}
        \caption{Interventional laws / Counterfactual law}
        \label{fig:int_cf}
     \end{subfigure}
    \caption{Differences between interventional and counterfactual assumptions on a dose-response curve from the example randomized controlled trial. The \textcolor{blue}{blue} highlights observed experimental results (identifiable in $\C$), whereas the \textcolor{red}{red} highlights untestable hypotheses (identifiable in $\M$). \Cref{fig:pop_indiv} tracks a \emph{same} patient across alternative realities where they received different values of the treatment. \Cref{fig:int_cf} represents the counterfactual coupling between the marginals for $t=4$ and $t=6$. Counterfactuals where generated via the covariance function $k(t,t') := \exp(\abs{t-t'}^2/(2 \sigma^2))$ with $\sigma = 2$.}
    \label{fig:counterfactual_dose}
\end{figure}

To address the interventional level, we assume that \texttt{t} is the treatment from a randomized controlled trial. This ensures that the variables are causally ordered according to the simple graph $\texttt{t} \rightarrow \texttt{y}$ denoted by $\G$, which implies that the conditional dependence of \texttt{y} in \texttt{t} represents causation. As such,  $P_{\texttt{y} \mid \texttt{t}}$ has a causal interpretation. The fact that $P_{\texttt{y} \mid \texttt{t}} (\bullet | t) = \mathcal{N}(m(t),1)$ with $m$ minimal at $0$ on $[0,10]$ signifies that the treatment works in the sense that it increases health in average. Note, however, that its efficiency decreases for $t > 7$. The knowledge of $P_{\texttt{t},\texttt{y}}$ and $\G$ forms a \emph{causal graphical model} $\C$. It completes the observational level with graphical assumptions to allow causal claims. \Cref{sec:cgm} furnishes a reminder on causal graphical models. Critically, such a model fully handles the interventional level, not the complete counterfactual level. For illustration, suppose that Alice received $t=4$ and experienced $y=5$. With $\C$ alone, an analyst can conclude that increasing the dose from $t=4$ to $t=6$ improves health by $m(6)-m(4)$ units in average for the general population, but cannot determine what would have been the outcome of Alice specifically had $t=6$. We refer to \cref{fig:pop_indiv} for an illustration. \cite{balakrishnan2025conservative} proposed similar graphics to highlight the nonidentifiability of counterfactual curves.

To tackle the whole counterfactual level, including individual causes, one needs stronger hypotheses than randomization or the knowledge of a causal graphical model. Encoding counterfactual assumptions on top of $\C$ can be achieved by postulating a structural causal model $\M$ compatible with $\C$. \Cref{sec:scm} details structural causal models. In this work, we focus on \emph{Markovian} models only. Such a model basically corresponds to a pair of independent exogenous random variables $(U_\texttt{t},U_\texttt{y})$ and a pair of measurable functions $(f_\texttt{t},f_\texttt{y})$ such that the endogenous random variables $(T,Y)$ defined by the assignments
\begin{align*}
    &T := f_\texttt{t}(U_\texttt{t}),\\
    &Y := f_\texttt{y}(T,U_\texttt{y}),
\end{align*}
meet $(T,Y) \sim P_{\texttt{t},\texttt{y}}$. The functional dependence of $Y$ in $T$ conforms to the graph $\G$. These assignments enables one to carry out interventions producing counterfactual variables. Concretely, intervening on $\texttt{t}$ defines the \emph{potential outcome} $Y_t := f_\texttt{y}(t,U_\texttt{y})$ representing the outcome had the treatment been equal to $t$ for $t \in [0,10]$. In $\M$, one can identify the marginal laws of $(Y_t)_{t \in [0,10]}$, called \emph{interventional laws}. More precisely, note that $U_\texttt{t} \independent U_\texttt{y}$ implies $T \independent U_\texttt{y}$, and thereby $Y_t \sim P_{\texttt{y} \mid \texttt{t}} (\bullet | t)$. As such, $\M$ contains all the assumptions encoded in $\C$, that is the observational and interventional rungs. Furthermore, because the variables $(Y_t)_{t \in [0,10]}$ share a common source of randomness $U_\texttt{y}$, they follow a joint probability distribution between the interventional marginals called a \emph{counterfactual law}. Such a law cannot be identified in $\C$ only. \Cref{fig:int_cf} illustrates the distinction between interventional and counterfactual laws, by representing a counterfactual coupling over $(Y_4,Y_6)$.

To summarize, one can formulate counterfactual assumptions (third rung) respecting observational and interventional knowledge (first and second rung) by postulating a structural causal model $\M$ compatible with the known graphical model $\C$. This raises several questions regarding the link between a structural causal model and the produced counterfactual distributions. How to modify a structural causal model as to change the counterfactual conception without modifying the observational and interventional assumptions? Are all couplings between two interventional marginals attributable to a structural causal model? How to unpack the counterfactual assumptions contained in a structural causal model? Conversely, how to design a structural causal models compatible with given counterfactual assumptions? This work precisely aims to address these questions.

\subsubsection{Challenges of structural counterfactual modeling}

To illustrate the challenges underlying these questions, we first consider the following assignments:
\begin{align*}
    &T := U_\texttt{t},\\
    &Y^+ := m(T) + U_\texttt{y},
\end{align*}
where $U_\texttt{t} \sim P_\texttt{t}$ and $U_\texttt{y} \sim \mathcal{N}(0,1)$. It characterizes a structural causal model $\M_{+}$ compatible with the graphical model $\C$ previously introduced: the observational and interventional laws remain the same as before. Let us unpack the counterfactual conception that this model represents. It entails the counterfactual coupling $(Y^+_4,Y^+_6) := (m(4)+U_\texttt{y}, m(6)+U_\texttt{y})$, which is the deterministic coupling between the interventional marginals characterized by the translation $y \mapsto m(6)-m(4) + y$ from $P_{\texttt{y} \mid \texttt{t}} (\bullet | 4)$ to $P_{\texttt{y} \mid \texttt{t}} (\bullet | 6)$. In terms of coupling procedure, it simply matches the quantiles of the marginals: counterfactuals are obtained by rank preservation between the interventional laws.

Consider now an analyst who would like to change this conception of counterfactuals while keeping the observational and interventional assumptions. They face two difficulties. Firstly, modifying the structural assignments without care also affects $\C$. Secondly, it is often unclear how a modification of the assignments modifies the counterfactuals. For instance, changing the coefficient 1 before $m(t)$ by $\alpha \neq 1$ does keep the same graph but not the same observational and interventions distributions: the causal kernel becomes $P^\alpha_{\texttt{y}|\texttt{t}}(\bullet|t) = \mathcal{N}(\alpha m(t),1)$. Regarding the change for the counterfactual transition from $t=4$ to $t=6$, the coupling switches from $(m(4)+U_\texttt{y}, m(6)+U_\texttt{y})$ to $(\alpha m(4)+U_\texttt{y}, \alpha m(6)+U_\texttt{y})$. While the associated distributions are mathematically unequal, they differ only by their marginals---not by how pairs of instances are coupled. They both consist in matching the quantiles of each marginal. All in all, altering the parameters of the structural assignments to modify specifically the counterfactuals failed at many regards. It preserved the graph, but changed the observational and interventional levels and not really the counterfactual level. Note that changing the law of $U_{\texttt{y}}$ (like its mean) while keeping the rest of the mechanism equal yields the same issues.

To show what sorts of model modifications change the counterfactuals, suppose that the analyst aims for the following counterfactual assumption: the counterfactual coupling matches the quantiles of $P_{\texttt{y} \mid \texttt{t}} (\bullet | 4)$ and $P_{\texttt{y} \mid \texttt{t}} (\bullet | 6)$ in opposition. The model $\M_{-}$ below reaches this conception while preserving $\C$:
\begin{align*}
    &T := U_\texttt{t},\\
    &Y^- := m(T) + \operatorname{sign}(5-T) U_\texttt{y}.
\end{align*}
To see it, simply note that $(Y^-_4,Y^-_6) = (m(4) + U_\texttt{y}, m(6) - U_\texttt{y}) = (Y^-_4, m(6)+m(4)-Y^-_4)$, meaning that $Y^-_6$ is a decreasing function of $Y^-_4$. Notably, this modification required to alter the additive role of the original noise. 
More generally, despite their popularity and seemingly generic form, additive noise model fully determine the counterfactuals knowledge, as counterfactual counterparts are obtained via rank preservation. Dealing with more various counterfactual conceptions requires thinking of other model architectures. It poses the problem of finding the set of modifications of structural causal models reaching the desired conception of counterfactuality, or finding an alternative class of models encoding counterfactual assumptions in a more convenient way.

\subsubsection{Navigating through the diversity of counterfactual conceptions}

Moving on, we aim at understanding the class of counterfactual assumptions generated by structural causal models and how to navigate across them. In the context of our medical example, a counterfactual conception corresponds to a joint probability distribution between the interventional marginals $\left( P_{\texttt{y}|\texttt{t}}(\bullet|t) \right)_{t \in [0,10]}$. Let us show how to represent diverse counterfactual conceptions compatible with the same graphical model $\C$ via structural causal models. 

\begin{figure}[tb]
    \centering
    \includegraphics[width=0.7\linewidth]{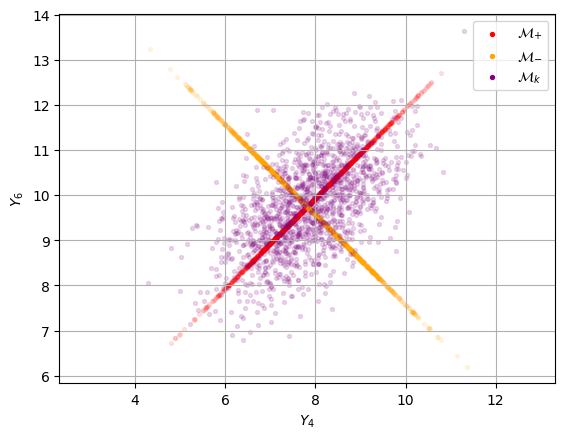}
    \caption{Representation of 1500 samples from the distribution of $(Y_4,Y_6)$ in $\M_+$ (in \textcolor{red}{red}), $\M_-$ (in \textcolor{orange}{orange}) and $\M_k$ (in \textcolor{purple}{purple}) to illustrate the indeterminacy of counterfactuals. Different behaviors are compatible with $\C$. Counterfactuals can be deterministic (\textcolor{red}{red} and \textcolor{orange}{orange}) or stochastic (\textcolor{purple}{purple}); comonotonic (\textcolor{red}{red} and \textcolor{purple}{purple}) or countermonotonic (\textcolor{orange}{orange}). The \textcolor{orange}{orange} distribution describes a situation where even though increasing the dose improves health in average it can degrades the health of some individuals.}
    \label{fig:noneq_cf}
\end{figure}

Let $U_\texttt{t} \sim P_\texttt{t}$ and $U^k_\texttt{y} := (U^k_{\texttt{y},t})_{t \in [0,10]}$ be a Gaussian process with mean function $m(t)$ and covariance function $k(t,t') := \exp(\abs{t-t'}^2/(2 \sigma^2))$ (called a \emph{Gaussian kernel}) with $\sigma = 2$, such that $U_\texttt{t} \independent U_\texttt{y}$. Then, define the structural causal model $\M_k$ given by
\begin{align*}
    &T := f_\texttt{t}(U_\texttt{t}) := U_\texttt{t},\\
    &Y^k := f_\texttt{y}(T,U^k_\texttt{y}) := U^k_{\texttt{y},T}.
\end{align*}
Here, $f_\texttt{y}(t',(u_t)_{t \in [0,10]})$ returns the projection of $(u_t)_{t \in [0,10]} \in \R^{10}$ at index $t' \in [0,1]$.\footnote{This trick is similar to how scholars show that structural causal models and potential-outcome models are equivalent. In particular, $(U^k_{\tty,t})_{t \in [0,10]}$ can be seen as a collection of potential outcomes. We refer to \cref{sec:po} for more details.} Like $\M_{+}$ and $\M_{-}$, the model $\M_k$ is compatible with $\C$. In particular, $Y^k_t := U^k_{\texttt{y},t} \sim \mathcal{N}(m(t),1) = P_{\texttt{y}|\texttt{t}}(\bullet|t)$ by construction. However, their counterfactual distributions differ. We notably have:
\[
    (Y^k_4,Y^k_6) \sim \mathcal{N}\left(\begin{bmatrix} m(4) \\ m(6) \end{bmatrix}, \begin{bmatrix} 1 & k(4,6) \\ k(4,6) & 1\end{bmatrix}\right).
\]
\Cref{fig:noneq_cf} illustrates how the three structural causal models encode very different singular causes compatible with the same general causes. Note that by changing the covariance function $k$, one can generate even more distinct counterfactual distributions compatible with the same observational and interventional distributions.

More generally, one can choose any distribution for $U^k_\texttt{y}$ as long as it meets two conditions: (1) $(t,\omega) \mapsto U^k_{\tty,t}(\omega)$ is measurable; (2) the marginal laws remain the same. Condition (1) ensures that $U^k_{\tty,T}$ is a well-defined random variable. While this technical point is fundamental for the sake of rigor and will be discussed in the paper (see \cref{sec:joint} and \cref{rem:measurability}), we emphasize that it is not necessary to understand the principle of our approach. In the specific case of the current example, measurability follows from the measurability of $m$ and the continuity of $k$. Condition (2) preserves the observational and intervention knowledge when changing the counterfactual assumption. As demonstrated later, this condition is not enough in the case of more general causal graphs $\G$, as a coherent counterfactual joint probability distribution must \emph{also} verify a factorization property relative to the graph.


This example solves the previous issues: it manages to readily translate counterfactual conceptions into structural causal models and to change such conceptions without touching the causal graphical model encoding the observational and interventional knowledge. The key ingredient was writing the assignment of $Y$ as random process $U_{\texttt{y},T}$ rather than as a functional relation $f_{\texttt{y}}(T,U_{\texttt{y}})$ with a single noise. In the literature, pedagogical examples of structural causal models that entail the same interventional distributions but distinct counterfactual distributions typically follow the same principle (like \citep[Section 3.4]{peters2017elements} and \citep[Example D.6]{bongers2021foundations}). This trends upholds the idea that this approach is cognitively natural to formalize counterfactual conceptions. \Cref{sec:comp_to_scm} details the advantages of this process viewpoint compared to the functional account.

Note that this strategy can be regarded as placing a probability distribution directly on the deterministic function that maps a cause to its effect rather than on an abstract exogenous variable. We recall that a random process induces a probability distribution on a function space: in the above example, a realization of $U^k_{\tty}$ is a deterministic function from $t \in [0,1]$ to $y \in \R$. \cite{balke1994counterfactual} originally proposed this type of reparameterization of structural causal models to bound counterfactual probabilities when only the causal graphical model is known. \cite{peters2017elements} also applied such a reformulation for pedagogical reasons and called it \emph{canonical representation}. To our knowledge, this representation has only been used as a theoretical construction to illustrate properties of structural causal models (causal hierarchy, invariance by exogenous reparameterizations) or as a technique to bound counterfactual queries, rather than as a general operational principle to formalize and implement counterfactuals compatible with a given causal graphical model. The objective of this paper is to address this point.

\subsubsection{Normalizing a canonical representation}

The above example of canonical representations has a practically convenient feature: since the interventional marginals are Gaussian, it suffices to make $U_{\tty}$ run in the set of Gaussian processes with mean function $m$ to generate various (but not all) counterfactual conceptions. To generalize this strategy to possibly non-Gaussian marginals, we propose a \emph{normalization} procedure inspired by copula theory. The first step consists in computing, for every $t \in [0,10]$, the unique non-decreasing transport map from $\Norm$ to $P_{\tty|\mathtt{t}}(\bullet|t)$, denoted by $\psi_{\tty}(\bullet|t)$. Note that the transformation $\psi_{\tty}$ is fully identified by the causal graphical model; it does not contain any counterfactual information. The second step amounts to specifying a centered Gaussian process (or any stochastic process with normal marginals) $E^{(\tty)} := (E^{(\tty)}_t)_{t \in [0,10]}$, which characterizes the counterfactual conception in a normalized space. The third step comes to transforming the normalized process into a counterfactual joint probability distribution with the preassigned interventional marginals through
\[
    Y_t := \psi_{\tty}(E^{(\tty)}_t|t) \sim P_{\tty|\mathtt{t}}(\bullet|t),
\]
for every $t \in [0,10]$. Next, one sets $T \sim P_{\mathtt{t}}$ and $Y := Y_T = \psi_{\tty}(E^{(\tty)}_T|T)$ for consistency.

Note that $E^{(\tty)}_T \sim \Norm$: the random process $E^{(\tty)}$ does not capture any marginal information; only cross-world information. The cross-world dependencies in the normalized space are preserved by the monotonicity of $\psi_{\tty}$. Notably, if the variables $(E^{(\tty)}_t)_{t \in [0,10]}$ are all equal, then $(Y_t)_{t \in [0,10]}$ are comonotonic.

This approach has several advantages compared to parameterizing counterfactuals via structural causal models. In particular, it disentangles the \emph{constraint} of fitting the interventional marginals from the \emph{choice} of the counterfactual conception. The idea is that a practitioner can choose intelligible cross-world dependencies by selecting a Gaussian process while the transform $\psi_{\tty}$ automatically maps these dependencies to the given marginals. As thoroughly discussed in \cref{sec:comp_to_scm}, this simplifies both modeling and numerical aspects.

Overall, the key contributions of this article can be summarized as follows. In \cref{sec:def_ctf}, we properly extend the construction of the canonical representation to arbitrary graphs, and then show the equivalence between structural causal models and canonical representations. In \cref{sec:normalizations}, we detail the normalization procedure to define counterfactual joint probability distributions with arbitrary marginals. In \cref{sec:regression}, we introduce a distributional regression framework to learn the transform $\psi_{\tty}$ at the core of the normalization strategy. In \cref{sec:exp}, we illustrate the method numerically.

\section{Notation and definitions}\label{sec:notation}

This section specifies the basic definitions and notation that will be used throughout the paper. We refer to \Cref{table:notation} for a summary, which anticipates on causal notation.

Let $d \geq 1$ be an integer and define $[d] := \{1,\ldots,d\}$. By convention, $[d] = \emptyset$ if $d \leq 0$. When $\V := \times^d_{i=1} \V_i$ is a Cartesian product indexed by $[d]$, we define for every $I \subseteq [d]$ the Cartesian product $\V_I := \times_{i \in I} \V_i$. Similarly, for any vector $v := (v_i)_{i \in [d]} \in \R^d$ and any subset $I \subseteq [d]$ we write $v_I := (v_i)_{i \in I}$. Whatever the space, we denote by $\Id$ the identity function.

In this article, an $i \in [d]$ typically refers to a variable with domain $\V_i$. In some illustrations (as in \cref{sec:intro_example}), we employ letters $(\texttt{x},\texttt{t}, \texttt{y},\ldots)$ rather than numbers $(1,2,3,\ldots)$ to denote variables for the sake of clarity. Moreover, we make distinctions between \emph{variables} $(\texttt{x},\texttt{t}, \texttt{y},\ldots)$, \emph{random variables} $(X,T,Y,\ldots)$, and \emph{values} $(x,t,y,\ldots)$.

\subsection{Probability measures and kernels}

\paragraph{Measures.} For any Polish space $\Xi$, we write $\B(\Xi)$ for the Borel $\sigma$-algebra of $\Xi$ and $\Prob{\Xi}$ for the set of Borel probability measures on $\Xi$. By measurable we always mean Borel measurable. For any $m \in \R^d$ and any positive semidefinite matrix $C \in \R^{d \times d}$, we denote by $\mathcal{N}(m,C)$ the (multivariate) Gaussian distribution with mean $m$ and covariance matrix $C$, which belongs to $\Prob{\R^d}$. In particular, we refer to $\Norm$ as the \emph{normal distribution}. Additionally, $\Lap(m,\lambda)$ is the Laplace distribution with mean $m \in \R$ and decay rate $\lambda>0$. For any set $\X$ that is either a finite set or a real interval, we denote by $\Unif{\X}$ the uniform distribution over $\X$, which belongs to $\Prob{\X}$. 

\paragraph{Marginalization, disintegration.} In what follows, let $\V := \times^d_{i=1} \V_i$, where each $\V_i$ is a Borel space. For every $P \in \mathcal{P}(\V)$ and every nonempty $I \subseteq [d]$, $P_I \in \mathcal{P}(\V_I)$ denotes the marginal distribution of $P$ over the $I$-variables. For every $J \subseteq [d]$ and $P_J$-almost every $v_J$, $P_{\mid J}(\bullet | v_J) \in \mathcal{P}(\V)$ denotes the disintegration of $P$ given the $J$-variables valued at $v_J$ while $P_{I \mid J}(\bullet | v_J) \in \mathcal{P}(\V_I)$ refers to the marginal of $P_{\mid J}(\bullet | v_J)$ over the $I$-variables. We fix $P_{I \mid \emptyset} = P_I$ by convention. Additionally, when $I$ or $J$ are singletons, we use the element rather than the set in the subscript notation (like $P_i$ instead of $P_{\{i\}}$). According to the disintegration formula: for any $B \in \B(\V_{I \cup J})$, $P_{I \cup J}(B) = \int_{\V_J} P_{I \mid J}(B_{v_J} | v_J) \mathrm{d}P_J(v_J)$ where $B_{v_J} := \{ v_I \in \V_I \mid v_{I \cup J} \in B\}$. A \emph{probability kernel} from $\X$ to $\Y$ is a function $K(\bullet | \bullet)$ from $\B(\Y) \times \X$ to $[0,1]$ such that for every $B \in \B(\Y), x \in \X$ the function $K(B | \bullet)$ is measurable and the function $K(\bullet | x)$ belongs to $\mathcal{P}(\Y)$. We denote by $K(\bullet | \X)$ the collection $(K(\bullet | x))_{x \in \X}$. In particular, $P_{I \mid J}(\bullet|\bullet)$ is a probability kernel from $\V_J$ to $\V_I$ \citep[Theorem 9]{pollard2002user}.

\paragraph{Independence.} We say that $I$ is $P$-independent to $J$ if for any $(B_I, B_J) \in \B(\V_I) \times \B(\V_J)$, $P_{I \cup J}(B_I \times B_J) = P_I(B_I) P_J(B_J)$, which we write as $P_{I \cup J} = P_I \otimes P_J$. Moreover, for any $H \subseteq [d]$ (possibly empty) we say that $I$ is $P$-independent to $J$ conditional to $H$ if $I$ and $J$ are $P( \bullet | v_H)$-independent for $P_H$-almost every $v_H$. This equivalently means that $P_{I \mid J \cup H}(\bullet | v_{J \cup H}) = P_{I \mid H}(\bullet | v_H)$  for $P_{J \cup H}$-almost every $v_{J \cup H}$.

\paragraph{Cumulative distribution and quantile functions.} For every $i \in [d] \setminus J$ and $P_J$-almost every $v_J$, we write $F^P_{i \mid J}(\bullet | v_J) : \V_i \to [0,1]$ for the cumulative distribution function of $P_{i \mid J}(\bullet | v_J)$ and $G^P_{i \mid J}(\bullet | v_J) :=  \inf_{v_i \in \V_i} \{ \bullet \leq F^P_{i \mid J}(v_i | v_J) \}$ for its generalized inverse, called the quantile function of $P_{i \mid J}(\bullet | v_J)$. Both $F^P_{i \mid J}(\bullet | v_J)$ and $G^P_{i \mid J}(\bullet | v_J)$ are non-decreasing. Additionally, they satisfy $G^P_{i \mid J}(\bullet | v_J) \# \operatorname{Unif}([0,1]) = P_{i \mid J}(\bullet | v_J)$ regardless of $P$, and $F^P_{i \mid J}(\bullet | v_J) \# P_{i \mid J}(\bullet | v_J) = \Unif{[0,1]}$ if $P_{i \mid J}(\bullet | v_J)$ is atomless. If $I$ and $J$ are $P$-independent conditional to $H$, then $F^P_{I \mid J \cup H}(\bullet | v_{J \cup H}) = F^P_{I \mid K}(\bullet | v_H)$  for $P_{J \cup H}$-almost every $v_{J \cup H}$. Crucially, $G^P_{i|J}(\bullet|\bullet)$ is \emph{jointly} measurable, as a special case of \citep[Theorem 2.1]{carlier2016vector}.

\subsection{Product maps and measure transportation}

\paragraph{Cartesian products of functions.} We consider two operations assembling a (possibly uncountable) family of functions into a single function taking values in a product space. First, let $(f_i)_{i \in I}$ be a collection of functions such that $f_i : \X_i \to \Y_i$ for every $i \in I$. We denote by $\underset{i \in I}{\times} f_i$ the function from $\underset{i \in I}{\times} \X_i$ to $\underset{i \in I}{\times} \Y_i$ given by $\left( \underset{i \in I}{\times} f_i \right)(x_I) := (f_i(x_i))_{i \in I}$. Second, let $(f_i)_{i \in I}$ be a collection of functions such that $f_i : \X \to \Y_i$ for every $i \in I$. We write $f_I$ or $(f_i)_{i \in I}$ (overloading notation) for the function from $\X$ to $\underset{i \in I}{\times} \Y_i$ given by $\left( f_i \right)_{i \in I}(x) := (f_i(x))_{i \in I}$. For example, $(X_1,X_2)$ can be equivalently a pair of random variables from $\Omega$ to $\R$ or a random vector $(X_1,X_2) : \Omega \to \R^2$. The main point to keep in mind is that the symbol $\times$ indicates that each function $f_i$ as its own input in the concatenation. In the case where all sets are Borel spaces and product spaces are equipped with product $\sigma$-algebras, if $f_i$ is measurable for every $i \in I$ then both $\left( \underset{i \in I}{\times} f_i \right)$ and $f_I$ are measurable---even when $I$ is uncountable \citep[Corollary 6.63]{driver2003analysis}. 

\paragraph{Push forward.} For $P \in \mathcal{P}(\X)$ and a measurable map $f : \X \to \Y$ we define the \emph{push-forward measure} of $P$ by $f$ as $f \# P := P \circ f^{-1}$, which belongs to $\mathcal{P}(\Y)$. In terms of random variables, if $X \sim P$ then $f(X) \sim f \# P$. Moreover, we write $f \# K$ for the push-forward of the probability kernel $K$ by $f$ defined by: for every $B \in \B(\Y), x \in \X$, $(f \# K)(B|x) = K(f^{-1}(B) | x)$. If two measurable functions $f$ and $g$ are $P$-almost everywhere equal, then $f \# P = g \# P$. Moreover, if $P \in \mathcal{P}(\X)$, $f : \X \to \Y$ and $g : \Y \to \Z$ are measurable, then $g \# (f \# P) = (g \circ f) \# P$. Given two probability distributions $P$ and $P'$ on possibly distinct domains, we call any measurable function $f$ such that $f \# P = P'$ a \emph{transport map} from $P$ to $P'$. If $f$ is an invertible transport map from $P$ to $P'$, then $f^{-1}$ is a transport map from $P'$ to $P$. We recall that the inverse of a (Borel) measurable map is also measurable \citep{souslin1917}. A coupling $\gamma$ with $P$ as left marginal and $P'$ as right marginal is said to be \emph{deterministic from left to right} if there exists a transport map $f$ from $P$ to $P'$ such that $\gamma = (\Id,f) \# P$, \emph{deterministic from right to left} if there exists a transport map $f$ from $P'$ to $P$ such that $\gamma = (f,\Id) \# P'$, and \emph{deterministic} if it is deterministic in both senses. We underline that being deterministic from left to right is different from being deterministic from right to left in general (see \cref{prop:deterministic_process}).

\paragraph{Monotonic rearrangements.} For every $P,Q \in \Prob{\R}$ such that $P$ is atomless, there exists a ($P$-almost everywhere) unique \emph{non-decreasing transport map} (respectively, \emph{non-increasing transport map}) from $P$ to $Q$: it is given by $G^Q \circ F^P$ (respectively, $G^Q \circ (1-F^P)$). In particular, for $P \in \Prob{\V}$, $G^P_{i \mid J}(\bullet | v_J)$ is the unique non-decreasing transport map from $\Unif{[0,1]}$ to $P_{i \mid J}(\bullet | v_J)$. Note that, if $P$ and $Q$ are absolutely continuous relative to the Lebesgue measure and have finite second-order moments, then the non-decreasing transport map coincides with the optimal-transport map for the squared Euclidean cost \citep{cuesta1989notes,brenier1991polar}.

\subsection{Random variables}

\paragraph{Probability space.} Throughout the paper, we consider a probability space $(\Omega, \Sigma, \P)$ with $(\Omega,\Sigma)$ a Borel space, and $\P : \Sigma \to [0,1]$ a probability measure. This space does not necessarily have a physical interpretation, it abstractly represents the possible underlying states of the world. Crucially, it serves as the common mathematical basis to define and compare random variables.

\paragraph{Laws and independence.} A \emph{random variable} $V$ (including \emph{random vectors}) is a measurable function from $\Omega$ to a Borel subset of an Euclidean space equipped with the Borel $\sigma$-algebra. It produces a probability distribution on its output space: we write $\mathbb{L}(V) := V \# \P$ and $\E[V] := \int V(\omega) \mathrm{d}\P(\omega)$ for respectively the \emph{law and expectation under $\P$} of a random variable $V$. We also use $V \sim P$ to signify that $\law{V} = P$. We emphasize that the laws of univariate random variables can be completely general in this paper, we do not suppose them to be either Lebesgue-absolutely continuous or discrete. For any Borel set $F$, we use the common probability-textbook notation $\{V \in F\}$ for the set $\{\omega \in \Omega \mid V(\omega) \in F\} \in \Sigma$. Two variables are \emph{equal in law under $\P$}, denoted by $\mathbb{L}(V_1) = \mathbb{L}(V_2)$, if $\P(V_1 \in F) = \P(V_2 \in F)$ for every Borel set $F$. The notation $V_1 \independent V_2$ means that $V_1$ and $V_2$ are \emph{independent under $\P$}, that is $\P(V_1 \in F_1, V_2 \in F_2) = \P(V_1 \in F_1) \cdot \P(V_2 \in F_2)$ for all Borel sets $F_1,F_2$.

\paragraph{Conditioning.} We denote by $\P(\bullet | V=v)$ the \emph{regular conditional probability measure with respect to $\{V=v\}$}, which exists and is unique for $\mathbb{L}(V)$-almost every $v$. Then, whenever they are well-defined, we write $\mathbb{L}(V_2 | V_1=v_1) := V_2 \# \P(\bullet | V_1=v_1)$ and $\E[V_2 | V_1=v_1] := \int V_2(\omega) \mathrm{d}\P(\omega | V_1=v_1)$ for respectively the \emph{law and expectation of $V_2$ conditional to $V_1=v_1$}. The expression $V_1 \independent V_2 | V_3$ means that $V_1$ and $V_2$ are \emph{independent conditional to $V_3$ under $\P$}, namely that $V_1$ and $V_2$ are independent under $\P(\bullet | V_3=v_3)$ for $\mathbb{L}(V_3)$-almost every $v_3$.

\subsection{Joint probability distributions with fixed marginals}\label{sec:joint}

\paragraph{Stochastic processes.} Let $\Y$ be a Borel space and $\X$ be a set (possibly infinite or uncountable). A \emph{stochastic (or random) process} with \emph{state space} $\Y$ and \emph{index set} $\X$ is a collection $(Y_x)_{x \in \X}$ of random variables $Y_x : \Omega \to \Y$. It induces a probability distribution $(Y_x)_{x \in \X} \# \P$ in $\mathcal{P}(\Y^\X)$. Two stochastic processes $(Y_x)_{x \in \X}$ and $(Y'_x)_{x \in \X}$ have the same law if for every $W \geq 1$ and $(x_1,\ldots,x_W) \in \X^W$, the finite-dimensional random vectors $(Y_{x_1},\ldots,Y_{x_W})$ and $(Y'_{x_1},\ldots,Y'_{x_W})$ are equal in law.

\paragraph{Projective families of distributions.} Let $\F(\X)$ be the set of finite subsets of $\X$, and define for any $A,B \in \F(\X)$ such that $B \subseteq A$ the projections $\pi^A_B : \Y^A \to \Y^B$ and $\pi_B : \Y^\X \to \Y^B$. A \emph{projective family of distributions} with state space $\Y$ and index set $\X$ is a collection of probability measures $Q := (Q_\xi)_{\xi \in \F(\X)}$ where $Q_\xi \in \mathcal{P}(\Y^\xi)$ for every $\xi \in \F(\X)$ satisfying: for any $A,B \in \F(\X)$ such that $B \subseteq A$, $Q_B = \pi^A_B \# Q_A$. The law $Q$ of a stochastic process $Y$ with state space $\Y$ and index set $\X$ entails the projective family of distributions $(\pi_\xi \# Q)_{\xi \in \F(\X)}$. Conversely, for any projective family of distributions $Q$ with state space $\Y$ and index set $\X$, there exists a stochastic process $Y$ with state space $\Y$ and index set $\X$ entailing $Q$ \citep[Theorem 8.23]{kallenberg1997foundations}.

\paragraph{Measurability and composability} All in all, stochastic-process distributions and projective families of distributions are equivalent. We refer to both of them as \emph{process measures} or \emph{process distributions}, and we write $\Q(\X,\Y)$ for the set of process distributions with state space $\Y$ and index set $\X$. We now turn to technical aspects in the case where $\X$ is a Borel space. A random process $Y$ is said to be \emph{measurable} if the function $(x,\omega) \to Y_x(\omega)$ is measurable. In the probability literature, this property notably serves to ensure that an object like $Y_X$, where $Y$ is a random process and $X$ a random variable, is a well-defined random variable. Given a process measure $Q$ and a random-process representation $Y$, the random \emph{variable} $Y_x$ is by design measurable for every $x \in \X$. However, the random \emph{process} $Y$ is not necessarily measurable without further assumptions \citep[Chapter 3]{doob1953stochastic}. Note that process measurability holds if $\X$ is discrete \citep[Chapter 2]{doob1953stochastic} and for centered Gaussian processes with continuous covariance functions \citep[Section 2.1.1]{henderson2024sobolev}. In the specific context of this work, we qualify a process measure as \emph{composable} if among the random-process representations given by \citep[Theorem 8.23]{kallenberg1997foundations} there exists a measurable one. We denote by $\Q^*(\X,\Y)$ the set of composable process distributions with state space $\Y$ and index set $\X$. In particular, $\Q(\X,\Y) = \Q^*(\X,\Y)$ if $\X$ is countable, as a consequence of a previous remark. We refer to \cref{rem:measurability} for more details.

This technicality plays a significant role in many causal frameworks, although the literature rarely highlights it. It is notably a necessary assumption of the potential-outcome framework (detailed in \cref{sec:po}) to write the so-called \emph{consistency rule} with uncountable treatment values. While \cite{abbring2005social} explicitly suppose measurability in such a scenario, this hypothesis is generally left implicit. We emphasize that some of our theoretical results require composable measures. Nonetheless, one can understand the concepts and methodology we develop without minding composability.

\paragraph{Joint probability distributions.}  

We say that a process measure $Q$ is \emph{deterministic} if there exist a probability distribution $P$ and an injective function $f$ such that $Q = f \# P$. This definition is consistent with the notion of deterministic coupling, as reminded (later) by \cref{prop:deterministic_process}. For any collection $(P^{(x)})_{x \in \X}$ of marginal distributions $P^{(x)} \in \mathcal{P}(\Y)$, we denote by $\Joint{\X}{\Y}{(P^{(x)})_{x \in \X}}$ the set of process distributions $Q$ such that $Q_x = P^{(x)}$ for all $x \in \X$. Similarly, we write $\Jointc{\X}{\Y}{(P^{(x)})_{x \in \X}}$ for the set of composable process distributions $Q$ such that $Q_x = P^{(x)}$ for all $x \in \X$. Finally, we use the shorthands $\Norma{\X} := \Joint{\X}{\R}{(\Norm)_{x \in \X}}$ and  $\Normac{\X} := \Jointc{\X}{\R}{(\Norm)_{x \in \X}}$.

\begin{table}
\caption{Some notation}
\begin{tabularx}{\textwidth}{@{}p{0.25\textwidth}X@{}}
\toprule
  \textbf{Probability} \\
  $\mathcal{P}(\V)$ & Set of Borel probability distributions over the space $\V$\\
  $P_{I | J}$ & Disintegration/marginalization over variables $I$/$J$ of $P \in \mathcal{P}(\V)$\\
  $G^P_{i | J}(\bullet|v_J)$ & Quantile function of $P_{i | J}(\bullet|v_J)$, where $P \in \mathcal{P}(\V)$\\
  $\Q(\X,\Y)$ & Set of process measures with state space $\Y$ and index set $\X$\\
  $\Q_{\text{joint}}(\X,\Y,(P^{(x)})_{x \in \X})$ & Measures in $\Q(\X,\Y)$ with marginals $(P^{(x)})_{x \in \X}$ in $\mathcal{P}(\Y)$\\
  $Q_{x_1,\ldots,x_W}$ & Projection of $Q \in \Q(\X,\Y)$ over the indices $(x_w)^W_{w=1} \in \X^W$\\
  \textbf{Causality} \\
  $i \in [d]$ & Index of an endogenous variable \\
  $w \in [W]$ & Index of a parallel world \\
  $\phi_{w,i}$ & Intervention transformation in world $w$ on variable $i$ \\
  $\C \in \CGM$ & Directed acyclic causal graphical model \\
  $K_i$ & Causal kernel of variable $i$ \\
  $\M \in \SCM$ & Markovian structural causal model \\
  $(f_i,P_{0,i})$ & Causal mechanism and exogenous distribution of variable $i$ \\
  $\A \in \CR$ & Counterfactual model \\
  $S^{(i)}$ & One-step-ahead counterfactual distribution of variable $i$\\
\bottomrule
\end{tabularx}\label{table:notation}
\end{table}

\section{Pearl's causal framework}\label{sec:setup}

This section presents Pearl's causal modeling from the bottom to the top of the causality ladder. It revisits well-known fundamental results of \cite{bongers2021foundations} and \cite{bareinboim2022pearl} by extending them to more general types of interventions than perfect interventions. This serves to keep the paper self contained by introducing key concepts and notation for the rest of this work. It starts in \cref{sec:cgm} with \emph{causal graphical models}, which allow causal inference at the interventional level, and continues in \cref{sec:scm} with \emph{structural causal models}, which allow causal inference up to the counterfactual level. We emphasize that we focus on \emph{Markovian models}: acyclic models with no hidden confounders.

Throughout, $d \geq 1$ is an integer representing the number of observational variables in a study. Each variable $i \in [d]$ takes values in a space $\V_i \subseteq \R$. The observations behave accordingly to a probability distribution $P$ over $\V := \times^d_{i=1} \V_i \subseteq \R^{d}$ called the \emph{observational (or endogenous) distribution}. Estimating features of $P$ enables one to make predictions from observations, as this distribution captures associations between variables. Nevertheless, it does not provide causal insight. The two types of models that we present below can be understood as additional information on top of $P$ to describe cause-effect relationships.

\subsection{Causal graphical models}\label{sec:cgm}

Association between two random variables does not distinguish between cause and effect as it is a symmetric notion. The first type of causal models addresses this limitation: it completes observational knowledge with a directed graph specifying cause-effect arrows between variables.  

\begin{definition}[Directed acyclic graphical model]
A \emph{directed acyclic graphical model} $\C$ over $d$ variables is tuple $(\V,\G,\K)$ where
\begin{enumerate}
    \item $\V := \times^d_{i=1} \V_i \subseteq \R^{d}$ is a Cartesian product of Borel subsets of $\R$, called the \emph{space of endogenous values};
    \item $\G$ is a directed acyclic graph (DAG) with nodes $[d]$ and parent function $\pa : [d] \to 2^{[d]}$, called the \emph{causal graph};
    \item $\K := (K_i)^d_{i=1}$ is a collection of probability kernels $K_i : \B(\V_i) \times \V_{\pa(i)} \to [0,1]$, called the \emph{causal kernels}.
\end{enumerate}
We denote by $\CGM_{\V}$ the set of directed acyclic graphical models with $\V$ as space of endogenous values.
\end{definition}

For simplicity, we refer to directed acyclic graphical models as \emph{causal graphical models} (CGMs). The result below formalizes how such a model entails the observational distribution $P$.

\begin{proposition}[Observational distribution]\label{prop:cgm_obs}
Let $\V := \times^d_{i=1} \V_i \subseteq \R^{d}$ be a Cartesian product of Borel subsets of $\R$. Every $\C \in \CGM_{\V}$ entails a probability measure $P \in \mathcal{P}(\V)$ given by
\[
\mathrm{d}P(v_1,\ldots,v_d) = \prod^d_{i=1} \mathrm{d} K_i(v_i | v_{\pa(i)}).
\]
It is such that for every $i \in [d]$ and every $v_{\pa(i)} \in \V_{\pa(i)}$, $P_{i | \pa(i)}(\bullet | v_{\pa(i)}) = K_i(\bullet | v_{\pa(i)})$. We call it the \emph{observational distribution} entailed by $\C$, and denote it by $\Obs{C}(\C)$.
\end{proposition}

Importantly, there always exists a causal model underlying a given observational distribution, as explained in the next proposition. We point out that such a model is, however, not uniquely determined by this distribution.

\begin{proposition}[Existence of a causal graphical model]\label{prop:cgm_exists}
Let $\V := \times^d_{i=1} \V_i \subseteq \R^{d}$ be a Cartesian product of Borel subsets of $\R$. For every $P \in \mathcal{P}(\V)$, there exists $\C \in \CGM_{\V}$ such that $\Obs{C}(\C) = P$.
\end{proposition}

A key concept of the causality literature is the notion of intervention. An intervention represents the action of altering some variables of the system while keeping the rest of the model untouched. In this work, we consider the following class of actions.

\begin{definition}[Intervention transformation]\label{def:intervention_transfo}
Let $\V := \times^d_{i=1} \V_i \subseteq \R^{d}$ be a Cartesian product of Borel subsets of $\R$. An \emph{intervention transformation} on $\V$ is a map $\Phi := \underset{i \in [d]}{\times} \phi_i$ with $\phi_i : \V_i \to \V_i$ measurable for every $i \in [d]$. We denote by $\I(\V)$ the set of intervention transformations on $\V$. Additionally, for every $I \subseteq [d]$ and $v^\star_I \in \V_I$ we denote by $\doint(I,v^\star_I)$ the intervention transformation $\Phi \in \I(\V)$ given by $\phi_i : v_i \mapsto v^\star_i$ if $i \in [d]$ and $\phi_i = \Id$ otherwise, called a \emph{perfect intervention}.
\end{definition}

An intervention $\Phi$ specifies how to alter each variable of the system. If, for example, one would like to understand the downstream effect of fixing variable n°1 to the value $v^\star_1 \in \V_1$, then they must choose $\doint(1,v^\star_1)$. Formally, an intervention transforms a causal model, leading to a modified endogenous distribution. 

\begin{definition}[Intervention on graphical models]\label{def:cgm_int}
Let $\V := \times^d_{i=1} \V_i \subseteq \R^{d}$ be a Cartesian product of Borel subsets of $\R$. For any $\C := (\V,\G, \K) \in \CGM_{\V}$ and $\Phi \in \I(\V)$, the intervention by $\Phi$ maps $\C$ to $\C^{\Phi} := (\V,\G,\K^\Phi) \in \CGM_{\V}$ where $\K^{\Phi} := (K^\Phi_i)^d_{i=1}$ is defined by $K^\Phi_i = \phi_i \# K_i$ for every $i \in [d]$. We call $\Int{C}(\C,\Phi) := \Obs{C}(\C^\Phi)$ the \emph{interventional distribution} of $\C$ under $\Phi$.
\end{definition}
The graphical framework we presented is inspired by the one of \cite{dance2024causal}. Notably, we consider more general interventions than fixing variables to constant values (namely, perfect do-interventions), as studied by \cite{sani2020identification}. Remark that in this configuration the graph remains the same after the surgery.

\subsection{Structural causal models}\label{sec:scm}

We now turn to the second type of causal models, which can be seen as augmented versions of graphical models.

\begin{definition}[Markovian structural causal model]\label{def:scm}
A \emph{Markovian structural causal model} $\M$ over $d$ variables is tuple $(\V,\G,\U,P_0,f)$ where
\begin{enumerate}
    \item $\V := \times^d_{i=1} \V_i \subseteq \R^{d}$ is a Cartesian product of Borel subsets of $\R$ and $\U := \times^d_{i=1} \U_i$ is a Cartesian product of Borel spaces, called, respectively, the space of \emph{endogenous variables} and the space of \emph{exogenous variables},
    \item $\G$ is a DAG with nodes $[d]$ and parent function $\pa : [d] \to 2^{[d]}$,
    \item $P_0 := \otimes^d_{i=1} P_{0,i}$ is a product probability distribution on $\U$, such that $P_{0,i} \in \mathcal{P}(\U_i)$, called the \emph{exogenous distribution},
    \item $f := (f_i)^d_{i=1}$ is a collection of measurable maps, called the \emph{causal mechanism}, such that $f_i : \V_{\pa(i)} \times \U_i \to \V_i$.
\end{enumerate}
We denote by $\SCM_{\V}$ the set of Markovian structural causal models with $\V$ as space of endogenous values. A solution $(V,U)$ of $\M \in \SCM_{\V}$ is a pair of random variables $V : \Omega \to \V$ and $U : \Omega \to \U$ such that $\L(U) = P_0$ and $V_i = f_i(V_{\pa(i)},U_i)$ for every $i \in [d]$.
\end{definition}

A structural causal model specifies how endogenous (observational) variables $V$ are generated from exogenous (latent) variables $U$. The above definition focuses on Markovian models, in which the exogenous variables are mutually independent and each endogenous variables is affected by at most one exogenous variable. For simplicity, we will frequently refer to Markovian structural causal models as \emph{structural causal models} (SCMs).

We point out that the measurability condition of $f$ is frequently left implicit in the literature. It is nonetheless a requirement to write assignments like $V_i = f_i(V_{\pa(i)},U_i)$, and we will address it with care throughout the paper.

\subsubsection{Observational and interventional levels}

The properties of Markovian SCMs make their solutions satisfy a local Markov property described below. For any graph $\G$ over $[d]$, we denote by $\desc(i)$ the set of descendants of $i$ in $\G$ and by $\ndesc(i)$ its complementary in $[d]$. Similarly, we denote by $\an(i)$ the set of ancestors of $i$ and by $\nan(i)$ its complementary. Both $\desc(i)$ and $\an(i)$ contain $i$.

\begin{lemma}[Markov property of solutions]\label{lem:markov}
Let $\V := \times^d_{i=1} \V_i \subseteq \R^{d}$ be a Cartesian product of Borel subsets of $\R$. Every $\M \in \SCM_{\V}$ admits a solution. Each solution $(V,U)$ satisfies the following properties for every $i \in [d]$:
\begin{enumerate}
    \item $U_i \independent V_{\ndesc(i)}$ and $V_i \independent U_{\nan(i)}$, in particular $U_i \independent V_{\pa(i)}$;
    \item $\law{V_i | V_{\pa(i)} = v_{\pa(i)}} = \law{f_i(v_{\pa(i)}, U_i)}$ for every $v_{\pa(i)} \in \V_{\pa(i)}$;
    \item $V_i \independent V_{\ndesc(i)} \mid V_{\pa(i)}$.
\end{enumerate}
\end{lemma}

Moreover, an SCM entails an observational distribution, fully determined by the source of randomness $P_0$ and the causal mechanism $f$. This characterizes its observational level.

\begin{proposition}[Observational distribution]\label{prop:scm_obs}
Let $\V := \times^d_{i=1} \V_i \subseteq \R^{d}$ be a Cartesian product of Borel subsets of $\R$. For any $\M \in \SCM_{\V}$, the law of $V$ for every solution $(V,U)$ is unique. We call it the \emph{observational distribution} of $\M$ and denote it by $\Obs{M}(\M)$.
\end{proposition}

Similar to graphical models, we define interventions for structural models. This determines the interventional level.

\begin{definition}[Intervention]\label{def:scm_int}
Let $\V := \times^d_{i=1} \V_i \subseteq \R^{d}$ be a Cartesian product of Borel subsets of $\R$. For any $\M := (\V, \G, \U, P_0,f) \in \SCM_{\V}$ and measurable product map $\Phi \in \I(\V)$, the intervention by $\Phi$ maps $\M$ to $\M^{\Phi} := (\V,\G,\U,P_0,\tilde{f}) \in \SCM_{\V}$ where $\tilde{f}_i := \phi_i \circ f_i$ for every $i \in [d]$. We define $\Int{M}(\M, \Phi) := \Obs{M}(\M^\Phi)$ the \emph{interventional distribution of $\M$ relatively to $\Phi$}. 
\end{definition}

We now turn to a conceptually important result. The observational and interventional distributions of an SCM correspond to those of a certain CGM. This means that the first two rungs of the causality ladder can be framed equivalently with SCMs or CGMs.

\begin{proposition}[Entailed causal graphical model]\label{prop:scm_cgm}
Let $\V := \times^d_{i=1} \V_i \subseteq \R^{d}$ be a Cartesian product of Borel subsets of $\R$, let $\M := (\V,\G,\U,P_0,f) \in \SCM_{\V}$, and define the probability kernels $\K := (K_i)^d_{i=1}$ given by: for every $i \in [d]$ and every $v_{\pa(i)} \in \V_{\pa(i)}$, $K_i(\bullet|v_{\pa(i)}) := f_i(v_{\pa(i)},\bullet) \# P_{0,i}$.\footnote{This can also be written as $K_i(\bullet|v_{\pa(i)}) := \mathbb{L}(f_i(v_{\pa(i)},U_i))$ where $\L(U_i) = P_{0,i}$.} Then, the tuple $\C_\M := (\V,\G,\K)$ belongs to $\CGM_{\V}$ and is such that:
\begin{enumerate}
    \item $\Obs{M}(\M) = \Obs{C}(\C_\M)$,
    \item $\C_{\M^\Phi} = \C^\Phi_\M$ and $\Int{M}(\M,\Phi) = \Int{C}(\C_\M,\Phi)$ for every $\Phi \in \I(\V)$.
\end{enumerate} 
We call $\C_\M$ the \emph{entailed causal graphical model} of $\M$. For every $\C \in \CGM_{\V}$, we define $\SCM_{\V,\C} := \{ \M \in \SCM_{\V} \mid \C_\M = \C \}$.
\end{proposition}

Basically, the quantity $f_i(v_{\pa(i)}, U_i)$ from an SCM can be seen as a random-variable representation of the causal kernel $K_i$ from a CGM. The next proposition ensures that one can always construct a structural model on top of a graphical model.

\begin{proposition}[Existence of a structural causal model]\label{prop:scm_exists}
Let $\V := \times^d_{i=1} \V_i \subseteq \R^{d}$ be a Cartesian product of Borel subsets of $\R$. The two following properties hold.
\begin{enumerate}
    \item For every $\C \in \CGM_{\V}$, $\SCM_{\V,\C} \neq \emptyset$.
    \item For every $P \in \mathcal{P}(\V)$, there exists $\M \in \SCM_{\V}$ such that $\Obs{M}(\M)=P$.
\end{enumerate}
\end{proposition}
This proposition and its proof are very similar to \citep[Proposition~9]{peters2014causal}, the main difference being that we do not make any assumption on the observational distribution $P$: it need not be absolutely continuous with respect to the Lebesgue measure.

To later introduce canonical representations of SCMs, we need to clarify which part of a causal model corresponds to which rung of the causal hierarchy. This analysis requires classical notions of equivalence between SCMs.

\begin{definition}[Observational and interventional equivalence]\label{def:int_equi}
Let $\V := \times^d_{i=1} \V_i \subseteq \R^{d}$ be a Cartesian product of Borel subsets of $\R$. For every $\M_1,\M_2 \in \SCM_{\V}$, we say that $\M_1$ and $\M_2$ are
\begin{enumerate}
    \item \emph{observationally equivalent} if $\Obs{M}(\M_1) = \Obs{M}(\M_2)$,
    \item \emph{interventionally equivalent} if $\Int{M}(\M_1,\Phi) = \Int{M}(\M_2,\Phi)$ for every $\Phi \in \I(\V)$.
\end{enumerate}
\end{definition}

Then, we show that, in a Markovian SCM, interventional knowledge is fully captured by the observational distribution and the causal graph, that is, by the entailed CGM.

\begin{proposition}[Interventional versus graphical assumptions]\label{prop:int_is_graph}
Let $\V := \times^d_{i=1} \V_i \subseteq \R^{d}$ be a Cartesian product of Borel subsets of $\R$. If $\M_1,\M_2 \in \SCM_{\V}$ have the same graph $\G$ and are such that $\Obs{M}(\M_1) = \Obs{M}(\M_2)$, then $\C_{\M_1} = \C_{\M_2}$. Therefore, $\M_1$ and $\M_2$ are interventionally equivalent.
\end{proposition}

Said differently, if two SCMs are compatible with the same graph $\G$ and entail the same observational distribution $P$, which means that they entail the same CGM, then they are interventionally equivalent. 

Nevertheless, an SCM encodes more causal assumptions than its entailed CGM. Not only the functional mechanism and the exogenous distribution produce causal kernels before and after intervention, given by the laws of $f_i(\bullet, U_i)$ and $f^\Phi_i(\bullet, U_i)$, but they also allow to reason about their \emph{joint} distribution given by the law of $\left( f_i(\bullet, U_i), f^\Phi_i(\bullet, U_i) \right)$ for the same $U_i$. This step up from interventional marginals to coupling across them amounts to climbing the causality ladder from the second to the third rung: the counterfactual level, that we address thereafter.

\subsubsection{Counterfactual level}

While the interventional rung focuses on single-world phenomena corresponding to general causes, the counterfactual rung focuses on cross-world phenomena corresponding to singular causes. For example, \say{the medical treatment works} is an interventional statement, whereas \say{had Alice taken the treatment she would have recovered} is a counterfactual statement. Mathematically, the interventional level concerns families of marginal distributions while the counterfactual level concerns joint probability distributions over these marginals. We define counterfactuals in SCMs via the notion of cross-world solutions.

\begin{definition}[Cross-world solution]
Let $\V := \times^d_{i=1} \V_i \subseteq \R^{d}$ be a Cartesian product of Borel subsets of $\R$, $\M \in \SCM_{\V}$, and $(\Phi_w)^W_{w=1} \in \I(\V)^W$, where $W \geq 1$. A \emph{cross-world solution} of $\M$ with respect to $(\Phi_w)^W_{w=1}$ is a random tuple $(V^{\Phi_1},\ldots,V^{\Phi_W},U)$ such that each $(V^{\Phi_w},U)$ is a solution of $\M^{\Phi_w}$. Importantly, this the same exogenous vector $U$ for all $w \in [W]$.
\end{definition}

The idea is that each block $V^{\Phi_w}$ of the cross-world solution represents the outcome in a parallel world where intervention $\Phi_w$ took place. These outcomes are entangled by a same common cause $U$. Similar to classical solutions, cross-world solutions meet a local Markov property.

\begin{proposition}[Joint Markov property]\label{prop:block_markov}
Let $\V := \times^d_{i=1} \V_i \subseteq \R^{d}$ be a Cartesian product of Borel subsets of $\R$ and $(\Phi_w)^W_{w=1} \in \I(\V)^W$, where $W \geq 1$. Each $\M := (\V,\G,\U,P_0,f) \in \SCM_{\V}$ admits a cross-world solution relatively to $(\Phi_w)^W_{w=1}$. Every cross-world solution, denoted by $(V^{\Phi_1},\ldots,V^{\Phi_W},U)$, satisfies for every $i \in [d]$:
\begin{enumerate}
    \item $U_i \independent (V^{\Phi_w}_{\ndesc(i)})^W_{w=1}$ and $(V^{\Phi_w}_{i})^W_{w=1} \independent U_{\nan(i)}$, in particular $U_i \independent (V^{\Phi_w}_{\pa(i)})^W_{w=1}$;
    \item $\mathbb{L}\left((V^{\Phi_w}_i)^W_{w=1} \mid (V^{\Phi_w}_{\pa(i)})^W_{w=1} = (v^{(w)}_{\pa(i)})^W_{w=1} \right) = \mathbb{L}\left( \left( (\phi_{w,i} \circ f_i)(v^{(w)}_{\pa(i)}, U_i) \right)^W_{w=1} \right)$ for every $(v^{(w)}_{\pa(i)})^W_{w=1} \in \V^W_{\pa(i)}$;
    \item $V_i \independent (V^{\Phi_w}_{\ndesc(i)})^W_{w=1} \mid (V^{\Phi_w}_{\pa(i)})^W_{w=1}$.
\end{enumerate}
\end{proposition}

The cross-world solution $(V^{\Phi_w})^W_{w=1}$ induces a joint probability distribution over $\V^W$ with interventional laws as marginals, called a \emph{counterfactual distribution}. Counterfactual distributions enable one to reason counterfactually by placing probabilities over cross-world statements.

\begin{proposition}[Counterfactual distribution]\label{prop:counterfactual_law}
Let $\V := \times^d_{i=1} \V_i \subseteq \R^{d}$ be a Cartesian product of Borel subsets of $\R$. For any $\M \in \SCM_{\V}$ and $(\Phi_w)^W_{w=1} \in \I(\V)^W$, where $W \geq 1$, all cross-world solutions relatively to $(\Phi_w)^W_{w=1}$ share the same probability distribution, called the \emph{counterfactual distribution} of $\M$ relatively to $(\Phi_w)^W_{w=1}$ and denoted by $\Ctf{M}(\M,(\Phi_w)^W_{w=1})$. Moreover, $\Ctf{M}(\M,(\Phi_w)^W_{w=1}) \in \Joint{\{\Phi_w\}^W_{w=1}}{\V}{\left(\Int{M}(\M,\Phi_w)\right)^W_{w=1}}$.
\end{proposition}

Let us make important comments regarding vocabulary. Strictly speaking, observational and interventional distributions are also \emph{joint} probability distributions, since the domain $\V$ is a Cartesian product (in general). Concretely, they encode dependencies between multiple variables observed in a \emph{same world}. We mostly reserve the term \say{joint} to counterfactual distributions to emphasize their fundamental characteristic:  they encode \emph{joint cross-world} dependencies. In the same vein, even when a single intervention represents a contrary-to-fact event, we call the produced distribution interventional. This serves to clarify at which causality layer we operate. Several references employ terms like \say{joint}
 and \say{counterfactual} at the interventional level (see \cref{sec:comparison}).

Note that other approaches to formally define counterfactual distributions (with possibly non-Markovian SCMs) consist in intervening on \emph{twin SCMs} \citep[Section 2.5]{bongers2021foundations} or \emph{counterfactual SCMs} \citep[Section 6.4]{peters2017elements}.\footnote{A twin SCM only produces counterfactual \emph{couplings} (with two marginals), but the principle can be readily extended to address counterfactual joint probability distributions (with many marginals) as we do.} In our setting, we prefer to rely on cross-world solutions rather than such models because both the twin SCM and the counterfactual SCM of a Markovian SCM are not necessarily Markovian. As such, this avoids introducing a more general concept of SCMs just for the purpose of defining counterfactuals.

\Cref{prop:block_markov} has critical consequences on counterfactual reasoning with Markovian models. Because counterfactual distributions must satisfy the joint Markov property, not all joint probability distributions across given interventional marginals correspond to the counterfactual distribution of some Markovian SCM. This means that some counterfactual conceptions are not admissible given a CGM. In \cref{sec:ctf_models}, we characterize the set of admissible joint probability distributions. Having specified the strongest causality rung in SCMs, we extend the notions of equivalence between models to this level.

\begin{definition}[Counterfactual equivalence]
Let $\V := \times^d_{i=1} \V_i \subseteq \R^{d}$ be a Cartesian product of Borel subsets of $\R$. For every $\M_1, \M_2 \in \SCM_{\V}$, we say that $\M_1$ and $\M_2$ are \emph{counterfactually equivalent} if for every $W \geq 1$ and $(\Phi_w)^W_{w=1} \in \I(\V)^W$, $\Ctf{M}(\M_1,(\Phi_w)^W_{w=1}) = \Ctf{M}(\M_2,(\Phi_w)^W_{w=1})$. 
\end{definition}
The next result formalizes the hierarchy between causal-inference layers for SCMs.
\begin{proposition}[Causal hierarchy]\label{prop:hierarchy}
Let $\V := \times^d_{i=1} \V_i \subseteq \R^{d}$ be a Cartesian product of Borel subsets of $\R$ and $\M_1, \M_2 \in \SCM_{\V}$. If $\M_1$ and $\M_2$ are counterfactually equivalent, then they are interventionally equivalent. If $\M_1$ and $\M_2$ are interventionally equivalent, then they are observationally equivalent.
\end{proposition}

As such, two counterfactually equivalent SCMs are undistinguishable at any task concerned with queries from the ladder of causation. We then turn to a fundamental result regarding the causality ladder and causal models. The proposition below shows that CGMs do not capture counterfactual assumptions. This means that SCMs sharing the same graph and the same observational distributions are not necessarily counterfactually equivalent.

\begin{proposition}[Counterfactual versus graphical assumptions]\label{prop:scm_vs_cgm}
There exist $\V := \times^d_{i=1} \V_i \subseteq \R^{d}$ a Cartesian product of Borel subsets of $\R$ and $\M_1,\M_2 \in \SCM_{\V}$ such that $\C_{\M_1} = \C_{\M_2}$ and $\M_1$ is not counterfactually equivalent to $\M_2$.
\end{proposition}

By overloading terminology, we call any combination of marginalizations and desintegrations of, respectively, an observational, interventional, or counterfactual distribution of a causal model (be it a CGM or an SCM) an observational, interventional, or counterfactual distribution, respectively.

\begin{remark}[A mathematical view on \say{CGMs versus SCMs}]\label{rem:equations}
\Cref{prop:scm_vs_cgm} shows that CGMs are underspecified to completely identify the counterfactual layer, in contrast to SCMs. The difference in mathematical content between these two types of models explains this dissimilarity. Let us further detail how it operates.

A CGM $\C \in \CGM_{\V}$ places a probability kernel $K_i$ at each node $i \in [d]$ of a graph. Instead, an SCM $\M \in \SCM_{\V,\C}$ places a pair $(f_i,P_{0,i})$ at each node $i \in [d]$ of the same graph, such that $f_i(v_{\pa(i)},\bullet) \# P_{0,i} = K_i(\bullet|v_{\pa(i)})$. Basically, $\M$ furnishes a functional representation of each causal kernel. In general, such a representation is not unique: one can find another pair $(\tilde{f}_i,\tilde{P}_{0,i})$ such that $\tilde{f}_i(v_{\pa(i)},\bullet) \# \tilde{P}_{0,i} = K_i(\bullet|v_{\pa(i)})$ \citep[Proposition 8.20]{kallenberg1997foundations}. Because they describe the same probability kernel (and as such, the same conditional probability distribution) the two mechanisms $(f_i,P_{0,i})$ and $(\tilde{f}_i,\tilde{P}_{0,i})$ may seem undistinguishable. However, this crucially depends on the considered causal-inference layer. While observational and interventional equivalence actually hold, it is not generally the case for counterfactual equivalence. Mathematically, this comes from the fact that
\[
f_i(v_{\pa(i)},\bullet) \# P_{0,i} = \tilde{f}_i(v_{\pa(i)},\bullet) \# \tilde{P}_{0,i},
\]
whereas, in general,
\[
\left( f_i(v_{\pa(i)},\bullet), f_i(v'_{\pa(i)},\bullet) \right) \# P_{0,i} \neq \left( \tilde{f}_i(v_{\pa(i)},\bullet), \tilde{f}_i(v'_{\pa(i)},\bullet) \right) \# P_{0,i}.
\]

All in all, there generally exist an infinite number of representations $(f_i,P_{0,i})$ of a same probability kernel $K_i$. From a causal perspective, navigating through these possible representations amounts to navigating through the admissible counterfactual assumptions relative to given interventional constraints.\footnote{Note that there also exist \emph{distinct} pairs of $(f_i,P_{0,i})$ representing the same kernel and producing the \emph{same} counterfactuals (see \cref{prop:concise}).} One can choose without loss of generality a specific representation for observational and interventional tasks but not for counterfactual ones. This significance of structural assignements can be hidden as long as one looks marginally at an endogenous solution rather than jointly at a cross-world endogenous solution. This is why specifying the causal-inference layer is particularly crucial when working with SCMs.

On this basis, we caution against statements like \say{$V_i = f_i(V_{\pa(i)},U_i)$ and $V_i = \tilde{f}_i(V_{\pa(i)},\tilde{U}_i)$ are basically the same assignment equations} or \say{without loss of generality one can write $V_i = f_i(V_{\pa(i)},U_i)$ with $U_i$ atomless and $f_i$ non-decreasing in $U_i$}. They can be misleading, especially in papers addressing counterfactual reasoning (as in \citep{plevcko2020fair,javaloy2024causal}). \emph{As soon as singular causation matters, the only valid justification for using a specific functional representation (namely, an SCM) is a choice of counterfactual conception.} See also \cref{rem:discussing}.
\end{remark}

To sum-up, the causal graph \emph{constrains} counterfactual distributions (\cref{prop:block_markov}) but does not determine them (\cref{prop:scm_vs_cgm}); the addition of functional constraints \emph{determines} them.

\subsection{Comments}

This section concludes the reminder on Pearl's causal framework by clarifications on the causality ladder. More specifically, it explains which conceptual and mathematical parts of the ladder we focus on, in order to make precise the scope of our contributions from the sections to come.

\subsubsection{Counterfactual inference with graphical models}

As demonstrated above, the entailed CGM of an SCM fully covers the interventional level but is underspecified at the counterfactual level. Nevertheless, the entailed CGM may identify some (but generally not all) counterfactual features: under specific graphical assumptions, some derivatives of counterfactual distributions can be expressed with interventional laws only. Notably, a purely graphical condition like the \emph{back-door criterion} enables analysts to answer contrary-to-fact questions \citep[Theorem 4.3.1]{pearl2016causal}. This means, in particular, that some counterfactual statements can be empirically tested by having falsifiable graphical implications \citep{shpitser2007what}. People familiar with the \emph{do operator} can follow a simple formal rule: a counterfactual effect is identifiable in a CGM if it can be written with the do operator. \Cref{rem:general_cause} below exemplifies counterfactuals that can be identified by a CGM.

\begin{figure}
    \centering
    \includegraphics[page=1]{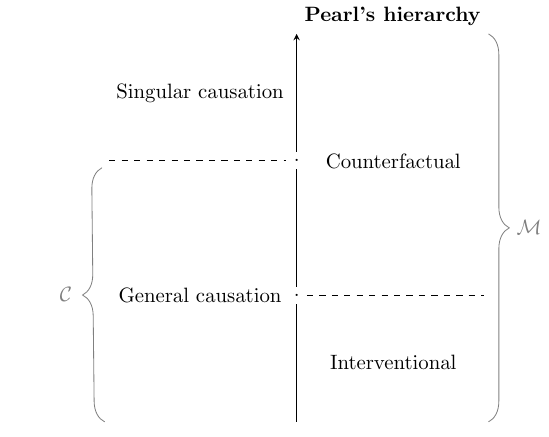}
    \caption{Layers of causation}
    \label{fig:hierarchy}
\end{figure}

This signifies that the split between GCMs and SCMs does not exactly align with the split between the interventional and counterfactual layers. According to \citep[Chapter 7]{pearl2009causality}, it conceptually corresponds to the distinction between \emph{general} and \emph{singular} causes. This distinction can help underlining the objective of our work: rather than addressing the whole counterfactual layer, it focuses on the modeling of singular causation within the counterfactual layer. We do not further explain these conceptual categories, as such a discussion lies beyond the scope of this paper. Actually, our objective has a concrete mathematical formulation that avoids philosophical debates. Assuming that $\C \in \CGM_{\V}$ is given, it focuses exactly on modeling probability distributions that \emph{can} be described by an $\M \in \SCM_{\V,\C}$ \emph{but not} by $\C$ alone. Note that this formal gap between CGMs and SCMs almost never collapse in a measure-theoretic sense \citep[Theorem 27.1]{bareinboim2022pearl}. \Cref{fig:hierarchy} further illustrates these points. Throughout the paper, we informally call a \emph{counterfactual conception} or \emph{counterfactual assumption} a vision of causation determining causal-inference statements beyond what a CGM can identify. 

\begin{remark}[Counterfactuals and general causes]\label{rem:general_cause}

Consider an SCM $\M := ( \V, \G, \U, f, P_0 )$ over the three variables $\{ \mathtt{x,t,y} \}$ and such that $\V_{\mathtt{t}} := \{0,1\}$. This could describe a problem with a covariate $\mathtt{x}$, a cause of interest $\mathtt{t}$, and an outcome $\mathtt{y}$. We focus on $\doint(\mathtt{t},0)$ and $\doint(\mathtt{t},1)$. We also write $V :=  (X,T,Y)$, $V^{\doint(\mathtt{t},0)} := (X_0,0,Y_0)$, and $V^{\doint(\mathtt{t},1)} := (X_1,1,Y_1)$ for the solutions of respectively $\M$, $\M^{\doint(\mathtt{t},0)}$, and $\M^{\doint(\mathtt{t},1)}$ for a same exogenous noise $U \sim P_0$. Finally, we assume that $\M$ is such that $0 < \P(T=1 | X) < 1$ almost surely (note that it is a statistically testable hypothesis).

Analysts typically aim at computing treatment effects such as $\E[Y_1 - Y_0]$ or $\E[Y_1 - Y_0 | X=x]$ without access to the complete causal mechanism (that is, the SCM); just by knowing the graph $\G$ and the observational distribution $P$ (that is, the CGM). These effects are features of $\law{Y_t | X=x}$ for $t \in \{0,1\}$, which are \emph{counterfactual} distributions in general: notice that $\law{Y_t \mid X=x}$ is a disintegration of $\law{(X,Y_t)}$, which is a marginalized law of the cross-world solution $(V,V^{\doint(\mathtt{t},t)})$. $\C_\M$ can identify these effects under specific assumption. Suppose that $\G$ is given by $\mathtt{x} \rightarrow \mathtt{t}$, $\mathtt{x} \rightarrow \mathtt{y}$, and $\mathtt{t} \rightarrow \mathtt{y}$. This typically corresponds to a treatment model. First, since $\mathtt{x}$ is not a descendant of $\mathtt{t}$ in $\G$, the solutions can be simplified as $V^{\doint(\mathtt{t},0)} = (X,0,Y_0)$ and $V^{\doint(\mathtt{t},1)} = (X,1,Y_1)$. Consequently, in this configuration, the counterfactual law $\law{Y_t | X=x}$ coincide with the interventional law $\law{Y_t | X_t=x}$ for $t \in \{0,1\}$. Second, as expected, such interventional laws are identifiable by $P := \law{V}$, since $\law{Y_t | X=x} = \law{Y | X=x, T=t}$ for every $t \in \{0,1\}$, due to the exogenous variables being independent.

This shows how one can carry out cross-layer causal inference in particular situations. But critically, some quantities remain unidentifiable by only $\C_\M$. For example, $\E[Y_1 - Y_0 | X=x, Y = y]$ comes from $\law{(V,V^{\doint(\mathtt{t},1)})}$ but cannot be simplified as a feature of interventional laws. Note that $\E[Y_1 - Y_0 | X=x]$ corresponds to a group causal effect while $\E[Y_1 - Y_0 | X=x, Y = y]$ corresponds to an individual causal effect. The former informs about the potential outcomes in the treated group versus the non-treated group within the population meeting $X=x$; the latter informs about the potential outcomes of single individuals within the population meeting $X=x$ across two distinct treatment statuses.
\end{remark}

\subsubsection{Falsifiability}

Many causality scholars interpret the absence of an arrow $\texttt{t} \rightarrow \texttt{y}$ in a CGM as: in every randomized controlled trial with treatment $\texttt{t}$ and outcome $\texttt{y}$, there is no statistically significant difference in distributions of outcomes between the groups \citep[Section 1.3]{peters2014causal}. This suggests that CGMs are objects that people can possibly falsify via experiments. This also echoes the idea that interventional distributions corresponds to outputs of randomized controlled trials. In contrast, the literature agrees that SCMs are generally not falsifiable. These models entail counterfactual distributions which, as aforementioned, imply untestable statements about singular causes. Thereby, it may seem that CGMs are falsifiable whereas SCMs are not. As explained below, the classification is actually more complicated.

Not all variables are arguably eligible to randomization. This includes variables for which we cannot conceive a randomization process. As examples, the causality literature often mentions immutable characteristics like sex or race, or more ambiguous properties like tolerance. Therefore, CGMs that contain arrows starting from such variables are frequently deemed not falsifiable.\footnote{This relates to the long-standing debate about manipulable causes. We refer to \citep{winship1999estimation,freedman2004graphical,berk2004regression,greiner2011causal,vanderweele2014causal,glymour2017evaluating,pearl2018does,pearl2019do} for more details.} This notably means that, even though SCMs can (almost) never be falsified, the objective of this paper cannot be summarized as a study of nonfalsifiable causation.

Interestingly, the nonfalsifiability of SCMs differs from the one of CGMs. For illustration, suppose that people agree on how to randomize race, or that analysts are able to design experiments to randomly assign this characteristic. Even in this fictional scenario, one would never be able to assess the outcome of a \emph{same} individual had their race changed. This highlights that the nonidentififiability of singular causes holds independently of the feasibility of the experiment, contrarily to general causes. As such, counterfactual assumptions seem metaphysical at a even higher level than conceiving interventions on nonmanipulable variables.

All in all, the key point to keep in mind is that regardless of the nature of the given CGM, the counterfactual assumptions that one places over it (for instance with an SCM) are not falsifiable. Therefore, they rest on an arbitrary modeling choice which must be made fully transparent. In what follows, we propose an alternative approach to SCMs for formalizing et choosing these assumptions.

\section{Counterfactual models}\label{sec:ctf_models}

Basically, a model to reason counterfactually is something built on top of a graphical model (or any interventional model) to describe cross-world effects. As explained above, this can be achieved via a structural model. But all in all, what truly matters is the joint probability distributions one can place over interventional marginals. We propose an alternative way of representing counterfactual assumptions based on this remark.

\subsection{Definition and equivalence to structural causal models}\label{sec:def_ctf}

For starters, we introduce the problem in a simple two-marginal case. Let $\Phi_1, \Phi_2$ be two actions, and $P^{\Phi_1}, P^{\Phi_2}$ be the resulting interventional distributions of a graphical model $\C$. We ask two questions. First, what are the remaining degrees of freedom (as permitted by SCMs) for the counterfactual couplings between $P^{\Phi_1}$ and $P^{\Phi_2}$? Second, can we find an intelligible representation of these couplings? Naturally, we extend these questions to any family of actions $(\Phi_w)^W_{w=1}$, where $W \geq 2$, leading to joint probability distributions between possibly many marginals.

To construct a joint probability that respects the causal ordering from $\C$, the idea is to firstly construct a joint probability distribution for each components from the Markov factorization. Concretely, for every $i \in [d]$, one designs a joint probability distribution over $P_{i | \pa(i)}(\bullet | \V_{\pa(i)})$. This is formalized by the definition below.

\begin{definition}[One-step-ahead counterfactual distribution]
Let $\V := \times^d_{i=1} \V_i \subseteq \R^{d}$ be a Cartesian product of Borel subsets of $\R$ and $\C := (\V,\G,\K) \in \CGM_{\V}$. For any $i \in [d]$, a \emph{one-step-ahead counterfactual distribution} adapted to $K_i$ is an $S^{(i)} \in \Jointc{\V_{\pa(i)}}{\V_i}{(K_i(\bullet | \V_{\pa(i)}))}$.
\end{definition}
The term \say{one-step-ahead} was originally proposed by \cite{richardson2013single}. It qualifies the fact that such a distribution describes how the variable $i$ reacts to interventions on all the variables positioned one step ahead in the causal ordering, namely, $\pa(i)$.

Note that one-steap-ahead counterfactual distributions are simply \emph{composable} process measures with marginals induced by a \emph{probability kernel}. Such a composable distribution always exists as a consequence of the existence of SCMs fitting a given observational distribution (\cref{prop:scm_exists}). We underline that composability and measurability issues may arise when the marginals are not derived from a probability kernel (see \cref{rem:unnormalize}).

\begin{figure}[t]
\centering
\includegraphics[page=2, scale=0.7]{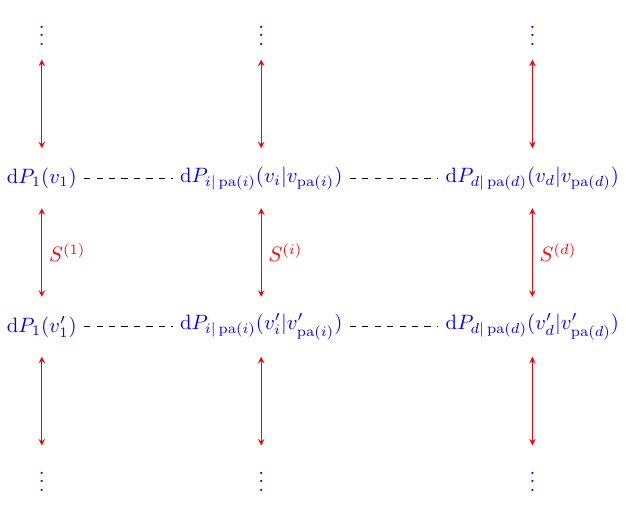}
\caption{Principle of a counterfactual model. The observational distribution $P$ is factorized (horizontally) relatively to the graph $\G$. Each factor $P_{i | \pa(i)}$ is then joined (vertically) across every possible realization by a process measure $S^{(i)}$. The color \textcolor{red}{red} highlights counterfactual information (what one may choose); the color \textcolor{blue}{blue} highlights interventional information (what is given).}
\label{fig:coupling_process}
\end{figure}

Stacking one-step-ahead counterfactual distributions along the topological order defines a \emph{counterfactual model}: an object built on top of a graphical model, from which one can derive counterfactual distributions compatible with this graphical model. \Cref{fig:coupling_process} illustrates the construction. Note that composability is essential.

\begin{definition}[Counterfactual model]\label{def:ctfm}
Let $\V := \times^d_{i=1} \V_i \subseteq \R^{d}$ be a Cartesian product of Borel subsets of $\R$ and $\C := (\V,\G,\K) \in \CGM_{\V}$. A \emph{counterfactual model} adapted to $\C$ is a pair $\A := (\C,\S)$ where $\S := (S^{(i)})^d_{i=1}$ is a collection of one-step-ahead counterfactual distributions such that $S^{(i)}$ is adapted to $K_i$ for every $i \in [d]$. We denote by $\CR_{\V,\C}$ the set of counterfactual models adapted to $\C \in \CGM_{\V}$ and define for every $\A \in \CR_{\V,\C}$:
\begin{enumerate}
    \item $\Obs{A}(\A) := \Obs{C}(\C)$, called the \emph{observational distribution} of $\A$;
    \item for every $\Phi \in \I(\V)$, $\Int{A}(\A, \Phi) := \Int{C}(\C, \Phi)$, called the \emph{interventional distribution} of $\A$ relatively to $\Phi$;
    \item for every $(\Phi_w)^W_{w=1} \in \I(\V)^W$, the process measure $Q \in \Q(\{\Phi_w\}^W_{w=1}, \V)$ given by
    \[
    \mathrm{d}Q \left((v^{(w)})^W_{w=1} \right) = \prod^d_{i=1} \mathrm{d} \left( \left( \underset{w \in [W]}{\times} \phi_{w,i} \right) \# S^{(i)}_{v^{(1)}_{\pa(i)}, \ldots, v^{(W)}_{\pa(i)}} \right) \left((v^{(w)}_i)^W_{w=1} \right),
    \]
    denoted by $\Ctf{A}(\A, (\Phi_w)^W_{w=1})$, and called the \emph{counterfactual distribution} of $\A$ relatively to $(\Phi_w)^W_{w=1}$.
\end{enumerate}
Moreover, we say that $Q \in \Q(\{\Phi_w\}^W_{w=1}, \V)$ is a \emph{counterfactual distribution adapted to $\C \in \CGM_{\V}$ relatively to $(\Phi_w)^W_{w=1}$} if there exists $\A \in \CR_{\V,\C}$ such that $Q = \Ctf{A}(\A, (\Phi_w)^W_{w=1})$.
\end{definition}

\begin{figure}
    \centering
    \includegraphics[page=3,scale=0.8]{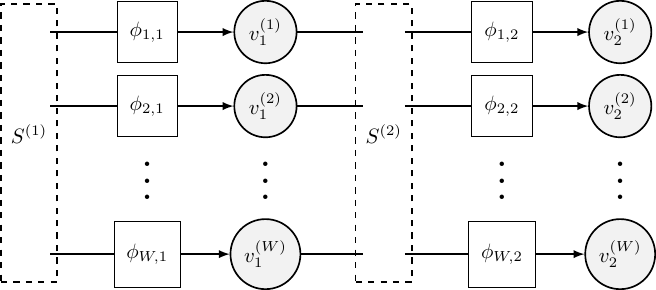}
    \caption{Sampling process from a counterfactual distribution with a counterfactual model. It represents the generation of a single sample $(v^{(w)}_i)_{ w \in [W], i \in [d]}$ from $\Ctf{A}(\A,(\Phi_w)^W_{w=1})$ highlighted in gray. For simplicity, the figure addresses the graph $1 \rightarrow 2$. One can extend the principle to any graph. It resembles a neural network with size $W$ and depth $d$, where each input corresponds to a parallel world indexed by $w \in [W]$ and each layer corresponds to a variable indexed by $i \in [d]$. The counterfactual process measure $S^{(i)}$ operates as a random activation function that introduces cross dependencies over its inputs, while the intervention $\phi_{w,i}$ deterministically activates input $i$ in world $w$.}
    \label{fig:sampling}
\end{figure}

\Cref{fig:sampling} illustrates the generative process described in the third item of \cref{def:ctfm}.\footnote{The TikZ code is inspired from \citep{love2024nntikz}.} It shares similarities with the one of deep Gaussian processes \citep{damianou2013deep}, where the outputs of a process are the inputs of the next one. In our case, each one-step-ahead counterfactual processes is composed along the topological order. As for structural causal models, counterfactual distributions of counterfactual models have interventional marginals.

\begin{proposition}[Interventional marginals]\label{prop:int_margins}
Let $\V := \times^d_{i=1} \V_i \subseteq \R^{d}$ be a Cartesian product of Borel subsets of $\R$, $\C \in \CGM_{\V}$, and $\A \in \CR_{\V,\C}$. Then, $\Ctf{A}(\A, (\Phi_w)^W_{w=1}) = \Q_{\text{joint}}(\{\Phi_w\}^W_{w=1}, \V,(\Int{A}(\A,\Phi_w))^W_{w=1})$.
\end{proposition}

The following proposition establishes the first part of the equivalence between SCMs and counterfactuals models: every Markovian structural causal model produces a counterfactual model with the same counterfactual distributions.
\begin{proposition}[Entailed counterfactual model]\label{prop:entailed_a}
Let $\V := \times^d_{i=1} \V_i \subseteq \R^{d}$ be a Cartesian product of Borel subsets of $\R$, $\C \in \CGM_{\V}$, and $\M := (\V,\G,\U, P_0, f) \in \SCM_{\V,\C}$. For every $i \in [d]$, write $S^{(i)} \in \Q^*(\V_{\pa(i)},\V_i)$ for the process measure $\left(f_i(v_{\pa(i)},\bullet) \right)_{v_{\pa(i)} \in \V_{\pa(i)}} \# P_{0,i}$,\footnote{That is, the law of the random process $\left(f_i(v_{\pa(i)},U_i)\right)_{v_{\pa(i)} \in \V_{\pa(i)}}$, which is measurable since $f_i$ is measurable.} called the \emph{entailed one-step-ahead counterfactual distribution} of $\M$ at $i$. Then, $\A_\M := (\C,\S)$, where $\S := (S^{(i)})^d_{i=1}$, is a counterfactual model adapted to $\C$, called the \emph{entailed counterfactual model} or \emph{canonical representation} of $\M$. Moreover, for every $W \geq 1$ and every $(\Phi_w)^W_{w=1} \in \I(\V)^W$, $\Ctf{M}(\M,(\Phi_w)^W_{w=1}) = \Ctf{A}(\A_\M,(\Phi_w)^W_{w=1})$.
\end{proposition}

The entailed one-step-ahead counterfactual distribution at $i$ mathematically corresponds to the distribution of the random process $\left( V^{\doint(\pa(i),v_{\pa(i)})}_i \right)_{v_{\pa(i) \in \V_{\pa(i)}}}$. Note that $S^{(i)}$ places a probability distribution on the set of functions ${\V_i}^{\V_{\pa(i)}}$, which inserts randomness to the causal links $\pa(i) \rightarrow i$ alternatively to how the exogenous noise $U_i$ inserts randomness in a structural equation. This changes of viewpoint aligns with the notion of \emph{canonical representation} of a structural equation proposed by \cite{peters2017elements}. Since we presented a counterfactual model as a primitive, not necessarily the derivative of some SCM, we did not use the expression \emph{canonical representation} from the start.

Conversely to \cref{prop:entailed_a}, every counterfactual model can be represented by a structural causal model. This notably characterizes, given $\C \in \CGM_{\V}$, which joint probability distributions are counterfactual distributions from a model in $\SCM_{\V,\C}$: those produced by the models in $\CR_{\V,\C}$. In other words, admissible counterfactual distributions essentially stem from \emph{composable} unitary process distributions $(S^{(i)})^d_{i=1}$ stacked along $\G$ so that they meet a \emph{joint Markov condition}.  

\begin{theorem}[Structural representation]\label{thm:representation}
Let $\V := \times^d_{i=1} \V_i \subseteq \R^{d}$ be a Cartesian product of Borel subsets of $\R$. For every $\C \in \CGM_{\V}$ and $\A \in \CR_{\V,\C}$, there exists $\M \in \SCM_{\V,\C}$ such that $\A = \A_\M$.
\end{theorem}

The proof is similar to how scholars demonstrate that a potential-outcome model corresponds to an SCM (see \cref{sec:po}). The idea is that a stochastic process $(Y_x)_{x \in \X}$ induces the structural mechanism $f_{\tty}(x,\omega_{\tty}) := Y_x(\omega_{\tty})$. The fact that $f_{\tty}$ must be measurable implies that $(Y_x)_{x \in \X}$ needs to be measurable, hence why we suppose one-steap-ahead counterfactual distributions to be composable. This shows that the composability of a counterfactual process distribution mirrors the measurability of a structural-assignment function.

To sum up, counterfactual models and structural causal models are equivalent for reasoning along the causality ladder. An interest of counterfactual models comes from their conciseness, as several distinct structural causal models may entail the same counterfactual model. In particular, changing $(f_i,P_{0,i})$ by $(f_i \circ (\Id \times \varphi_i), \varphi^{-1} \# P_{0,i})$, where $\varphi_i$ is an invertible noise reparametrization, does not modify the counterfactuals.\footnote{This remark is the foundation of the notion of \emph{exogenous isomorphism equivalence} between SCMs, which implies counterfactual equivalence \citep{chen2025exogenous}.}

\begin{proposition}[Conciseness of canonical representations]\label{prop:concise}
Let $\V := \times^d_{i=1} \V_i \subseteq \R^{d}$ be a Cartesian product of Borel subsets of $\R$. For every $\C \in \CGM_{\V}$, there exist $\M_1, \M_2 \in \SCM_{\V,\C}$ such that $\M_1 \neq \M_2$ and $\A_{\M_1} = \A_{\M_2} \in \CR_{\V,\C}$.
\end{proposition}

As such, canonical representations rule out irrelevant degrees of freedom. This property was already noted by \cite{peters2017elements} for single structural equations. More precisely, we prove that counterfactual models characterize the counterfactual equivalence classes of SCMs. In a sense, this justifies the expression \say{canonical representation}, as differences between causal models beyond the counterfactual layer are irrelevant.

\begin{theorem}[Characterization of counterfactual equivalence]\label{thm:ctf_eq}
Let $\V := \times^d_{i=1} \V_i \subseteq \R^{d}$ be a Cartesian product of Borel subsets of $\R$. Every $\M_1, \M_2 \in \SCM_{\V}$ are counterfactually equivalent if and only if $\A_{\M_1} = \A_{\M_2}$. 
\end{theorem}
Critically, \cref{thm:ctf_eq} and \cref{prop:concise} imply that the counterfactual layer identifies a counterfactual model but does not identify an SCM.

Beyond their conciseness, canonical representations clearly separate the observational, interventional, and counterfactual components: by changing $\S$ while preserving $\C$, one can modify their counterfactual conception without touching the interventional assumptions. In contrast, changing $f$ or $P_0$ in an $\M \in \SCM_{\V,\C}$ generally breaks the compatibility to $\C$.

\subsection{Normalizations of counterfactual models}\label{sec:normalizations}

We provided a characterization of the counterfactual distributions compatible with a given causal graphical model (which encodes the complete observational and interventional knowledge) via a canonical representation (called a counterfactual model) based on process measures. We explained how this representation could be more straightforward than structural causal models to formalize counterfactual assumptions specifically. To further illustrate this point and to tackle practical considerations, we now introduce \emph{normalizations of counterfactual models}.

This concept is motivated by the following remark. Given $\C := (\V,\G,\K) \in \CGM_{\V}$, when one selects a one-step-ahead counterfactual process $S^{(i)}$ among $\Jointc{\V_{\pa(i)}}{\V_i}{K_i(\bullet|\V_{\pa(i)})}$, what truly matters at the counterfactual level are the cross dependencies that $S^{(i)}$ places over the preassigned marginals $K_i(\bullet|\V_{\pa(i)})$. Consider, for instance, the process measure that preserves the ranks across all marginals: the counterfactual assumption lies in the rank-preserving property---not in the marginals. This underlines the need for a framework allowing to choose the cross dependencies independently of the process's marginals. Borrowing ideas from the theory of copulas \citep{sklar1959fonctions} and warped random processes \citep{snelson2003warped, wilson2010copula}, we propose to define the cross dependencies in a latent space through a normalized process measure with marginals equal to $\Norm$, and then to transport the dependencies over the marginals $K_i(\bullet|\V_{\pa(i)})$. This proposition has two interests: a conceptual one and a practical one. First, it fully achieves the disentanglement of counterfactual assumptions from the observational and interventional knowledge. Moreover, thinking in a normalized space notably permits analysts to intelligibly formalize their counterfactual beliefs. Second, it provides a practical framework to implement such beliefs.

\subsubsection{Definition}

Normalizations are defined at the one-variable level, namely, at the transition $\pa(i) \to i$ for each $i \in [d]$. Similar to the kernels of causal graphical models, the mechanisms of structural causal models, and the one-step-ahead distributions of counterfactual models, one gets a normalization for the whole graph $\G$ by stacking unitary normalizations. We underline that such an iterative construction heavily relies on the Markov assumption.

For a given $i \in [d]$, the first step consists in transporting the marginals $(\Norm)_{v_{\pa(i)} \in \V_{\pa(i)}}$ onto $K_i(\bullet|\V_{\pa(i)})$. We do so via the following diagonal non-decreasing map.
\begin{proposition}[Unnormalizing maps]\label{prop:unnormalize}
Let $\V := \times^d_{i=1} \V_i \subseteq \R^{d}$ be a Cartesian product of Borel subsets of $\R$ and $\C := (\V,\G,\K) \in \CGM_{\V}$. For every $i \in [d]$ and every $v_{\pa(i)} \in \V_{\pa(i)}$, denote by $\psi^{\C}_i(\bullet | v_{\pa(i)})$ the non-decreasing transport map from $\Norm$ to $K_i(\bullet | v_{\pa(i)})$. Additionally, consider the function $\Psi^{\C}_i : \R^{\V_{\pa(i)}} \to \V_i^{\V_{\pa(i)}}$ given by
$\Psi^{\C}_i := \left( \underset{v_{\pa(i)} \in \V_{\pa(i)}}{\times} \psi^{\C}_i(\bullet|v_{\pa(i)}) \right)$. Then,
\begin{enumerate}
    \item $\psi^{\C}_i(\bullet|\bullet)$ is measurable;
    \item $\Psi^{\C}_i(\bullet)$ is measurable.
\end{enumerate}
\end{proposition}
We emphasize that the measurability of $\psi^C_i$ strongly rests on the fact that the collection of marginals come from a probability kernel $K_i$. For an arbitrary collection of marginals, $\psi^C_i$ may not be measurable in its second argument (see \cref{rem:unnormalize}).

Note that the maps $\psi^{\C}_i$ and $\Psi^{\C}_i$ are uniquely determined by $\C$, and as such do not capture counterfactual knowledge. The second steps consists in placing cross-world dependencies in the normalized space. Remind that $\Normac{\X} := \Jointc{\X}{\R}{(\Norm)_{x \in \X}}$.
\begin{definition}[Normalizations of counterfactual models]\label{def:norm}
Let $\V := \times^d_{i=1} \V_i \subseteq \R^{d}$ be a Cartesian product of Borel subsets of $\R$ and $\C := (\V,\G,\K) \in \CGM_{\V}$. Then, define the two following objects:
\begin{enumerate}
    \item for every $i \in [d]$ and every one-step-ahead counterfactual distribution $S^{(i)}$ adapted to $K_i$, a \emph{normalization} of $S^{(i)}$ is any process distribution $N^{(i)} \in \Normac{\V_{\pa(i)}}$ such that $\Psi^{\C}_i \# N^{(i)} = S^{(i)}$;
    \item for every $\A := (\C,\S) \in \CR_\C$, a \emph{normalization} of $\A$ is any collection $\mathcal{N} := (N^{(i)})^d_{i=1}$ such that $N^{(i)}$ is a normalization of $S^{(i)}$ for every $i \in [d]$.
\end{enumerate}
\end{definition}

\begin{figure}[t]
\centering
\includegraphics[page=4,scale=0.8]{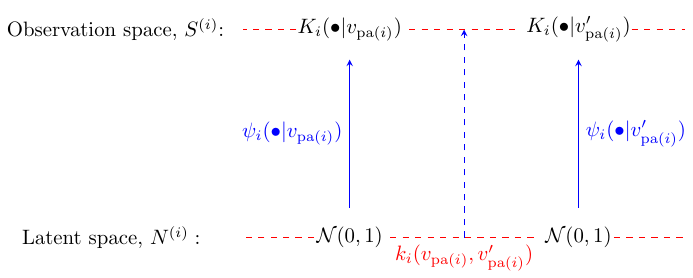}

\caption{Generating a joint probability distribution between given marginals through a normalization. The idea is to specify the dependencies in a latent Gaussian space and then to transport them monotonically in the observation space. The color \textcolor{red}{red} highlights the cross-marginal dependencies (that is, the counterfactual conception); the color \textcolor{blue}{blue} highlights the transport between spaces (which is not counterfactual). Note that all the represented distributions are univariate.}
\label{fig:generating_process}
\end{figure}

The \say{disentangled} expression $S^{(i)} := \Psi^{\C}_i \# N^{(i)}$ separates $\Psi^{\C}_i$, an interventional object fully determined by $\C$, from $N^{(i)}$, which captures only the counterfactual dependencies of the joint probability distribution $S^{(i)}$. The idea in practice is to \emph{choose} the counterfactual conception via a normalization $N^{(i)}$---we remind that there is no empirical basis to falsify counterfactuals---and then to generate $S^{(i)}$ via $\Psi^{\C}_i$ (which is not arbitrary) in order to produce counterfactual models without altering interventional and observational knowledge. We also emphasize that even though the marginals of $N^{(i)}$ are Gaussian, it need not be a Gaussian-process measure.

We point out that, while $\Psi^\C_i$ is always measurable, and thereby $\Psi^{\C}_i \# N^{(i)}$ produces a well-defined process distribution, it may not be obvious that $\Psi^{\C}_i \# N^{(i)}$ is composable if $N^{(i)}$ is composable. Composability of the unnormalized process follows from the joint measurability of $\psi^\C_i$. This allows to construct counterfactual models via the parametrization $S^{(i)} := \Psi^{\C}_i \# N^{(i)}$.
\begin{lemma}[Unnormalization preserves composability]\label{lem:norm_comp}
Let $\X,\Y,\Y'$ be Borel spaces and suppose that $\psi(\bullet|\bullet) : \Y \times \X \to \Y'$ is measurable. If $Q \in \Q^*(\X,\Y)$, then $\left( \underset{x \in \X}{\times} \psi(\bullet|x) \right) \# Q \in \Q^*(\X,\Y')$.
\end{lemma}

The next proposition establishes conditions for the existence and uniqueness of normalizations. It rests on the fundamental result on copulas. Sklar's theorem proves that the cross dependencies of a joint probability distributions can always be chosen intrinsically by normalizing the marginals to $\Unif{[0,1]}$. An invertible reparameterization readily gives the same property with marginals normalized to $\Norm$.

\begin{proposition}[On existence and uniqueness]\label{prop:norm}
Let $\V := \times^d_{i=1} \V_i \subseteq \R^{d}$ be a Cartesian product of Borel subsets of $\R$. For every $\C := (\V,\G,\K) \in \CGM_{\V}$ and every $i \in [d]$, define the mapping
\[
\Upsilon^{\C}_i : \Q(\V_{\pa(i)},\R) \to \Q(\V_{\pa(i)},\V_i), Q \mapsto \Psi^{\C}_i \# Q.
\]
Then, for every $i \in [d]$,
\begin{enumerate}
    \item \begin{multline*}
        \Upsilon^{\C}_i\left( \Normac{\V_{\pa(i)}} \right) \subseteq \Jointc{\V_{\pa(i)}}{\V_i}{K_i(\bullet|\V_{\pa(i)})} \\ \subseteq \Joint{\V_{\pa(i)}}{\V_i}{K_i(\bullet|\V_{\pa(i)})} = \Upsilon^{\C}_i\left( \Norma{\V_{\pa(i)}} \right);
    \end{multline*}
    \item if $\psi^{\C}_i(\bullet|v_{\pa(i)})$ is injective for every $v_{\pa(i)} \in \V_{\pa(i)}$, then $\Upsilon^{\C}_i$ is bijective from $\Norma{\V_{\pa(i)}}$ to $\Joint{\V_{\pa(i)}}{\V_i}{K_i(\bullet|\V_{\pa(i)})}$ and from $\Normac{\V_{\pa(i)}}$ to $\Jointc{\V_{\pa(i)}}{\V_i}{K_i(\bullet|\V_{\pa(i)})}$.
\end{enumerate}
\end{proposition}

\Cref{prop:norm} ensures existence and uniqueness of the normalization of every one-step-ahead counterfactual distribution $S^{(i)}$ as soon as $\psi^{\C}_i(\bullet|v_{\pa(i)})$ is injective for every $v_{\pa(i)} \in \V_{\pa(i)}$, which is notably the case when $K_i(\bullet|v_{\pa(i)})$ is absolutely continuous with respect to the Lebesgue measure for every $v_{\pa(i)} \in \V_{\pa(i)}$. If injectivity does not hold, then a proper normalization may not exist, as it could fail to be composable.\footnote{More precisely, while $\Psi^\C_i \#$ can actually transform a non-composable measure into a composable one, whether a composable measure always admits a composable antecedent by $\Psi^\C_i \#$ is an open question.} 

Such a push-forward reparameterization, $S^{(i)} := \Psi^{\C}_i \# N^{(i)}$, can be also be obtained with unconstrained (unnecessarily monotonic) transport maps. Nevertheless, choosing $\Psi_i^{\C}$ composed of non-decreasing transport maps is important for two reasons. First, this guarantees identifiability: $\Psi^{\C}_i$ is uniquely determined by $K_i$, as $\psi^{\C}_i(\bullet|v_{\pa(i)})$ is the unique non-decreasing transport map from $\Norm$ to $K_i(\bullet|v_{\pa(i)})$ for every $v_{\pa(i)} \in \V_{\pa(i)}$. Second, this permits a natural translation from dependencies in the latent space to dependencies in the observation space: the process $S^{(i)}$ inherits by monotonicity the dependencies in $N^{(i)}$. For instance, if $N^{(i)}$ preserves the ranks across all marginals, then so does $S^{(i)}$. The next section explores in details the questions of translating intelligible counterfactual conceptions into normalizations and of how the properties of a normalization reverberates onto the one-step-ahead counterfactual process.

\begin{figure}
    \centering
    \includegraphics[page=5,width=0.8\linewidth]{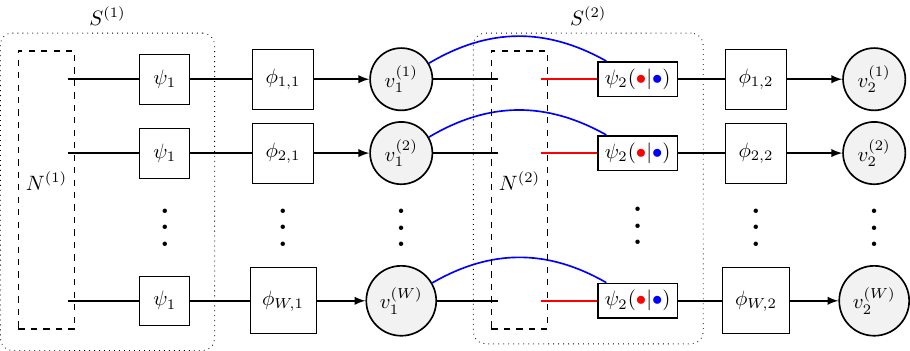}
    \caption{Sampling process from a counterfactual distribution with a normalized counterfactual model. The only inference step before this prediction step is the computation of $\psi_1$ and $\psi_2$.}
    \label{fig:sampling_normalized}
\end{figure}

\begin{remark}[Random-variable formalism]\label{rem:variables}
For readability, we propose a random-variable notation of normalized counterfactual models, in line with solutions of SCMs. For simplicity, we focus on one variable and its parents. Suppose that $\mathtt{x} := \{\mathtt{x}_1, \ldots, \mathtt{x}_d \} = \pa(\mathtt{y})$ in some CGM $\C \in \CGM_{\V}$. Then, consider $(X,Y)$, the projection on $\{\ttx,\tty\}$ of a solution of some SCM $\M \in \SCM_{\V,\C}$. The assignment of $Y$ can be written as
\begin{equation}\label{eq:structural_assigment}
    Y = f_{\tty}(X,U_{\tty}),
\end{equation}
where $U_{\tty} \independent X$. The one-steap-ahead potential outcomes, namely, the $\tty$ outputs of $(\doint(\ttx,x))_{x \in \X}$, are given by
\[
    Y_x := f_{\tty}(x,U_{\tty})
\]
for every $x \in \X$, and entail the random process $(Y_x)_{x \in \X} \sim S^{(\tty)}$. Observe that they meet $Y = Y_X$, which is referred to as \emph{consistency}.

Moreover, let $\psi^\C_{\tty}$ be defined as in \cref{def:norm}. We recall that this function is fully determined by $\C$ only. According to \cref{thm:ctf_eq} and assuming the existence of a normalization $N^{(\tty)}$, there exists a measurable random process $E^{(\tty)} := (E^{(\tty)}_x)_{x \in \X} \sim N^{(\tty)}$ such that $E^{(\tty)} \independent X$ and
\[
    Y_x = \psi^\C_{\tty}\left( E^{(\tty)}_x | x \right).
\]
Therefore, consistency leads to
\begin{equation}\label{eq:normalization_assignment}
    Y = \psi^\C_{\tty}\left( E^{(\tty)}_X | X \right).
\end{equation}

The normalization \cref{eq:normalization_assignment} provides a similar assignment than the SCM \cref{eq:structural_assigment}: $Y$ is causally determined by $X$ and an independent source of randomness. Note that $\law{E^{(\tty)}_X | X=x} = \law{E^{(\tty)}_x}$ since $E^{(\tty)} \independent X$, and hence $\law{E^{(\tty)}_X | X=x} = \Norm$, which means that $E^{(\ttx)}_X \independent X$. Nonetheless, $E^{(\tty)}_X$ behaves differently from $U_{\tty}$ under interventions. Letting $\Phi$ be an intervention transformation that only acts on $\mathtt{x}$ through a product map $\phi_{\ttx} := \times^d_{i=1} \phi_{\ttx_i}$, one has
\[
    Y^{\Phi} := f_{\tty}(\phi_{\ttx}(X),U_{\tty}) = \psi^\C_{\tty}\left( E^{(\tty)}_{\phi_{\ttx}(X)} | \phi_{\ttx}(X) \right).
\]
In this sense, and as later detailed in \cref{rem:noises}, the noises $U_{\tty}$ and $E^{(\tty)}_X$ play radically different roles while defining the same factual and potential outcomes. Furthermore, note that $E^{(\tty)}_X \sim \Norm$ and $(E^{(\tty)}_X,E^{(\tty)}_{\phi_{\ttx}(X)}) \sim N^{(\tty)}_{X,\phi_{\ttx}(X)}$. As such, the index of $(E^{(\tty)}_x)_{x \in \X}$ does not provide any information at the marginal level but determines the joint probability distribution of $(Y,Y^{\Phi})$. Finally, we emphasize that changing the counterfactuals while preserving $\C$ in the normalized assignment only requires changing the normalization $N^{(\tty)}$---the function $\psi^\C_{\tty}$ remains fixed. We will use this formalism for counterfactual models in \cref{sec:exp}.
\end{remark}

\begin{remark}[On the measurability of unnormalization]\label{rem:unnormalize}
As aforementioned, the measurability of $\psi^{\C}_i$ comes from the fact that the target marginals stem from a probability kernel $K_i$. To illustrate this point, let $\X$ be an uncountable Borel space and let $(P^{(x)})_{x \in \X}$ be a family from $\Prob{\R}$. Crucially, $(P^{(x)})_{x \in \X}$ is not necessarily induced by a probability kernel in this configuration.\footnote{If $\X$ is countable, then $(P^{(x)})_{x \in \X}$ automatically defines a probability kernel from $\X$ to $\R$.} For every $x \in \X$, define $\psi(\bullet|x)$ as the non-decreasing transport map from $\Norm$ to $P^{(x)}$. Now, let $m : \X \to \R$ be non-measurable and take $P^{(x)} := \mathcal{N}(m(x),1)$. By uniqueness, $\psi(e|x) = e + m(x)$, which is not measurable.
\end{remark}

\subsubsection{Formalization}

To illustrate how to concretely formalize counterfactual conceptions, we introduce a specific class of normalizations: the centered Gaussian-process measures.

\begin{definition}[Gaussian normalizations]
Let $\X$ be a set. For every positive semidefinite kernel function $k : \X \times \X \to \R$ such that $k(x,x)=1$ for all $x \in \X$, we write $N^k$ for the law of a Gaussian process with mean function $x \mapsto 0$ and covariance function $k$ (regardless of the index and target spaces, slightly abusing notation). In particular, $N^k \in \Norma{\X}$.

Let $\X$ be a Borel space. If $N^k$ is composable (that is, if $N^k \in \Normac{\X}$) then we call it the \emph{Gaussian normalization} associated to $k$. Additionally, we define $N^\uparrow := N^k \in \Normac{\X}$ for $k$ given by $k(x,x')=1$ for every $x,x' \in \X$, and call it the \emph{comonotonic normalization}. This corresponds to the law of a random process $(Y_x)_{x \in \X}$ where $Y_x = Y_{x'} \sim \Norm$ for every $x,x' \in \X$.
\end{definition}

\begin{figure}
    \def\width{0.4}
    \def\sep{1.3cm}
    \centering
    \begin{subfigure}[b]{\width \textwidth}
        \scalebox{0.7}{\includegraphics[page=6]{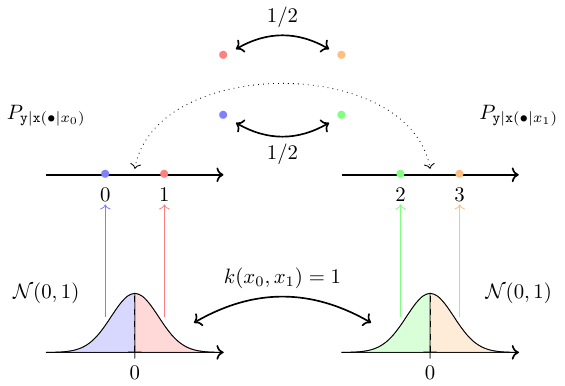}}
    \caption{Comonotonic coupling}
    \end{subfigure}
    \hspace{\sep}
    \begin{subfigure}[b]{\width \textwidth}
        \scalebox{0.7}{\includegraphics[page=7]{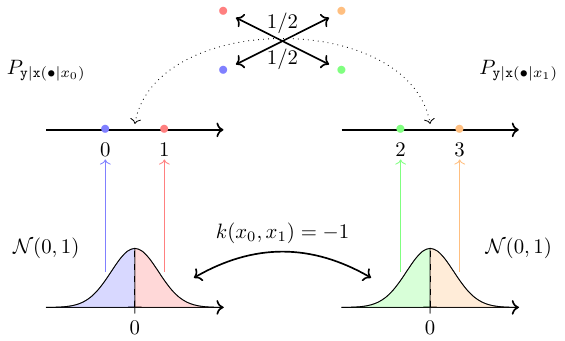}}
    \caption{Countermonotonic coupling}
    \end{subfigure}
    \begin{subfigure}[b]{\width \textwidth}
        \scalebox{0.7}{\includegraphics[page=8]{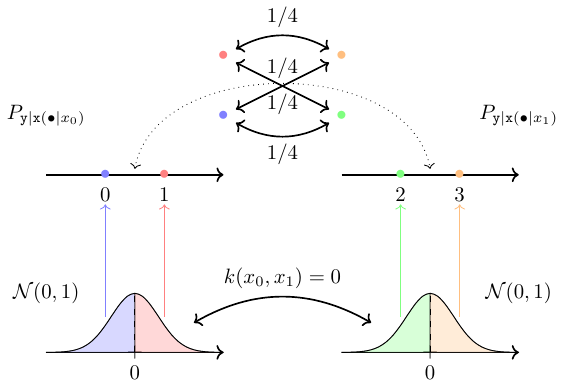}}
    \caption{Independent coupling}
    \end{subfigure}
    \hspace{\sep}
    \begin{subfigure}[b]{\width \textwidth}
        \scalebox{0.7}{\includegraphics[page=9]{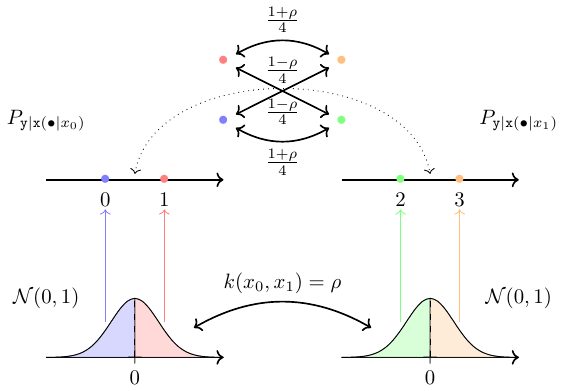}}
    \caption{General case ($-1 \leq \rho \leq 1$)}
    \label{fig:general_case}
    \end{subfigure}
    \caption{Four couplings produced with latent Gaussian couplings. The marginals are $P_{\texttt{y}|\texttt{x}(\bullet|x_0)} := \operatorname{Unif}(\{0,1\})$ and $P_{\texttt{y}|\texttt{x}(\bullet|x_1)} := \operatorname{Unif}(\{2,3\})$. The transform $\psi_{\texttt{y}}$ is represented by the colored arrows: it is the same in all figures. The only parameter that changes between the figures is the covariance $k$, which entails different couplings.}
    \label{fig:latent_feature_1}
\end{figure}

Naturally, considering such process measures as normalizations restricts the set of attainable one-step-ahead counterfactual distributions, since there can be non-Gaussian cross-dependencies between Gaussian marginals. Nevertheless, this provides a rich, intelligible, and computable class of dependencies across such marginals: one can in particular choose if counterfactuals are deterministic or stochastic with a control over the variance. This is why we propose this class to encode basic counterfactual conceptions, as well as to compute normalizations in practice (later in \cref{sec:exp}). Note that, in specific situations, the link between Gaussian normalizations and process measures with preassigned marginals is bijective (like in \cref{fig:general_case}).

In this Gaussian framework, the counterfactual assumptions at the one-step-ahead level (that is the level of $\pa(i) \rightarrow i$) are fully encoded by the covariance function $k_i$ (more precisely its values beyond the diagonal) of $N^{(i)} := N^{k_i}$. The choice of $k_i$ allows to control essentially two properties of the counterfactual conception in the latent space: whether the counterfactuals are \emph{deterministic} or \emph{stochastic} and whether they are \emph{comonotonic} or \emph{countermonotonic}. Concretely, if $\abs{k_i(v_{\pa(i)}, v'_{\pa(i)})} = 1$ then the latent coupling $N^{(i)}_{v_{\pa(i)},v'_{\pa(i)}}$ is deterministic, whereas if $\abs{k_i(v_{\pa(i)}, v'_{\pa(i)})} < 1$ then it is stochastic. Moreover, if $k_i(v_{\pa(i)}, v'_{\pa(i)}) = 1$ then the latent coupling is comonotonic (it notably coincides with the unique non-decreasing transport map), while if $k_i(v_{\pa(i)}, v'_{\pa(i)}) = -1$ then the coupling is countermonotonic. To sum-up, the sign and intensity of $k_i(v_{\pa(i)}, v'_{\pa(i)})$ determine the properties of the counterfactual coupling \emph{in the latent space}. We underline that globally, when considering all pairs $(v_{\pa(i)}, v'_{\pa(i)})$, $k_i$ must remain a \emph{positive semidefinite} kernel that ensures \emph{composability} of the Gaussian process measure.

\begin{figure}
    \centering
    \begin{subfigure}[b]{0.4 \textwidth}
        \scalebox{0.7}{\includegraphics[page=10]{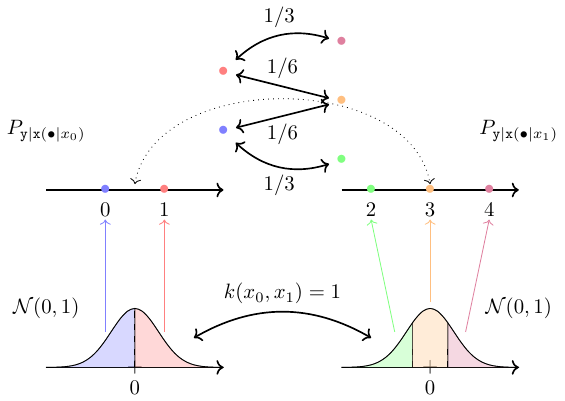}}
    \caption{Comonotonic coupling}
    \end{subfigure}
    \hspace{1.3cm}
    \begin{subfigure}[b]{0.4 \textwidth}
        \scalebox{0.7}{\includegraphics[page=11]{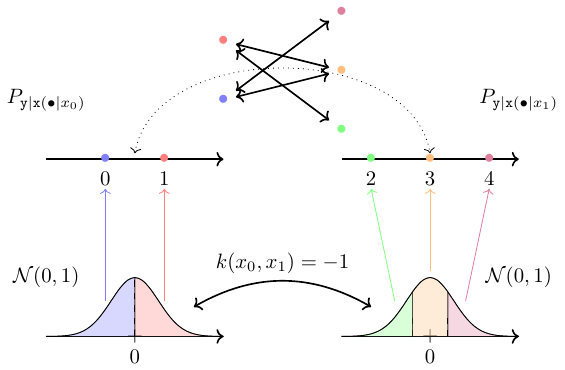}}
    \caption{Countermonotonic coupling}
    \end{subfigure}
    \caption{Two couplings between the marginals $P_{\texttt{y}|\texttt{x}(\bullet|x_0)} := \operatorname{Unif}(\{0,1\})$ and $P_{\texttt{y}|\texttt{x}(\bullet|x_1)} := \operatorname{Unif}(\{2,3,4\})$. Because there does not exist any deterministic coupling between these marginals, the counterfactual couplings are stochastic even when the latent Gaussian couplings are deterministic.}
    \label{fig:latent_feature_2}
\end{figure}

The monotonicity of the transformations $\psi^{\C}_i(\bullet|v_{\pa(i)})$ for every $v_{\pa(i)} \in \V_{\pa(i)}$ permits to transfer---as much as possible---these properties into the feature space. In \say{well-behaved situations}, notably when $K_i(\bullet|v_{\pa(i)})$ and $K_i(\bullet|v'_{\pa(i)})$ are both Lebesgue-absolutely continuous or both uniform over $n \geq 2$ points, then all the properties in the latent space get transported in the feature space. \Cref{fig:latent_feature_1} illustrates this behaviour. In particular, being deterministic in latent space means being deterministic in feature space. However, there exist cases where $K_i(\bullet|v_{\pa(i)})$ and $K_i(\bullet|v'_{\pa(i)})$ do not admit deterministic couplings. In such scenarios, a deterministic coupling in the latent space produces a stochastic coupling in the feature space, as represented by \cref{fig:latent_feature_2}. Another interest of fixing the counterfactual conception in the latent space is that it separates when stochasticity is choice (one chooses a stochastic normalization in the latent space) to when it is a constraint (imposed by the nature of the interventional marginals).

The comonotonic normalization $N^\uparrow$ corresponds to an intuitive and widely applied conception of counterfactuals, which states that counterfactually paired outcomes fit the same quantile of their respective distributions. Let us apply this conception to the cause-effect relationship between gender and height for illustration. According to the comonotonic conception, taller women are always associated to taller men. In particular, if Alice is a 170cm tall woman, then had she been a man she would have been 192cm tall: the two height values correspond to the 89th percentile of the women's and men's height distributions (of adults in the USA in 2015-2016 \citep{NHANES2015}). We say that $S^{(i)}$ is comonotonic if it admits $N^\uparrow$ as normalization (even if $S^{(i)}$ is stochastic).

\begin{remark}[About comonotonicity]
Most articles on singular causation employ deterministic comonotonic counterfactuals due to their intuitive rank-preserving nature and their relation to optimal transport maps \citep{plevcko2020fair,bothmann2023causal,machado2025sequential,balakrishnan2025conservative}. According to us, these are perfectly legitimate arguments to defend such a counterfactual conception. Nevertheless, one must always keep in mind that it remains a specific choice, it is not identifiable by observations and interventions.

We emphasize that the goal of this paper is not only to provide a practical framework to implement counterfactual reasoning (explained and illustrated in the following sections), but also to make fully transparent the various arbitrary choices that coexist at the singular-causation level. Perhaps, all practitioners will systematically prefer the comonotonic conception, rendering otiose the flexibility of the normalization strategy to test many counterfactual assumptions. But this will have at least sparked a crucial debate within the causality community, as counterfactual choices are rarely discussed.
\end{remark}

\begin{remark}[About countermonotonicity]
Can one define a \emph{countermonotonic} normalization $N^\downarrow$, corresponding to a quantile-reversing conception of counterfactuals? Perhaps surprisingly, not always. In the case where $\V_{\pa(i)}$ contains only two distinct values, normalizations are simply couplings. To create a countermonotonic coupling with normal marginals, one can define
\[
    N^\downarrow := \mathcal{N}\left( \begin{bmatrix} 0 \\ 0\end{bmatrix}, \begin{bmatrix}
        1 & -1 \\ -1 & 1
    \end{bmatrix} \right).
\]
From a counterfactual perspective, the $-1$ coefficients beyond the diagonal signifies that worlds n°1 and n°2 have a negative deterministic correlation. Critically, this construction does not extend to the case where $\V_{\pa(i)}$ contains three distinct values or more. In particular,
\[
    \mathcal{N}\left( \begin{bmatrix} 0 \\ 0 \\ 0\end{bmatrix}, \begin{bmatrix}
        1 & -1 & -1 \\ -1 & 1 & -1 \\ -1 & -1 & 1
    \end{bmatrix} \right).
\]
is ill-defined since the proposed covariance matrix is not positive semidefinite. Less formally, note that if world n°1 is negatively correlated with worlds n°2 and n°3, then worlds n°2 and n°3 are necessarily positively correlated. Consequently, one cannot impose the above correlations. Instead, for example, one can have
\[
    N := \mathcal{N}\left( \begin{bmatrix} 0 \\ 0 \\ 0\end{bmatrix}, \begin{bmatrix}
        1 & -1 & -1 \\ -1 & 1 & 1 \\ -1 & 1 & 1
    \end{bmatrix} \right).
\]
Among the three cross-dependencies, two are countermonotonic and one is comonotonic. To sum up, in contrast to comonotonicity, countermonotonicity can occur locally (at the coupling scale) but not globally (at the process scale). It is a well-known result from copula theory \citep{kotz1992stochastic,joe2014dependence}.
\end{remark}

\begin{remark}[About measurability, composability, and independence]\label{rem:measurability}
\Cref{def:norm} imposes normalizations to be composable, so that one-step-ahead counterfactual distributions are also composable (\cref{lem:norm_comp}). Composability is notably crucial in \cref{thm:representation}. This naturally raises the question on which process distributions are composable or not. The literature on stochastic processes has already established rather technical sufficient and/or necessary conditions \citep{ambrose1940measurable,chung1965fields,cohn1972measurable,hoffmann1973existence}. We point out two simple specific cases that are relevant to define Gaussian normalizations in practice: $\Q(\X,\Y) = \Q^*(\X,\Y)$ if $\X$ is countable and $N^k \in \Q^*(\X,\R)$ if $k$ is continuous. We refer to \cite{bogachev1998gaussian} for more general conditions for Gaussian measures. 

We also emphasize that if $k(x,x')=0$ for all $x \neq x'$ and if $\X$ is uncountable then $N^k$ is not composable, and thereby is not a proper normalization. This critically means that in many settings \emph{there is no \say{independent} counterfactual conceptions that is admissible.}
\end{remark}

\subsubsection{Implementation}\label{sec:implementation}

To apply the above framework, one needs to learn $\psi^\C_i : \R \times \V_{\pa(i)} \to \V_i$, which defines $\Psi^\C_i$. In general cases, it would be intractable to learn separately each monotonic transport $\psi^\C_i(\bullet | v_{\pa(i)})$ for every $v_{\pa(i)} \in \V_{\pa(i)}$. This is why learning $\psi^\C_i$ as a single model with two inputs is necessary. To achieve this, we carry out a distributional regression under monotonicity constraints to obtain a non-decreasing conditional generative model. Concretely, we approach
\begin{equation}\label{eq:learning}
    \min_{\psi \in \digamma_i} \E\left[ D \left( \psi(\bullet | V_{\pa(i)}) \# \Norm, K_i(\bullet | V_{\pa(i)}) \right) \right],
\end{equation}
where $\digamma_i$ is a class of measurable functions from $\R \times \R^{\abs{\pa(i)}}$ to $\R$ that are non-decreasing in the first input and $D : \mathcal{P}(\R) \times \mathcal{P}(\R) \to \R_+$ is the (power of) a distance of divergence between probability measures. This optimization problem borrows idea from \citep{alquier2024universal,shen2024engression,dance2025counterfactual}.

Let us briefly show that the problem is well posed (we refer to \cref{prop:posedness} for a more general proof). If $\E\left[ D \left( \psi(\bullet | V_{\pa(i)}) \# \Norm, K_i(\bullet | V_{\pa(i)}) \right) \right] = 0$ then by nonnegativeness $D \left( \psi(\bullet | V_{\pa(i)}) \# \Norm, K_i(\bullet | V_{\pa(i)}) \right)=0$ almost surely, which implies that $\psi(\bullet | v_{\pa(i)}) \# \Norm = K_i(\bullet | v_{\pa(i)})$ for (almost) every $v_{\pa(i)} \in \V_{\pa(i)}$. Therefore, if $\digamma_i$ possesses $\psi^\C_i$, then solving \cref{eq:learning} returns $\psi^\C_i$ (up to optimization errors) by identifiability. Otherwise it returns an approximation of $\psi^\C_i$ due to model misspecification.

This means that designing $\digamma_i$ is crucial; it must be as close as possible to the set of functions non-decreasing in their first input. In practice, we choose $\digamma_i := \{ \psi_\theta, \theta \in \Theta \}$ as a parametric class of neural networks with fixed architecture and monotonicity constraints, and optimize the parameters of the model. More precisely, we rely on the work of \cite{runje2023constrained} to build such networks. This paper recalls that constraining neural networks to be non-decreasing in specific inputs is a simple problem: it suffices to take non-decreasing activation functions and to force positivity on specific weights. However, it highlights that if one employs only classical activation functions (like ReLU) which are convex, then the obtained networks will be \emph{convex non-decreasing} in the concerned inputs, thereby will fail to approximate general non-decreasing functions. The interest of its contribution lies in considering fancier versions of classical activation functions to guarantee universal approximation of non-decreasing functions. We refer to \cref{sec:regression} for more details on the optimization procedure and to \cref{sec:exp} for the numerical experiments.

We crucially emphasize that solving \cref{eq:learning} only requires $\C$. Notably, there is no counterfactual conception at work at this stage of the implementation. After estimating $\Psi^\C_i$, the analyst can simply choose a normalization $N^{(i)}$ formalizing their counterfactual conception (which has no computational cost), sample from $N^{(i)}$ (which has a light computational cost), and then pass the samples through $\Psi^\C_i$ to sample from $S^{(i)}$. Note that the separation of the interventional structure from the intrinsic counterfactual dependencies permitted by the normalization makes testing various counterfactual conceptions almost free. The fact that the content of the counterfactual model corresponding to the singular-causation layer (namely, the normalization) does not require inference from data is consistent with the fact that individual counterfactuals cannot be tested: they must be chosen. This contrasts against the SCM framework, in which one must learn a new model compatible with $\C$ for each counterfactual conception. In what follows, we compare in more details the structural approach with the canonical approach.

\subsection{Comparison to the structural approach}\label{sec:comp_to_scm}

We proposed a framework to choose, formalize, and implement counterfactual conceptions via normalizations of counterfactual models. This section discusses how these aspects compare with structural causal models.

\subsubsection{Modeling counterfactuals}\label{sec:modeling}

We start with the modeling aspects. We distinguish two questions: \emph{reading} a counterfactual conception from a causal model and \emph{writing} a counterfactual conception into a causal model. A reader claims: \say{Give me your causal model and I will tell you the counterfactuals in which you believe in.} A writer claims: \say{Tell me the counterfactuals in which you believe in and I will give you a causal model.}

Let us commence with \emph{reading}. In a normalized counterfactual model, the counterfactual conception is plainly expressed by the dependence structure of its normalization (by its covariance function in the Gaussian setting). In a structural causal model, the counterfactual conception derives from the form of the structural equations. To illustrate this point, we present notorious connections (see for instance \citep{nasr2023counterfactual}) between a one-variable causal mechanism $(f_i,P_{0,i})$ and its entailed one-step-ahead counterfactual distribution $S^{(i)}$ for $i \in [d]$, adapted to our framework and notation.


\begin{proposition}[From structural equations to counterfactuals]\label{prop:scm_to_ctf}
Let $\V := \times^d_{i=1} \V_i \subseteq \R^{d}$ be a Cartesian product of Borel subsets of $\R$, $\C \in \CGM_{\V}$, $\M:= (\V,\G,\U,P_0,f) \in \SCM_{\V,\C}$, and $i \in [d]$. The two following properties hold.
\begin{itemize}
    \item[(i)] If the function $f_i(v_{\pa(i)},\bullet)$ is injective for every $v_{\pa(i)} \in \V_{\pa(i)}$, then the entailed one-step-ahead counterfactual distribution $S^{(i)}$ is deterministic.
    \item[(ii)] If the functions $\left(f_i(v_{\pa(i)},\bullet)\right)_{v_{\pa(i)} \in \V_{\pa(i)}}$ are all non-decreasing or all non-increasing, then the entailed one-step-ahead counterfactual distribution $S^{(i)}$ admits $N^{\uparrow}$ as normalization.
\end{itemize}
\end{proposition}
 
We argue that reading the conceptual conception carried by a structural causal model is less direct than with a normalization. While the conditions (i) and (ii) from the above proposition are not particularly involved, they still require a mathematical analysis of $f_i$. Additionally, in the case where $f_i$ entails a stochastic one-step-ahead counterfactual distribution, there is no obvious indicator of the stochasticity of the counterfactuals, whereas the covariance function of a Gaussian normalization naturally includes this information. More generally, precisely reading counterfactual conceptions from SCMs that do not fit \cref{prop:scm_to_ctf} remains a largely open question.


\begin{remark}[On the role of noises]\label{rem:noises}
Note that, if $f_i(v_{\pa(i)},\bullet)$ is non-decreasing and $P_{0,i} = \Norm$, then $f_i(v_{\pa(i)},\bullet) = \psi^\C_i(\bullet|v_{\pa(i)})$ by uniqueness of the non-decreasing transport map from $\Norm$ to $K_i(\bullet|v_{\pa(i)})$. This might mislead people to think that $f_i(v_{\pa(i)},\bullet)$ and $\psi^\C_i(\bullet|v_{\pa(i)})$ play a similar role in general. We underline that it is not the case, even in the specific situation where the two functions mathematically coincide.

Let us denote by $N^{(i)}$ a normalization of $S^{(i)}$, the one-step-ahead counterfactual distribution entailed by some SCM $\M$ at $i$. The following equality always holds:
\[
\left( f_i(v_{\pa(i)},\bullet) \right)_{v_{\pa(i)} \in \V_{\pa(i)}} \# P_{0,i} = \left( \underset{v_{\pa(i)} \in \V_{\pa(i)}}{\times} \psi^\C_i(\bullet|v_{\pa(i)}) \right) \# N^{(i)}.
\]
Even if $f_i(v_{\pa(i)},\bullet) = \psi^\C_i(\bullet|v_{\pa(i)})$ and $P_{0,i} = \Norm$, the left-hand side and the right-hand side operate differently due to the distinct nature of their respective input noises (namely, sources of randomness). The left-hand-side function transports a single distribution towards a joint probability distribution, whereas the right-hand-side function transports a joint probability distribution towards a joint probability distribution with same dimension. Moreover, changing the noise entering $\left( f_i(v_{\pa(i)},\bullet) \right)_{v_{\pa(i)} \in \V_{\pa(i)}}$ is radically different from changing the noise entering $\Psi^\C_i$. When $P_{0,i}$ varies in $\Prob{\U_i}$ the marginals of the random process $\left( f_i(v_{\pa(i)},\bullet) \right)_{v_{\pa(i)} \in \V_{\pa(i)}} \# P_{0,i}$ changes; when $N^{(i)}$ varies in $\Normac{\V_{\pa(i)}}$ the marginals of $\Psi^\C_i \# N^{(i)}$ remain equal to $K_i(\bullet|\V_{\pa(i)})$.
\end{remark}

Let us advance to \emph{writing}. For this task, the structural approach faces two challenges: maintaining compatibility to the given CGM and properly translating a counterfactual conception into its formalism. To explain the question of compatibility, consider an analyst that aims at implementing a comonotonic conception of counterfactuals via an SCM entailing a given CGM $\C$. It seems by \cref{prop:scm_to_ctf} that it suffices to select $f_i$ such that $f_i(v_{\pa(i)},\bullet)$ is strictly monotonic for every $v_{\pa(i)} \in \V_{\pa(i)}$. However, this condition solely ensures the comonotonic property---not the compatiblity to $\C$. More precisely, choosing any strictly monotonic function is not enough: the mechanism must also meet $f_i(v_{\pa(i)},\bullet) \# P_{0,i} = P_{i|\pa(i)}(\bullet|v_{\pa(i)})$ for every $v_{\pa(i)} \in \V_{\pa(i)}$. Failing to meet this transport requirement would entail counterfactual distributions inconsistent with the observational and interventional distributions. Critically, finding a mechanism simultaneously matching the desired counterfactual conception \emph{and} satisfying compatibility to the CGM is difficult. To further illustrate this point, we study the most common type of SCMs: those in which the exogenous noises are additive terms of the structural equations.

\begin{definition}[Additive noise models]\label{def:anm}
Let $\V := \times^d_{i=1} \V_i \subseteq \R^{d}$ be a Cartesian product of Borel subsets of $\R$. An SCM $\M:= (\V,\G,\U,P_0,f) \in \SCM_{\V}$ is an \emph{additive noise model} (ANM) if for every $i \in [d]$ there exists a measurable function $h_i : \V_{pa(i)} \to \R$ such that
\[
    f_i(v_{\pa(i)},u_i) = h_i(v_{\pa(i)}) + u_i,
\]
for every $(v_{\pa(i)},u_i) \in \V_{\pa(i)} \times \U_i$. We say that a CGM $\C \in \CGM_{\V}$ admits an ANM if there exists an ANM $\M$ such that $\C_\M = \C$.
\end{definition}

ANMs became popular for various reasons, by notably allowing \emph{causal discovery} (see \cref{rem:discovery} below). But also because they provide a cognitively natural and \emph{seemingly} general form for the structural equations. Before all, we remind that despite their apparent generality they fully determine the counterfactual layer: since $f_i(v_{\pa(i)},\bullet)$ is strictly monotonic for every $v_{\pa(i)} \in \V_{\pa(i)}$, it entails the deterministic comonotonic conception of counterfactuals. As such, we invite interventional analysts manipulating such models to be aware that this mathematical form matters at the counterfactual level (recall \cref{rem:equations}) and we recommend counterfactual analysts to explicitly mention the specific counterfactual choice they are making by choosing such a structure. That being said, let us turn back to the question of compatibility. We emphasize that---in the Markovian case---ANMs cannot fit any probability distributions: for some CGMs, compatible strictly monotonic causal mechanisms can be far from being ANMs.

\begin{proposition}[Additive noise models do not always exist]\label{prop:not_anm}
Let $P_{\mathtt{x,y}} \in \mathcal{P}(\R^2)$ be defined by any $P_{\mathtt{x}} \in \mathcal{P}(\R)$ and $P_{\mathtt{y}|\mathtt{x}}(\bullet|x) := \mathcal{N}(m(x),\sigma(x)^2)$ for every $x \in \R$, where $m : \R \to \R$ and $\sigma : \R \to \R$ are two functions such that $P_{\mathtt{y}|\mathtt{x}}$ is a well-defined probability kernel. Additionally, let $\G$ be the graph given by $\mathtt{x} \rightarrow \mathtt{y}$. Then, $\C := (\R^2,\G,\K)$ with $\K := (P_{\mathtt{x}},P_{\mathtt{y}|\mathtt{x}})$ admits an ANM if and only if $\sigma^2$ is constant.
\end{proposition}
On the contrary, when specifying a counterfactual conception with a normalization, the produced counterfactual model fits $\C$ by design through the functions $(\Psi^\C_i)^d_{i=1}$. Conveniently, the analyst can choose the intrinsic counterfactual cross-dependencies independently of $\C$, thereby without altering the observational and interventional knowledge. In particular, every $\C$ always admits a comonotonic one-steap-ahead counterfactual processes $(S^{(i)})^d_{i=1}$ but not necessarily an ANM. This shows the advantage of separating $\C$ from purely counterfactual assumptions. Note that people often use SCMs in toy examples, where the full model serves as a primitive describing the world's functioning at the three causal-inference layers. In such hypothetical scenarios, the SCM is not constrained to fit a given CGM; it defines the correct CGM. Consequently, compatibility to $\C$ holds by design. However, while interesting in theory, notably for pedagogical reasons, toy examples do not correspond to real-world inference tasks. This is why compatibility remains a significant practical issue in the process of formalizing counterfactual beliefs. 

Moreover, even leaving aside the question of compatibility, it feels unclear how to encode a given counterfactual conception by a class of functions (representing structural equations) beyond the cases presented in \cref{prop:scm_to_ctf}. In contrast, writing any counterfactual conception is straightforward with normalizations of counterfactual models.

\begin{remark}[SCMs in causal discovery]\label{rem:discovery}
Several papers about causal inference rely on SCMs in problems that do not specifically deal with the counterfactual layer of Pearl's hierarchy. This typically concerns \emph{causal discovery} problems, in which the objective consists in identifying the causal graph (or at least a collection of admissible graphs) from the observational distribution. As explained in \citep[Section 4.2]{peters2017elements}, several causal-discovery methods are based on the assumption that the observational distribution follows an ANM \citep{hoyer2008nonlinear,shimizu2014lingam,peters2014causal}. As such, it might seem that these techniques rests on a \emph{counterfactual} assumption (that is, the functional form of the causal model) to make \emph{interventional} inference. To clarify, notice that the assumption can be formally written as: given $\C \in \CGM_{\V}$, there exists an ANM $\M_\star \in \SCM_{\V,\C}$ such that $\C_{\M_\star} = \C$. This only signifies that the causal kernels $\K$ of $\C$ are such that the deterministic comonotonic counterfactual conception over $\C$ takes the form of an ANM---it does not impose to work with this conception. Therefore, it is not a counterfactual assumption. If an analyst aims to carry out counterfactual reasoning on top of the learned $\C$, they may perfectly choose an $\M \in \SCM_{\V,\C}$ or an $\A \in \CR_{\V,\C}$ that can be different from the $\M_\star$ that served to learn $\C$. To summarize, supposing that a CGM admits an ANM radically does not mean that the latent SCM is an ANM. Additionally, we emphasize that causal discovery never deals with learning an SCM---which is not identifiable from observational data---only the graph. In this sense, \cite{peters2014causal} clearly states that their ANM assumption cannot be used to infer counterfactual queries.

Moreover, people frequently use an SCM (typically an ANM) as a tool to generate observational and interventional data, notably in toy examples. At this level, the functional form of the SCM does not fundamentally matter, it simply provides a convenient representation to simulate data.

All in all, we caution against misinterpreting the role of SCMs in causal-inference research. At the interventional level, they serve to formalize assumptions on the CGM or to generate data. At the counterfactual level, they traduce an arbitrary choice of counterfactual conception. In particular, despite the prominence of ANMs, an SCM never needs to be an ANM given only a CGM. This is why we recommend causality scholars to explicitly specify the layer at which they operate in their works and to justify their assumptions and choices.
\end{remark}

\begin{remark}[Discussing counterfactual assumptions]\label{rem:discussing}
As aforementioned, we think that the causality literature that addresses individual causation does not discuss enough the choices of the SCMs in light of the counterfactual conception. For instance, \cite{kusner2017counterfactual} define a notion of individual-wise algorithmic fairness, called \emph{counterfactual fairness}, requiring equal treatment between every input and their counterfactual counterparts had the protected attribute (typically, sex, race, or age) changed. This criterion crucially depends on the counterfactual conception: a different conception could yield radically distinct conclusions on fairness.\footnote{\cite{wu2019counterfactual} focus on bounding the unidentifiability of counterfactual fairness.} As such, it is critical that the conception be purposely and plainly selected. However, the vast literature on counterfactual fairness and its variations generally designs the SCMs with a predefined forms (often, as an ANM by default) without discussing the implications of such a choices onto the counterfactual conception \citep{chiappa2019path,zuo2022counterfactual,alvarez2023counterfactual,zuo2024lookahead,zhou2024counterfactual,alvarez2025counterfactual}.

We point out that \citep[Section 4.2]{kusner2017counterfactual} mentions that many structural assignments are compatible with the same observational and intervention data, and thereby underline that such components of SCMs are prone to discussion. Actually, the authors dedicate another article to considering multiple causal assumptions (including counterfactual ones) \citep{russell2017worlds}. Nevertheless, these works do not explicitly relate the form of the SCM to the counterfactual distribution (like being an ANM and entailing comonotonic counterfactuals).
\end{remark}

Overall, we note that while the causality literature frequently highlights that SCMs allow counterfactual reasoning (contrarily to CGMs), it rarely explains and justifies the employed SCMs through the prism of the (implicitly) decided counterfactual conception. A virtue of CGMs comes from their transparency: a nonscientific audience can readily understand the interventional information carried by causal arrows between variables. However, SCMs do not present counterfactual assumptions as intelligibly. Clarifying these assumptions is crucial in counterfactual reasoning problems, notably because they cannot be falsified. Canonical representations of SCMs and their normalizations serve to highlight the arbitrariness of counterfactual assumptions, by separating what is possibly falsifiable from what is never testable in an SCM, and to furnish an intelligible formalization of these assumptions. Thereafter, we discuss more practical considerations.


\subsubsection{Estimating a model}

As previously explained, learning a counterfactual model over a given CGM $\C$ by normalization demands two fundamentally different steps. The first is to \emph{learn} the mapping $\Psi^\C_i$ by solving \cref{eq:learning}. The second is to \emph{choose} a normalization $N^{(i)}$. In contrast, learning an SCM over a given CGM $\C$ blends these two steps. Typically, one selects a parametric hypothesis class $\H_i := \{ (f_{\theta,i}, P_{0,i}), \theta \in \Theta_i \}$ and optimizes over $\theta$ so that $f_{\theta,i}(V_{\pa(i)}, \bullet) \# P_{0,i} \approx P_{i | \pa(i)}(\bullet | V_{\pa(i)})$ on samples. Note that this can be achieved by solving a similar problem to \cref{eq:learning}:
\[
    \min_{(f_i,P_{0,i}) \in \H_i} \E\left[ D(f_i(V_{\pa(i)},\bullet) \# P_{0,i}, P_{i | \pa(i)}(\bullet | V_{\pa(i)})\right].
\]
Critically, the design of $\H_i$ matters for both the compatibility to $\C$ and the choice of the counterfactual conception. On the one hand, $\H_i$ must be expressive enough so that $f_{\theta,i}(V_{\pa(i)}, \bullet) \# P_{0,i} \approx P_{i | \pa(i)}(\bullet | V_{\pa(i)})$. On the other hand, it fixes admissible counterfactual conceptions, which are implicitly encoded in the function class that $\H_i$ aims at approximating (\textit{e.g.}, ANMs entail deterministic comonotonic counterfactuals).

This approach has several disadvantages. Notably, the vulnerability to model misspecification. Some classical choices of $\H_i$ (linear additive models, additive noise models, \ldots) are often \emph{overspecified}: they could poorly fit $P_{i | \pa(i)}$, leading to inaccurate observational and interventional estimators (as detailed in \cref{sec:modeling}). In contrast, by confining the compatibility to $\C$ into the estimation of $\Psi^\C_i$, estimating a counterfactual model fits the interventional marginals by design. Attempting to solve this issue by choosing a too large $\H_i$ can render the model \emph{underspecified}, as it could fail to identify a single counterfactual conception. For example, taking a class of unconstrained neural networks would permit to match $P_{i | \pa(i)}$, but the obtained $f_{\theta,i}(v_{\pa(i)},\bullet)$ could equivalently be monotonic or nonmonotonic. In such cases, $\H_i$ contains several conceptions but leaves the determination of the implemented conception to the arbitrariness of the optimization process. On the contrary, the normalizing mapping $\Psi^\C_i$ of counterfactual models is identifiable. Then, choosing the counterfactual conception amounts to choosing a normalization. Of course, an SCM analyst could remedy this specification issue by choosing a large enough $\H_i$ that still uniquely identifies the desired counterfactual conception. For instance, if they want to implement a comonotonic conception of counterfactuals, they could use the same class of constrained monotonic neural networks that we employ for \cref{eq:learning} (which has universal approximation guarantees). However, this solution only works for comonotonic counterfactuals. It is unclear which class of functions correspond to less conventional counterfactual conceptions. All in all, normalizations form a simpler framework to estimate a counterfactual model than classes of functions.

This points towards another limitation of the structural approach for counterfactual modeling: the computational cost of implementing various counterfactual conceptions. Say that an analyst aims at trying two distinct counterfactual conceptions. To do so with SCMs, they must propose two hypothesis classes $\H_i$ and $\H'_i$ capturing these conceptions (which can already be a difficult task) and then optimize the two corresponding models. On the contrary, with counterfactual models they must only learn the transport model $\Psi^\C_i$ \emph{once}, and then feed it with two distinct latent processes $N^{(i)}$ and $N'^{(i)}$. More generally, this enables them to test any counterfactuals without additional estimation tasks.\footnote{This feature of normalizations can notably be useful in frameworks taking multiple counterfactual assumptions into account, like \citep{russell2017worlds}.} Importantly, learning $\psi^\C_i$ (which offers the flexibility to specify any counterfactual conceptions) brings no extra computational cost compared to learning an $f_i$ (which fixes the counterfactuals). In both cases, one must solve a conditional generative modeling problem. Overall, canonical representations encoded through normalizations are more practical than SCMs to explore diverse counterfactual assumptions compatible with given interventional knowledge.

\section{Distributional regression}\label{sec:regression}

In practice, learning $\Psi^\C_i$ amounts to solving a distributional regression problem under monotonicity constraints: \cref{eq:learning}. This requires two ingredients: a computable finite-sample loss and an expressive class of partially non-decreasing models. For the sake of generality, we study this learning problem in a generic form that need not be motivated by causality. In particular, $i$ exceptionally indexes a sample rather than a variable in this section.

\subsection{Optimization problem}

We address the generic distributional regression problem of the form
\begin{equation}\label{eq:generic_learning}
    \min_{\psi \in \digamma} \E\left[ D \left( \psi(\bullet | X) \# P_{\mathtt{e}}, P_{\mathtt{y} | \mathtt{x}}(\bullet | X) \right) \right]
\end{equation}
where $P_{\mathtt{x,y}}$ is the target observational distribution over $\R^d \times \R$, $P_{\mathtt{e}}$ is a fixed \emph{atomless} input noise distribution over $\R$, $(X,Y) \sim P_{\mathtt{x},\mathtt{y}}$, $D$ is from $\Prob{\R}^2$ to $[0,+\infty]$, and $\digamma$ is a set of measurable functions from $\R \times \R^d$ to $\R$. Note that \cref{eq:learning} fits this problem. The following result studies the well-posedness of \cref{eq:generic_learning} to carry out distributional regression, with and without monotonicity constraint.

\begin{proposition}[Well-posedness of distributional regression]\label{prop:posedness}
Consider the problem from \cref{eq:generic_learning} with $D : \Prob{\R}^2 \to [0,+\infty]$ such that $D(P,P') = 0 \iff P=P'$.
\begin{enumerate}
    \item If there exists $\tilde{\psi} \in \digamma$ such that $\tilde{\psi}(\bullet|x) \# P_{\mathtt{e}} = P_{\mathtt{y} | \mathtt{x}}(\bullet | x)$ for $P_{\mathtt{x}}$-almost-every $x \in \R^d$, then every solution $\psi^*$ to \cref{eq:generic_learning} achieves $\psi^*(\bullet|x) \# P_{\mathtt{e}} = P_{\mathtt{y} | \mathtt{x}}(\bullet | x)$ for $P_{\mathtt{x}}$-almost-every $x \in \R^d$.
    \item If $\digamma$ is such that $\psi(\bullet|x)$ is non-decreasing for every $\psi \in \digamma$ and every $x \in \R^d$, and if there exists $\tilde{\psi} \in \digamma$ such that $\tilde{\psi}(\bullet|x) \# P_{\mathtt{e}} = P_{\mathtt{y} | \mathtt{x}}(\bullet | x)$ for $P_{\mathtt{x}}$-almost-every $x \in \R^d$, then \cref{eq:generic_learning} admits a unique solution.
\end{enumerate}
\end{proposition}

This approach is very similar to ones of \cite{shen2024engression} and \cite{dance2025counterfactual}, and actually incorporates several of their ideas. The specificity is that we focus on the case where the model class $\digamma$ is such that $\psi(\bullet|x)$ is non-decreasing for every $x \in \R^d$. Not only the monotonicity constraint is necessary in the context of normalized counterfactual models and quantile regression (as detailed below), but it also guarantees identifiability of the solution (as proved above).

\subsubsection{Finite sample loss}\label{sec:finite_loss}

To solve this program numerically, we consider a finite sample version. This demands two steps since there are two probability distributions at work. First, suppose that we have access to a dataset $\{(x_i,y_i)\}^n_{i=1}$ of $n$ independent samples from $P_{\mathtt{x},\mathtt{y}}$. Replacing $P_{\mathtt{x, y}}$ by its empirical counterpart $\frac{1}{n} \sum^n_{i=1} \delta_{(x_i,y_i)}$ in \cref{eq:generic_learning} gives   
\begin{equation}\label{eq:generic_learning_sample_1}
    \min_{\psi \in \digamma} \frac{1}{n} \sum^n_{i=1} D \left( \psi(\bullet | x_i) \# P_{\mathtt{e}}, \delta_{y_i} \right),
\end{equation}
where $\delta_{y_i}$ can be seen as an estimation $P_{\mathtt{y} | \mathtt{x}}(\bullet | x_i)$. Second, we sample for each $i \in [n]$ a batch $\{e_{i,j}\}^{n'}_{j=1}$ of $n'$ independent samples from $P_{\mathtt{e}}$. In practice, we always take $n'=64$. Replacing $P_{\mathtt{e}}$ by its empirical counterpart $\frac{1}{n'} \sum^{n'}_{j=1} \delta_{e_{i,j}}$ for every $i \in [n]$ in \cref{eq:generic_learning_sample_1} gives
\begin{equation}\label{eq:generic_learning_sample_2}
    \min_{\psi \in \digamma} \frac{1}{n} \sum^n_{i=1} D \left( \psi(\bullet | x_i) \# \frac{1}{n'} \sum^{n'}_{j=1} \delta_{e_{i,j}}, \delta_{y_i} \right).
\end{equation}
To continue, we follow \cite{alquier2024universal} and \cite{dance2025counterfactual} and take $D$ as the square of a \emph{Maximum Mean Discrepancy} (MMD) \citep{gretton2012kernel}. For every positive definite kernel $K : \R \times \R \to \R$, the squared MMD between $P,P' \in \Prob{\R}$ relative to $K$ is given by
\begin{multline*}
    \MMD^2_K(P,P') := \iint K(y,y') \rmd P(y) \rmd P(y') - 2 \iint K(y,y') \rmd P(y) \rmd P'(y')\\ + \iint K(y,y') \rmd P'(y) \rmd P'(y').
\end{multline*}
If $K$ is bounded, continuous, and characteristic, then $\MMD_K$ is a distance over $\Prob{\R}$, differentiable in each of its input measures. Note that this kernel $K$ plays a different role that the kernel $k$ of Gaussian normalizations. Writing $y^\psi_{i,j} := \psi(e_{i,j} | x_i)$ and considering the expression of the MMD between empirical measures yields
\begin{equation}\label{eq:mmd_learning_sample_}
    \min_{\psi \in \digamma} \frac{1}{n} \sum^n_{i=1} \frac{1}{{n'}^2} \sum^{n'}_{j,j'=1} K(y^\psi_{i,j},y^\psi_{i,j}) - 2 K(y^\psi_{i,j},y_i) + K(y_i,y_i).
\end{equation}
Since only the $y^\psi_{i,j}$ depend on the model $\psi$ that we aim at learning, one can drop the constant term $K(y_i,y_i)$ in the above minimization problem. In practice, we minimize, using automatic differentiation and gradient-descent-based routines, the loss function
\[
    \Loss_K(\psi,\{(x_i,y_i)\}^n_{i=1},n') := \frac{1}{n} \sum^n_{i=1} \frac{1}{{n'}^2} \sum^{n'}_{j,j'=1} K(y^\psi_{i,j},y^\psi_{i,j'}) - 2 K(y^\psi_{i,j},y_i)
\]
in the parameters of $\psi$. Recall that $K(y,y')$ increases with the similarity of $y$ and $y'$ (as opposed to a distance). Therefore, the term $- 2 K(y^\psi_{i,j},y_i)$ promotes the $j$th random prediction at $x_i$ to be close to the sole known output at $x_i$, while the term $K(y^\psi_{i,j},y^\psi_{i,j'})$ introduces variance within the predictions at $x_i$ so that they fit a better approximation of $P_{\mathtt{y} | \mathtt{x}}(\bullet | x_i)$ than merely $\delta_{y_i}$. As such, it is this variance penalty that makes this problem a \emph{distributional} regression rather than a classical point-wise regression.


\subsubsection{Regression models}\label{sec:models}

This short section presents how we construct $\digamma$ with monotonicity constraints in practice. For any dataset corresponding to a population distribution $P_{\mathtt{x,y}}$ over $\R^d \times \R$, we define the model class $\digamma$ as a set of partially monotonic feed-forward neural networks $\psi : \R \times \R^d \to \R$ with a fixed architecture of $L$ hidden layers of size $R$. The partial monotonicity constraint renders the function $\psi(\bullet | x)$ non-decreasing for every $x \in \R^d$ by applying the monotonic layers introduced by \cite{runje2023constrained}. For concision, and to keep the paper self-contained, we summarize their method in \cref{sec:monotonic_networks}. Their approach notably combines a convex base activation with concave and bounded versions, and we employ ReLUs as base activation functions in practice. 

\subsection{Applications}

We present several statistical and causal estimators that one can derive from an estimation $\widehat{\psi}_{\mathtt{y}}$ of the solution to \cref{eq:generic_learning}.

\subsubsection{Classical regressions}

A first possibility consists in using $\widehat{\psi}_{\mathtt{y}}$ as a classical regression function to make pointwise predictions. For illustration, let $(X,Y) \sim P_{\mathtt{x,y}}$, where $X \independent E \sim P_{\mathtt{e}}$. The population solution $\psi_{\mathtt{y}}$ recovers the conditional mean $m(x) := \E[Y | X=x]$ (that is, the solution of mean-squared regression) via $\E[Y | X=x] = \E[ \psi_{\mathtt{y}}(E|x) ] = \int \psi_{\mathtt{y}}(e|x)  \rmd P_{\mathtt{e}}(e)$. One can construct an estimator by replacing the population solution with the finite-sample solution and through sampling from $P_{\mathtt{e}}$. Given a collection of independent samples $\{e_j\}^n_{j=1}$ from $P_{\mathtt{e}}$, one estimates the conditional mean function of $P_{\mathtt{y}|\mathtt{x}}$ as
\[
    \widehat{m}(x) := n^{-1} \sum^n_{j=1} \widehat{\psi}_{\mathtt{y}}(e_j | x).
\]
One can proceed similarly for conditional quantiles. Notice that, by monotonicity, $\psi_{\mathtt{y}}(G^{P_{\mathtt{e}}}(\bullet) | x)$ is the quantile function of $P_{\mathtt{y}|\mathtt{x}}(\bullet|x)$. As such, $\psi_{\mathtt{y}}(G^{P_{\mathtt{e}}}(q) | \bullet)$ is $m_q$: the quantile regression function of order $q$ of $P_{\mathtt{y}|\mathtt{x}}$, which is also the solution of the mean-pinball-loss regression of order $q$. Consequently, for every $q \in [0,1]$, one estimates $m_q(x)$ as
\[
    \widehat{m}_q(x) := \widehat{\psi}_{\mathtt{y}}(G^{P_{\mathtt{e}}}(q) | x).
\]
In contrast to the conditional mean, there is no need to sample from $P_{\mathtt{e}}$ to make this estimation. Note that when $G^{P_{\mathtt{e}}}$ is unknown, one can replace it by an estimator $\widehat{G}^{P_{\mathtt{e}}}$. 

In the context of normalized counterfactual models, one takes $P_{\mathtt{e}} := \Norm$. But if the objective is to carry out quantile regression, then it is more straightforward to take $P_{\mathtt{e}} := \Unif{[0,1]}$ so that $G^{P_{\mathtt{e}}} = \Id$. Interestingly, doing quantile regression by solving \cref{eq:generic_learning} allows to learn simultaneously \emph{all} the quantile of $Y$ conditional to $X=x$ whereas standard quantile regression solves a distinct optimization problem for each quantile. Therefore, it makes infinite quantile learning computationally tractable (see the work of \cite{brault2019infinite} for an alternative solution). Moreover, it prevents, by monotonicity, the typical quantile crossing problem that occurs when learning quantile curves separately \citep{bassett1982empirical}. We refer to \cref{sec:quantile_regression} for numerical experiments.

\subsubsection{Causal estimations}\label{sec:causal_estimation}

A second possibility consists in using $\widehat{\psi}_{\mathtt{y}}$ to normalize counterfactual models and derive causal effects. We keep similar notation to above and assume that $P_{\mathtt{e}} := \Norm$. For readability, we rely on the random-variable formalism presented in \cref{rem:variables}. We distinguish two types of estimands: general-causation effects and singular-causation effects. 

A general-causation estimand can be formulated without specifiying the normalization. For example, $m^\Phi(x) := \E[Y^\Phi | X=x] = \E\left[\psi_{\tty}\left((E^{(\tty)}_{\phi_{\ttx}(x)}|\phi_{\ttx}(x)\right)\right] = \int \psi_{\mathtt{y}}(e | \phi_{\mathtt{x}}(x)) \rmd P_{\mathtt{e}}(e)$ since $E^{(\tty)}_{\phi_{\ttx}(x)} \sim N^{(\tty)}_{\phi_{\ttx}(x)} = \Norm$ by marginalization. Plugging in $\widehat{\psi}_{\mathtt{y}}$ and a collection $\{e_j\}^n_{j=1}$ of independent samples from $\Norm$ furnishes the estimator
\[
    \widehat{m^\Phi}(x) := n^{-1} \sum^n_{j=1} \widehat{\psi}_{\mathtt{y}}(e_j | \phi_{\mathtt{x}}(x)).
\]
A singular causation estimand requires specifying the normalization $N^{(\mathtt{y})}$. Consider $m^{\Phi_1,\Phi_2}(x) := \E[(Y^{\Phi_1}, Y^{\Phi_2}) | X=x]$ as a first example. It can be written as
\begin{multline*}
m^{\Phi_1,\Phi_2}(x) = \E\left[\left(\psi_{\tty}\left(E^{(\tty)}_{\phi_{1,\ttx}(x)}|\phi_{1,\ttx}(x)\right), \psi_{\tty}\left(E^{(\tty)}_{\phi_{2,\ttx}(x)}|\phi_{2,\ttx}(x)\right)\right)\right]\\ = \int (\psi_{\mathtt{y}}(e_1 | \phi_{1,\mathtt{x}}(x)), \psi_{\mathtt{y}}(e_2 | \phi_{2,\mathtt{x}}(x)))  \rmd N^{(\tty)}_{\phi_{1,\ttx}(x),\phi_{2,\ttx}(x)}(e_1,e_2).  
\end{multline*}
By drawing independent samples $\{(e_{1,j},e_{2,j})\}^{n'}_{j=1}$ from $N^{(\mathtt{y})}_{\phi_{1,\mathtt{x}}(x),\phi_{2,\mathtt{x}}(x)}$, one can estimate this effect with
\[
    \widehat{m^{\Phi_1,\Phi_2}}(x) := n^{-1} \sum^n_{j=1} (\psi_{\mathtt{y}}(e_{1,j} | \phi_{1,\mathtt{x}}(x)), \psi_{\mathtt{y}}(e_{2,j} | \phi_{2,\mathtt{x}}(x))).
\]
Nevertheless, most singular causation estimands are expressed with a conditional distribution rather than a joint distribution. Consider $\E[Y^{\Phi_2} | Y^{\Phi_1}=y, X=x]$ as a second example. Note that it is given by the population solution of the mean-squared regression problem between of $Y^{\Phi_2}$ over $(Y^{\Phi_1}, X)$. As such, the estimation can be done in two steps. First, take independent samples $\{x_i\}^n_{i=1}$ of $P_{\mathtt{x}}$ and draw independent samples $\{(e_{1,i,j},e_{2,i,j})\}^{n'}_{j=1}$ from $N^{(\mathtt{y})}_{\phi_{1,\mathtt{x}}(x_i),\phi_{2,\mathtt{x}}(x_i)}$. This enables us to generate samples from $\law{(X,Y^{\Phi_1},Y^{\Phi_2})}$ by push foward, that is, $\left\{ (x_i, \phi_{\mathtt{y}}\left(\widehat{\psi}_{\mathtt{y}}(e_{1,j} | \phi_{\mathtt{x}}(x_i)) \right), \phi_{\mathtt{y}}\left(\widehat{\psi}_{\mathtt{y}}(e_{2,j} | \phi_{\mathtt{x}}(x_i)) \right)) \right\}_{i \in [n],j \in [n']}$. Second, compute the conditional mean function of $Y^{\Phi_2} | (Y^{\Phi_1},X)$ using classical regression techniques with mean square error and evaluate the estimated regression function at $(Y^{\Phi_1},X)=(y,x)$. 

We compute such estimators in the numerical experiments presented below. For the sake of concision and clarity, \cref{sec:regression} purposely focused on concepts and methodologies rather than statistical guarantees. In future works, we could follow \cite{alquier2024universal} and \cite{dance2025counterfactual}, who minimize a very similar finite-sample loss and provide such guarantees, to prove the convergence of our estimators.

\section{Numerical experiments}\label{sec:exp}

This section illustrates the learning of a distributional regression function and its application to counterfactual reasoning on artificial and true datasets. The code is available at \url{https://github.com/lucasdelara/distributional_regression-counterfactuals}. The simulations were run on a personal computer.

\subsection{Toy example}

We start with a toy dataset inspired by the motivating example from \cref{sec:intro_example}. Let $P_{\mathtt{x,y}}$ be given by $P_{\mathtt{x}} := \Unif{[0,10]}$ and $P_{\mathtt{y} | \mathtt{x}}(\bullet | x) := \Lap(m(x),1)$, where $m(x) = 10 \sin(\frac{\pi x}{14})$, and $\G$ be defined by $\mathtt{x} \rightarrow \mathtt{y}$. The mean signal $m$ is the same as in the introduction but the output noise is now Laplacian instead of Gaussian. This serves to render the experiment more significant. If $P_{\mathtt{y} | \mathtt{x}}(\bullet | x)$ were $\Norm$, then $\psi^\C_{\mathtt{y}}(\bullet|x)$ would simply be a translation of parameter $m(x)$. With $P_{\mathtt{y} | \mathtt{x}}(\bullet | x)$ as $\Lap(m(x),1)$ we have instead $\psi^\C_{\mathtt{y}}(\bullet|x) := m(x) + G^{\Lap(0,1)} \circ F^{\Norm}$, which requires a more complex model to be estimated. We will rely on this closed-form expression to validate our method.

\begin{figure}[tb]
    \centering
    \begin{subfigure}[b]{0.4\textwidth}
        \centering
        \includegraphics[width=\linewidth]{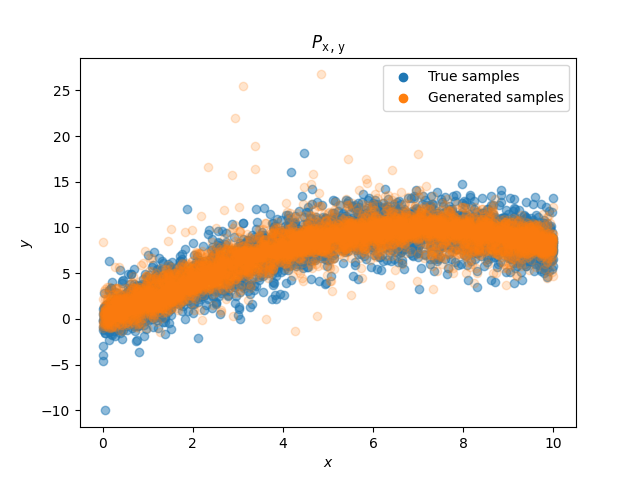}
        \caption{Global generation}
        \label{fig:toy_global_gen}
     \end{subfigure}
     \begin{subfigure}[b]{0.4\textwidth}
        \includegraphics[width=\linewidth]{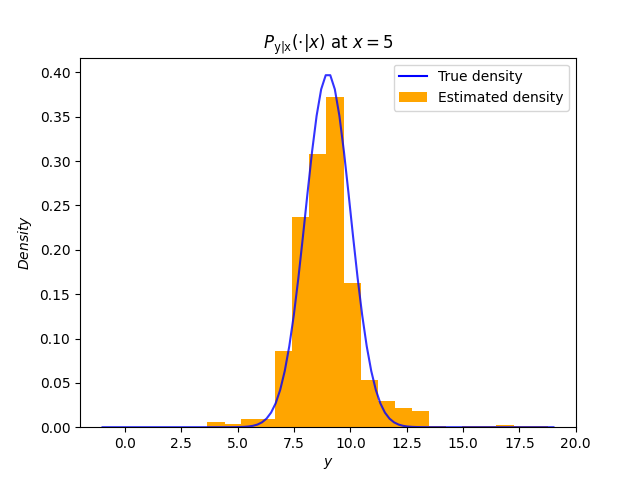}
        \caption{Conditional generation}
        \label{fig:toy_conditional_gen}
     \end{subfigure}
     \begin{subfigure}[b]{0.4\textwidth}
        \centering
        \includegraphics[width=\linewidth]{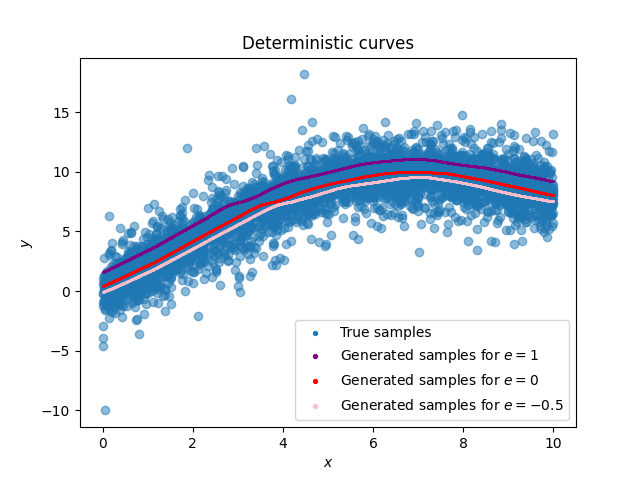}
        \caption{Fixed $e$, variable $x$}
        \label{fig:toy_curves}
     \end{subfigure}
     \begin{subfigure}[b]{0.4\textwidth}
        \includegraphics[width=\linewidth]{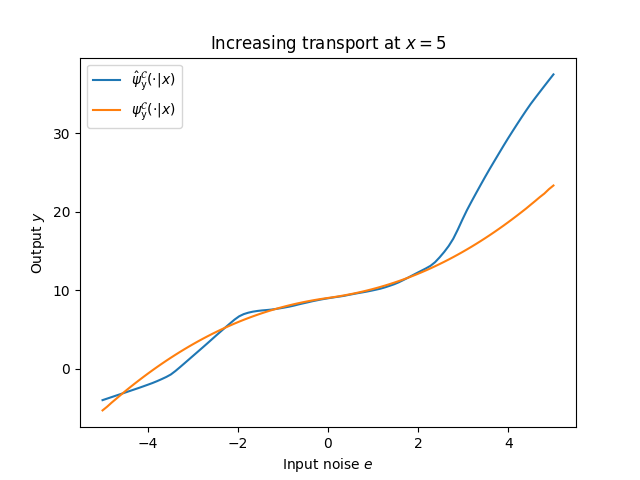}
        \caption{Fixed $x$, variable $e$}
        \label{fig:toy_transport}
     \end{subfigure}
    \caption{Validation of $\widehat{\psi}^\C_{\mathtt{y}}$ in the toy example.}
    \label{fig:toy_psi}
\end{figure}

In a first time, we estimate $\psi^\C_{\mathtt{y}}$. The model class corresponds to the architecture described in \cref{sec:models} with $L=4$ and $R=64$. The training procedure uses Adam optimizer with learning rate $10^{-4}$ and a batch size of $64$ during $60$ epochs. Both the training and testing phases employ $5000$ samples. The training leverages normalized data using a standard (Z-score) scaler. \Cref{fig:toy_psi} showcases several performances of $\widehat{\psi}^\C_{\mathtt{y}}$ on testing data. \Cref{fig:toy_global_gen} compares the true data $P_{\mathtt{x,y}}$ with the generated data $\widehat{\psi}^\C_{\mathtt{y}} \# (P_{\mathtt{e}} \otimes P_{\mathtt{x}})$ on samples. \Cref{fig:toy_conditional_gen} focuses on the generative ability at a single $x$ value. \Cref{fig:toy_conditional_gen} displays the graph of the function $\widehat{\psi}^\C_{\mathtt{y}}(e|\bullet)$ for several values of $e$. In particular, $e=0$ corresponds to the estimation of the conditional median. \Cref{fig:toy_transport} compares at a given $x$ the estimation $\widehat{\psi}^\C_{\mathtt{y}}(\bullet|x)$ with the true transport $\psi^\C_{\mathtt{y}}(\bullet|x) := m(x) + G^{\Lap(0,1)} \circ F^{\Norm}$. This highlights that the estimation procedure works. We note some outliers in the predictions, mostly positioned at high $\tty$ values.

\begin{figure}[tb]
    \centering
    \includegraphics[width=0.5\linewidth]{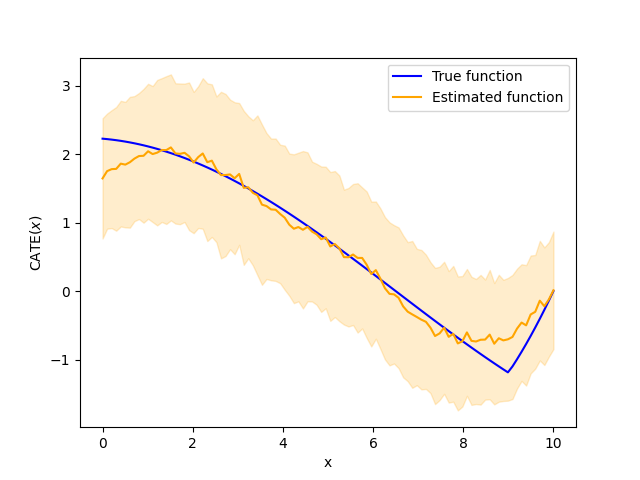}
    \caption{Graph of $\E[Y^\Phi-Y|X=x]$. The width of the error band corresponds to the standard deviations.}
    \label{fig:causal_effect}
\end{figure}

In a second time, we apply the learned model to estimate a general-causation effect. Using the notation from \cref{sec:causal_estimation}, for $\phi_{\ttx}(x) = \min(x+1,10)$ we have $\E[Y^\Phi-Y|X=x] = \E[\psi^\C_{\mathtt{y}}(E|\phi_{\ttx}(x))-\psi^\C_{\mathtt{y}}(E|x)] = m(\phi_{\ttx}(x))-m(x)$. We check the quality of the estimator based on $\widehat{\psi}^\C_{\tty}$ for a sample of size $n=1000$ from $\Norm$. \Cref{fig:causal_effect} depicts the mean values of the estimations along with their standard deviations. The mean function captures the trend of the true effect, except at the first border ($x=0$) and at the breaking point ($x=9)$. Additionally, the variance is quite large, which highlights the importance of taking the mean over a large sample from $P_{\texttt{e}}$ to obtain an accurate estimation.

\begin{figure}[tb]
    \centering
    \begin{subfigure}[b]{0.4\textwidth}
        \centering
        \includegraphics[width=\linewidth]{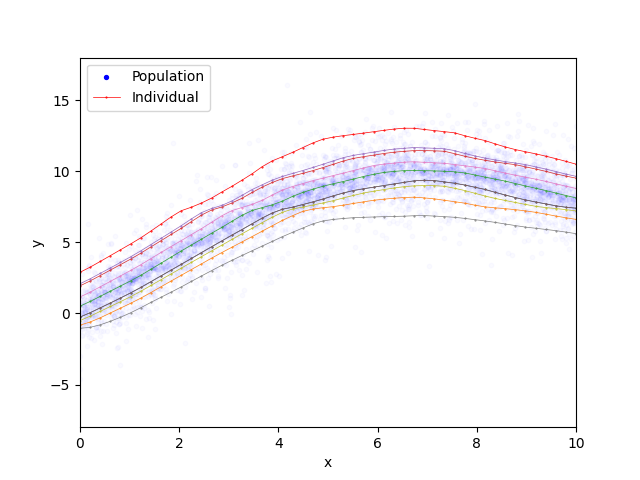}
        \caption{Individual curves, $N^\uparrow$}
        \label{fig:toy_comonotonic_indiv}
     \end{subfigure}
     \begin{subfigure}[b]{0.4\textwidth}
        \includegraphics[width=\linewidth]{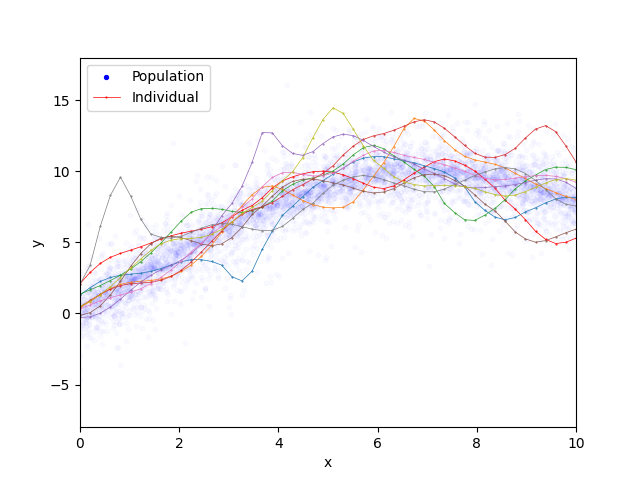}
        \caption{Individual curves, $N^k$}
        \label{fig:toy_gaussian_indiv}
     \end{subfigure}
     \begin{subfigure}[b]{0.4\textwidth}
        \centering
        \includegraphics[width=\linewidth]{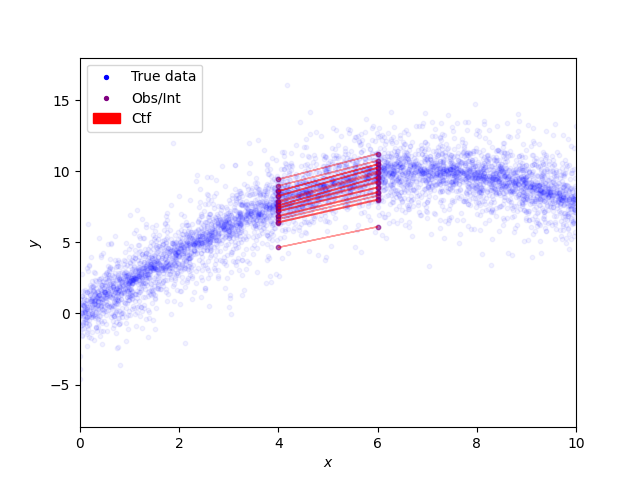}
        \caption{Coupling, $N^\uparrow$}
        \label{fig:toy_comonotonic_coupling}
     \end{subfigure}
     \begin{subfigure}[b]{0.4\textwidth}
        \includegraphics[width=\linewidth]{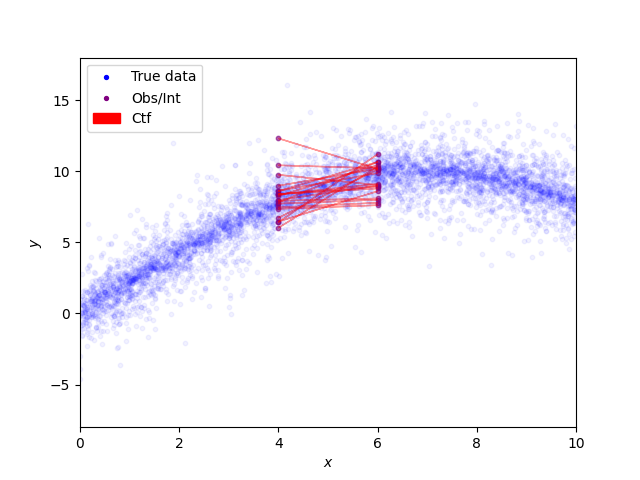}
        \caption{Coupling, $N^k$}
        \label{fig:toy_gaussian_coupling}
     \end{subfigure}
    \caption{Two counterfactuals conceptions in the toy example: $\widehat{\Psi}^\C_{\mathtt{y}} \# N^\uparrow$ (left) and $\widehat{\Psi}^\C_{\mathtt{y}} \# N^k$ (right).}
    \label{fig:toy_ctf}
\end{figure}

In a third time, we try several counterfactual conceptions. To do so, we specify two normalizations: $N^\uparrow$ and $N^k$, where $k$ is a Gaussian kernel with parameter $\sigma = 0.8$. \Cref{fig:toy_ctf} represents counterfactual information derived from these normalizations. \Cref{fig:toy_comonotonic_indiv} and \cref{fig:toy_gaussian_indiv} show 10 individuals across 50 parallel worlds. One individual corresponds to one sample from the normalization and to one color. \Cref{fig:toy_comonotonic_coupling} and \cref{fig:toy_gaussian_coupling} display the counterfactual coupling $\left( \widehat{\psi}^\C_{\mathtt{y}}(\bullet|4) \times \widehat{\psi}^\C_{\mathtt{y}}(\bullet|6) \right) \# N^{(\mathtt{y})}_{4,6}$ for each normalization. We emphasize that this step does not require any inference. We refer to the code to see how a user can simply choose a normalization once $\widehat{\psi}^\C_{\mathtt{y}}$ is trained. The figure illustrates how counterfactual conceptions describing pretty distinct forms of singular causation may coexist on top of a same causal graphical model.

Overall, the results we obtained on this toy example demonstrate that our method works and is perfectible. We underline that, while several aspects of the optimization procedure could certainly be improved to gain in computational time and prediction accuracy, our goal is not to make competitive simulations. Rather, we aim to show the principles of implementing and visualizing, via distributional regression, several counterfactual conceptions compatible with a same graphical model.

\subsection{Effect of 401(k) pension plan eligibility on
asset holdings}

Similar to \cite{dance2025counterfactual}, we apply our method on the real dataset studied by \cite{chernozhukov2004effects}. It contains 11 variables $\{\mathtt{z}_1, \ldots, \mathtt{z}_9, \mathtt{t}, \mathtt{y}\}$, where $\mathtt{t}$ is a binary indicator for eligibility to enroll in a 401(k) pension plan, the $\mathtt{z}_i$s are covariates, and $\mathtt{y}$ represent net financial assets. We denote by $P_{\mathtt{z,t,y}}$ the dataset's distribution. The objective consists in computing causal effects of $\mathtt{t}$ onto $\mathtt{y}$.

Interestingly, one does not need the complete knowledge of $\G$ to do so. The classical graphical assumption for this dataset is the following: the covariates are pretreatment variables. Formally, $\mathtt{t} \notin \pa(\mathtt{z}_i)$ for every $i \in [10]$, while $\mathtt{t} \in \pa(\mathtt{y})$. In particular, we do not specify relationships between covariates. As previously, we leverage the random-variable formalism from \cref{rem:variables}. Let $(Z,T,Y) \sim P_{\mathtt{z,t,y}}$ and $E^{(\tty)} \sim N^{(\tty)}$ be such that $(Z,T) \independent E^{(\tty)}$ and $Y = \psi^{\C}_{\tty}\left( E^{(\tty)}_{Z,T} | Z, T \right)$, where $N^{(\tty)}$ is an unspecified normalization (at this stage). Then, we define $Y_t := \psi^{\C}_{\tty}\left( E^{(\tty)}_{Z,t} | Z,t \right)$ for $t \in \{0,1\}$. The potential outcome $Y_t$ corresponds to the result of $\doint(\mathtt{t},t)$ in a structural representation of the counterfactual model. Crucially, $Z$ is not modified in the expression of $Y_t$ due to the graphical assumptions. This is why computing counterfactual distributions of $\mathtt{y}$ relative to changes of $\mathtt{t}$ only requires $\psi^\C_{\mathtt{y}}$: determining $\psi^\C_{\mathtt{z}_i}$ for some $i$ is useless.

We revisit the experiment of \cite{dance2025counterfactual}. To showcase the ability of their framework to carry out counterfactual inference, they estimate the causal effect $\E[Y_1 - Y_0 | Y_0=G_0(\bullet)]$, where $G_0$ is the quantile function of $Y_0$. Notice that this is a proper singular-causation effect, as it involves the joint probability distribution of potential outcomes. However, the authors do not mention that their model rest of the implicit assumption that counterfactuals are comonotonic, and hence do not emphasize enough that the curves they obtain are not universal. Different counterfactual assumptions would provide different curves than the ones in \citep[Figure 12]{dance2025counterfactual}. We propose to illustrate this point.

\begin{figure}[tb]
    \centering
    \begin{subfigure}[b]{0.47\textwidth}
        \centering
        \includegraphics[width=\linewidth]{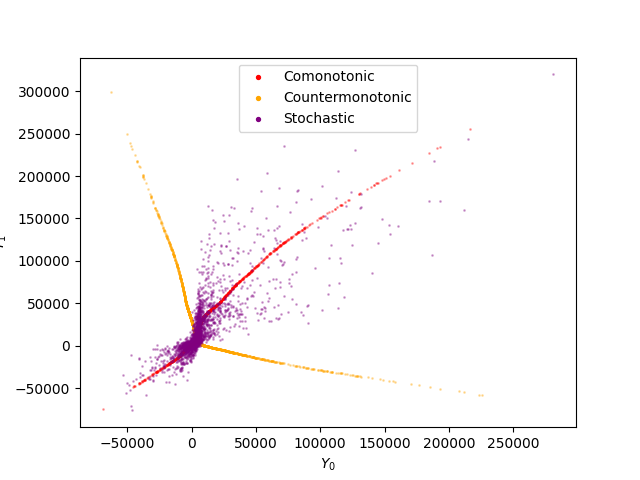}
        \caption{Samples from $\law{(Y_0,Y_1) | Z=\E[Z]}$}
        \label{fig:e401k_coupling_at_z}
     \end{subfigure}
     \hfill
     \begin{subfigure}[b]{0.47\textwidth}
        \includegraphics[width=\linewidth]{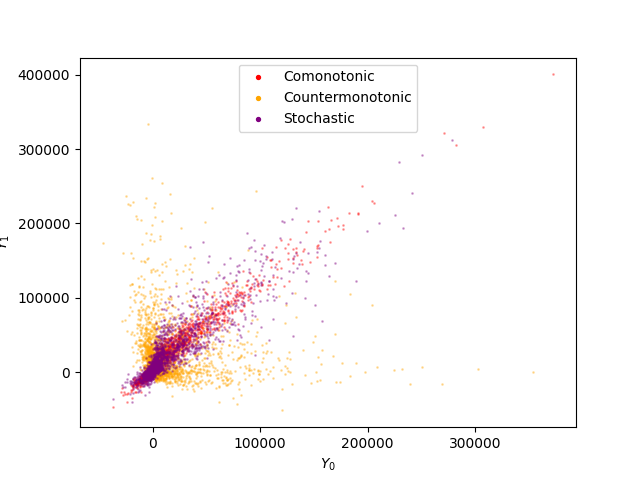}
        \caption{Samples from $\law{(Y_0,Y_1)}$}
        \label{fig:e401k_coupling}
     \end{subfigure}
    \caption{Counterfactual couplings for three different counterfactual conceptions.}
    \label{fig:e401k_ctf}
\end{figure}

In a first time, we estimate $\psi^\C_{\mathtt{y}}$. The model class corresponds to the architecture described in \cref{sec:models} with $L=2$ and $R=128$. The training procedure uses Adam optimizer with learning rate $10^{-3}$, a batch size of $128$, during $40$ epochs. Both the training phase employ $7000$ samples while the testing phase employs $2915$ samples. As before, the training leverages normalized data using a standard scaler. In a second time, we try three counterfactual conceptions: $N^\uparrow$, $N^\downarrow$, and $N^k$, where $k$ is a Gaussian kernel with bandwidth parameter $\sigma=2$ (hence, a stochastic comonotonic conception). The normalization specifies the joint probability distribution of $(E^{(\tty)}_{Z,0},E^{(\tty)}_{Z,1})$, which determines the joint probability distribution of $(Y_0,Y_1)$. In the case of $N^k$, the covariance of the Gaussian couple $(E^{(\tty)}_{Z,0},E^{(\tty)}_{Z,1})$ is given by
\begin{multline*}
\E[E^{(\tty)}_{Z,0} \cdot E^{(\tty)}_{Z,1}] = \E\left[\E[E^{(\tty)}_{Z,0} \cdot E^{(\tty)}_{Z,1} | Z] \right] = \E\left[\E[k\left( (Z,0), (Z,1) \right)|Z]\right] \\= \E\left[\E[\exp\left( -\norm{(Z,0)-(Z,1)}^2/\sigma \right)|Z]\right] = \exp\left( -1/\sigma \right)
\end{multline*}
Notably, due to the form of $k$ and the invariance of $Z$ by $\doint(\mathtt{t},t)$, the covariance has a simple expression. Before estimating the causal effect $\E[Y_1 - Y_0 | Y_0=G_0(\bullet)]$ for each counterfactual model, we visualize the three conceptions in \cref{fig:e401k_ctf}. More specifically, \cref{fig:e401k_coupling_at_z} illustrates how the properties carried by each normalization (deterministic comonotonic, deterministic countermonotonic, stochastic comonotonic) determine the \emph{one-step-ahead} counterfactual couplings.\footnote{Actually, there is no guarantee that $\E[Z]$ belongs to the support of $\law{Z}$. We merely use $\E[Z]$ as an arbitrary reference value for the conditioning on $Z$ in the illustration.} In contrast, \cref{fig:e401k_coupling} does not represent one-step-ahead counterfactual couplings, hence why even couplings generated by deterministic normalization are not deterministic. From a regression perspective, the noisiness of \cref{fig:e401k_coupling} compared to \cref{fig:e401k_coupling_at_z} comes from the exclusion of relevant explanatory variables: $Z$.

\begin{figure}
    \centering
    \includegraphics[width=0.9\linewidth]{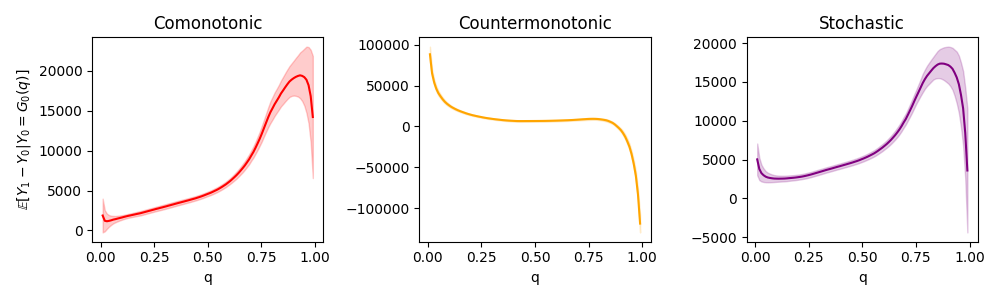}
    \caption{Average counterfactual treatment effect for three distinct counterfactual conceptions on the 401(k) dataset. The graphics represents the mean estimation along with the standard deviation over $10$ bootstrap repetitions.}
    \label{fig:e401k_quantile_effect}
\end{figure}

Noting that $\E[Y_1 - Y_0 | Y_0=G_0(\bullet)] = \E[Y_1 | Y_0=G_0(\bullet)] - G_0(\bullet)$, we then estimate $\E[Y_1 | Y_0=\bullet]$ by mean-square regression for each model. To do so, we rely on simple neural networks ($L=1$ and $R=128$). To account for variability, we repeat the whole training/evaluation procedure of the distributional regression function and of the conditional-expectation regression functions over 10 bootstrap replications of the dataset. \Cref{fig:e401k_quantile_effect} displays the results. We do obtain a similar curve profile to \cite{dance2025counterfactual} in the comonotonic scenario. As expected, the two other conceptions show different behaviors. Critically, one \emph{cannot conclude from data} that the causal effect of eligibility to the 401(k) scheme is largest for those with the largest financial assets among individuals who were not eligible, since the contrary trend happens in the countermonotonic scenario. Rather, one \emph{can believe} in such a trend.

\section{Related work}\label{sec:comparison}

To conclude this paper, we detail the similarities and differences between our framework and other related approaches to causal analysis and counterfactual modeling. This serves to make precise the scope of our contributions.

\subsection{The potential-outcome framework}\label{sec:po}

In the seminal causal approach of \cite{neyman1923applications} and \cite{rubin1974estimating}, the analyst does not have access to a model from which they could derive properties of a post-intervention variable like $V^{\Phi_w}_i$. Instead, they predetermine an outcome $Y : \Omega \to \Y$ together with a cause $T : \Omega \to \T$ of interest, and posit a collection of \emph{potential} outcomes $(Y_t)_{t \in \T}$ consistent with observational knowledge in the sense that $Y_T = Y$.\footnote{One may assume that the potential outcomes correspond to the post-intervention variables of some SCM. However, such a connection to do-interventions on SCMs is not necessary, as reminded in \citep{delara2025clarification}.} Note that $(Y_t)_{t \in \T}$ needs to be measurable, which does not always hold when $\T$ is uncountable. To carry out causal inference, which amounts to computing probabilistic features of $(Y_t)_{t \in \T}$ from observational data, they then make assumptions generally framed as conditional-independence constraints. For instance, \emph{conditional ignorability} relatively to a tuple of covariates $X : \Omega \to \X$ means that $(Y_t)_{t \in \T} \independent T | X$ and implies that $\law{Y_t | X=x} = \law{Y | X=x, T=t}$ for every $t \in \T$ provided that $\P(T=t | X=x) > 0$.

Some refined versions of this framework borrow elements from Pearl's framework to deal with more possible causes and outcomes, as well as to read off conditional-independence constraints from a graph. Notably, \cite{richardson2013single} postulate a graph $\G$ and one-step-ahead counterfactual random processes $\left( V^{\doint(\pa(i),v_{\pa(i)})}_i \right)_{v_{\pa(i)} \in \V_{\pa(i)}}$, for every $i \in [d]$. This potential-outcome approach and the canonical-representation approach are analogous: they both focus on a random-process perspective without explicitly relying on functional relationships. Additionally, note that some demonstrations of the equivalence between the potential-outcome framework and the structural framework (see \citep[Section 7]{pearl2009causality} and \citep{ibeling2023comparing}) share similarities with the proofs of \cref{prop:entailed_a} and \cref{thm:representation} \citep{shpitser2022multivariate, delara2025clarification}.

Crucially, the fundamental difference between our approach and potential outcomes does not concern how they formally represent counterfactuals, but the modeling and inference purposes of these representations. In practice, potential-outcome models serve to study general causation rather than singular causation, and as such are more comparable to CGMs than SCMs and counterfactual models. Notably, none of their typical assumptions (like conditional ignorability) identify \emph{joint} probability distributions of one-step-ahead counterfactuals; they simply \emph{constrain} this law with conditional-independence assumptions. In contrast, counterfactual models \emph{directly fully specify} the law $S^{(i)}$ of the whole random process $\left( V^{\doint(\pa(i),v_{\pa(i)})}_i \right)_{v_{\pa(i)} \in \V_{\pa(i)}}$. Note that this is not a critique of the potential-outcome framework: only a remark on its practical scope. We emphasize that the underspecification of potential-outcome models and causal graphical models for singular causation is arguably a strength in some cases, since it avoids introducing superfluous hypotheses in the context of general causation \citep{dawid2000causal,richardson2013single}.

\subsection{Joint-distribution frameworks}

We advance to frameworks that avoid using a fully specified SCM to describe counterfactual distributions. The general idea of counterfactual models is to directly design the joint counterfactual distributions rather than specifying an SCM. In this sense, any approach to model dependencies between preassigned marginals is a natural option for counterfactual reasoning. This is notably the case of copulas and optimal transport. There has actually been a surge in applications of these two frameworks in causality in general---not uniquely in counterfactual inference. As such, some aligns with our work; other greatly differ. We propose hereafter a brief review of this literature, in order to clarify the scope of our contributions.

\subsubsection{Copulas}

The idea of modeling joint probability distributions independently of their marginals, on which we heavily rely, appeared in several causal-inference papers. While these references all employ the term \say{counterfactual}, they sometimes address distinct problems: some concern general causation, other singular causation.

For example, \cite{lai2024counterfactual} and \cite{pereda2024estimation} model \emph{separately} the joint observational distribution and a joint interventional distribution with copulas called, respectively, the \emph{actual} copula and \emph{counterfactual copula}. Their goal amounts to learning the counterfactual copula from observational data. As such, their counterfactual copula capture cross dependencies between variables within same worlds, not between variables across different worlds (the actual and the counterfactual) as we do.

Other papers do address proper counterfactual inference. \cite{haugh2023counterfactual} rely on copulas to model different counterfactual distributions in a discrete dynamic latent state model. However, they focus on a particular setting, where the copula serves to encode the cross-dependencies of exogenous variables across worlds. Our setting differ, as the exogenous vector $U$ remains the same in all worlds. \cite{gunsilius2025primer} uses a copula view to represent the couplings $\law{((U_i,V_i) | V_{\pa(i)} = v_{\pa(i)}}$ for every $v_{\pa(i)} \in \V_{\pa(i)}$ and through it relevantly mentions that comonotonicity is a restrictive choice of counterfactuals. We take a similar perspective when modeling counterfactuals, but we design the cross-dependencies directly between $\law{V_i | V_{\pa(i)} = v_{\pa(i)}}$ for every $v_{\pa(i)} \in \V_{\pa(i)}$. This notably avoids to involve the exogenous noise $U_i$, which can be reparametrized without loss of generality even at the counterfactual level (\cref{prop:concise}). Overall, to our knowledge, copulas have never been used to design (one-step-ahead) counterfactual distributions.

\subsubsection{Optimal transport}

Optimal transport theory is a mass-transportation framework which addresses optimization problems of the form
\begin{equation}\label{eq:ot}
   \OT_c(P,P') = \min_{\gamma \in \Gamma(P,P')} \iint c(y,y') \mathrm{d}\gamma(y,y'),
\end{equation}
where $P,P' \in \Prob{\Y}$, $\Gamma(P,P')$ is the set of couplings from $P$ to $P'$ and $c : \Y \times \Y \to \R_+$ is a cost function. Optimal-transport analysts are interested in either the \emph{minimum value} $\OT_c(P,P')$ returned by this program or by the \emph{minimizers} of this program. Interestingly, for $c$ a distance over $\Y$ and $p \geq 1$ an integer, the function $\OT^{1/p}_{c^p}$ defines a distance over $\Prob{\Y}$ called a \emph{Wasserstein distance}, while a minimizer $\gamma^*$ provides an optimal coupling (or optimal transport plan) between the marginals. Recently, contributions at the intersection between optimal transport and causal inference have increased \citep{gunsilius2025primer}. Similar to copula-based causality papers, they all employ similar keywords but often address radically different problems. We propose to classify some of these papers and to clarify their relations to our framework. The classification rests on two fundamental questions. Regarding optimal transport, is the quantity of interest the optimal value or the optimal transport coupling? Regarding causal inference, what is the considered layer of causation?

\cite{cheridito2025optimal} adapt the optimal transport problem to marginals and couplings that factorize according to a same graph $\G$. This provides sharper notions of Wasserstein distances according to which average treatment effects are notably continuous. Thus, they leverage the minimum value of a refined version of \cref{eq:ot} to address \emph{general causation}. In this sense, it deviates from our framework. Nevertheless, we notice that the factorization constraint they impose on couplings coincides with the joint Markov property (\cref{prop:block_markov}). As such, while they do not ultimately apply the feasible couplings of their optimal transport problem, these couplings are valid \emph{counterfactual} couplings (in the sense that they correspond to a Markovian SCM with graph $\G$).

\cite{lin2025tightening} follow a similar approach but in the graph-free potential-outcome framework. Noting that $\law{(Y_0,X)}$ and $\law{(Y_1,X)}$ are identifiable under conditional ignorability relatively to $X$, they adapt \cref{eq:ot} to couplings between these marginals. Then, they use the optimal value to derive partial identification bounds on causal effects based on $\law{(Y_0,Y_1)}$. In their case, the optimal-transport coupling itself is not used and no counterfactual conception is specified. Instead, the optimal value serves to quantify the uncertainty steaming from the various possible counterfactual conceptions. Again, this does not deal with the full specification of singular causation.

\cite{black2020fliptest}, \cite{charpentier2023optimal}, \cite{torous2024optimal}, and \cite{delara2024transport} define \emph{counterfactual couplings} as optimal transport plans (or maps) between fixed interventional marginals. Said differently, they apply the minimizers of \cref{eq:ot} to address \emph{singular causation}. Their objective clearly aligns with ours. Actually, our paper is meant to extend their works to \emph{multi-marginals} joint probability distributions and to account for graphical knowledge. Since this deserves further precisions, we reserve the thorough details of this point to the section below.

\subsection{Mass-transportation approaches}\label{sec:transport}

Recent developments in causal inference have focused on a mass-transportation viewpoint of counterfactual reasoning \citep{black2020fliptest, charpentier2023optimal,torous2024optimal,delara2024transport,machado2025sequential,dance2025counterfactual}. It is based on the fact that an intervention not only produces an interventional distribution but also a matching between two probability distributions. Studying paired instances across marginals, rather than only the marginals, exactly corresponds to studying singular causation. As such, our work closely follows this line of research that we present in more details.


\subsubsection{Transport-based counterfactual models}

In \citep{delara2024transport}, which in inspired by \citep{black2020fliptest}, the authors propose to circumvent the full specification of an SCM by directly placing couplings between every pair of \emph{multivariate} marginals $P^{\Phi}$ and $P^{\Phi'}$.\footnote{In fact, they address a specific setting where the considered interventional marginals are observational, but this can be generalized to the case where the marginals are identifiable (for instance by a CGM).} They call such a collection a \emph{transport-based counterfactual model}. Given two marginals, the chosen coupling characterizes the semantics of the counterfactual statements. It can notably be deterministic or stochastic. Then, they further study in theory and in practice the case where the coupling corresponds to the solution of the quadratic optimal-transport problem between the marginals.

Canonical representations clearly follow a similar principle, as they define counterfactuals by placing joint probability distributions over preassigned marginals. Crucially, they address two limitations of transport-based counterfactual models in the context of equivalence to SCMs: the absence of causal ordering constraints (as noted by \cite{machado2025sequential}) and the absence of algebraic constraints (as noted by \cite{dance2025counterfactual}). The solutions that canonical representations introduce to address these limitations build upon ideas from the work of \cite{dance2025counterfactual}, that we present next.

\subsubsection{Counterfactual cocycles}

Our work borrows and extends many ideas of \cite{dance2025counterfactual} (and from \citep{dance2024causal}, a previous version of their article). For the sake of simplicity, we will only present the main concepts and refer to their paper for more details. Their framework for counterfactual inference relies on objects called \emph{cocycles}, defined relatively to probability kernels. In the context of causal kernels, given a CGM $\C := (\V,\G,\K)$ and a variable $i \in [d]$, a cocycle $s^{(i)}$ basically specifies how an intervention on $\pa(i)$ (the inputs) transforms $i$ (the output). More precisely, a cocycle $s^{(i)}$ for the $i$th variable determines a collection of measurable mappings $\{ s^{(i)}_{v_{\pa(i)},v'_{\pa(i)}} \}_{v_{\pa(i)},v'_{\pa(i)} \in \V_{\pa(i)}}$, called \emph{counterfactual maps}, such that for every $v_{\pa(i)},v'_{\pa(i)},v''_{\pa(i)} \in \V_{\pa(i)}$:
\begin{align}
    &K_i(\bullet | v'_{\pa(i)}) = s^{(i)}_{v_{\pa(i)},v'_{\pa(i)}} \# K_i(\bullet | v_{\pa(i)}),\label{eq:transport}\\
    &s^{(i)}_{v_{\pa(i)},v''_{\pa(i)}} = s^{(i)}_{v'_{\pa(i)},v''_{\pa(i)}} \circ s^{(i)}_{v_{\pa(i)},v'_{\pa(i)}}.\label{eq:composition}
\end{align}
\Cref{eq:transport} is a mass-transportation constraint, called \emph{kernel adaptedness}, which signifies that a counterfactual map provides a deterministic transport plan between every two states of the causal kernel. \Cref{eq:composition} is an algebraic constraint, called \emph{path independence}, which means that only the initial and final states matter. Informally, a cocycle $s^{(i)}$ resembles a one-step-ahead counterfactual process $S^{(i)}$, as they both specify cross-dependencies between interventional marginals. We purposely used a similar notation. However, cocycles model deterministic dependencies, and thereby correspond to deterministic counterfactuals only. In what follows, we formally prove these points to illustrate that one-step-ahead counterfactual distributions generalize cocycles to the stochastic case.

More precisely, we provide two theoretical results on general process measures, and then analyze their implications on causal cocycles and one-step-ahead counterfactual processes. In a first time, we show that a process measure is deterministic if and only if all its entailed couplings are deterministic. We point out that this result is not trivial in the \say{if} sense, as it demands to deduce a property of a whole joint probability distribution from some of its marginalizations. Interestingly, it renders the notion of deterministic \emph{coupling} and the notion of deterministic \emph{process} measure consistent.

\begin{proposition}[Characterization of deterministic couplings and processes]\label{prop:deterministic_process}
Let $\Y,\Y'$ be Borel spaces and $\X$ be a set. The two following propositions hold.
\begin{enumerate}
    \item Let $P \in \Prob{\Y}$ and $P' \in \Prob{\Y'}$. For every coupling $\gamma$ between $P$ and $P'$, $\gamma$ is deterministic (from left to right and right to left) if and only if there exists an injective measurable map $T : \Y \to \Y'$ such that $\gamma = (\Id,T) \# P = (T^{-1},\Id) \# P'$.
    \item For every $S \in \Q(\X,\Y)$, $S$ is deterministic if and only if $S_{x,x'}$ is a deterministic coupling for every $x,x' \in \X$.
\end{enumerate}
\end{proposition}

A \emph{single} deterministic coupling is characterized by a transport map between its marginals. The result below shows that a \emph{family} of deterministic couplings is stable under composition of their transport maps if and only if this family derives from a deterministic process measure.

\begin{proposition}[Composition constraint for stochastic couplings]\label{prop:composition}
Let $\Y$ be a Borel space, $\X$ a set, and $(P^{(x)})_{x \in \X}$ be a family of probability distributions in $\mathcal{P}(\Y)$. The two following items hold.
\begin{enumerate}
    \item Let $(s_{x,x'})_{x,x' \in \X}$ be a collection of measurable maps from $\Y$ to $\Y$. If for every $x,x',x'' \in \X$,
    \begin{itemize}
        \item[(i)] $s_{x, x''} = s_{x', x''} \circ s_{x, x'}$ and 
        \item [(ii)] ${s_{x,x'}} \# P^{(x)} = P^{(x')}$,
    \end{itemize}
    then there exists $S \in \Q_{\text{joint}}(\X,\Y,(P^{(x)})_{x \in \X})$ such that $S_{x,x'} = (\Id, s_{x, x'}) \# P^{(x)}$ for every $x,x' \in \X$. In particular, $S$ is deterministic.
    \item Let $S \in \Q_{\text{joint}}(\X,\Y,(P^{(x)})_{x \in \X})$. If $S$ is deterministic then there exists a collection of measurable maps $(s_{x, x'})_{x,x' \in \X}$ such that for every $x,x',x'' \in \X$,
    \begin{itemize}
        \item[(a)] $s_{x, x''} = s_{x', x''} \circ s_{x, x'}$ and
        \item[(b)] ${s_{x, x'}} \# P^{(x)} = P^{(x')}$ and
        \item[(c)] $S_{x, x'} = (\Id, s_{x,x'}) \# P^{(x)}$.
    \end{itemize}
\end{enumerate}
\end{proposition}
Basically, the first item states that a collection of \emph{deterministic} couplings between every pair of a given family of marginals derives from a \emph{same} process measure if the couplings (more precisely, their associated transport maps) satisfy a \emph{composition} constraint (namely, path independence). Conversely, the second item signifies that if a process measure is deterministic, then the derived couplings satisfy path independence. In other words, being a projective family of distributions extends to the stochastic case the path-independence constraint for a family of deterministic couplings. Note that this sort of composition concerns couplings across fixed marginals and as such fundamentally differs from composability, which concerns using outputs of a random process as input indices of another one.

To sum-up, according to \cref{prop:deterministic_process} and \cref{prop:composition}, one-step-ahead counterfactual processes generalize to the stochastic case the fundamental properties \cref{eq:transport} and \cref{eq:composition} of cocycles. Therefore, our framework generalizes the one of \cite{dance2025counterfactual}. This extension is significant for notably two reasons. First, as aforementioned, deterministic joint probability distributions do not always exist for some families of marginals. Crucially, our approach allows counterfactual reasoning given any CGM---including CGMs that do not admit deterministic counterfactuals. Second, even with CGMs compatible with deterministic counterfactuals, an analyst could aim to implement a stochastic conception of counterfactuals. Using our approach, they can model a richer class of counterfactuals than solely deterministic ones.

Moreover, as previously mentioned, \cite{dance2025counterfactual} addressed several limitations of the transport-based framework of \cite{delara2024transport}. By broadening the scope of \citep{dance2025counterfactual}, our framework addresses these limitations in the general stochastic setting. Let us further detail this point to explain how we improve the contributions in \citep{delara2024transport}. The main limitation of transport-based counterfactual models concerns the question of equivalence to SCMs. In Markovian SCMs, the causal ordering from the graph imposes factorization constraints on the counterfactual distributions (remind \cref{prop:block_markov}). Therefore, not all couplings correspond to some Markovian SCM.\footnote{Nonetheless, any joint probability distribution fits a non-Markovian SCM \citep[Proposition 5]{delara2025clarification}.} By not satisfying these constraints by design, transport-based counterfactual couplings do not necessarily align with SCM-based counterfactual couplings. On the contrary, \cite{dance2025counterfactual} can construct deterministic couplings between two multivariate interventional marginals $P^\Phi$ and $P^{\Phi'}$ by defining unitary transport maps between their Markov factors. This forces the counterfactual couplings to respect the causal ordering, and thereby to correspond to a Markovian SCM. Our construction of counterfactual models follows the same principle with stochastic couplings rather than transport maps. Additionally, in any SCM (be it Markovian or not), counterfactual couplings respect a kind of \say{composition} property, which follows from the \emph{consistency rule} of counterfactuals \citep{pearl2010consistency}. It exactly corresponds to the algebraic constraint of cocycles \cref{eq:composition} in the deterministic case. By not imposing this composition constraint to its couplings, the transport-based approach fails to ensure systematic equivalence with SCMs. Counterfactual models extend this constraint to the general case, as demonstrated above, and therefore are equivalent to Markovian SCMs, as shown by \cref{prop:entailed_a} and \cref{thm:representation}.

To conclude this discussion on \citep{dance2025counterfactual}, we point out other advantages of our framework, that are not related to the generalization from deterministic to stochastic. Cocycles do not separate counterfactual modeling from compatibility to the CGM, and hence suffer from the same issues as SCMs: testing several counterfactual conceptions requires learning a new model each time; our approach based on normalizations requires learning one model. Furthermore, similar to a structural assignment, the form of a cocycle does not matter for general causation (as long as it fits the causal kernel) but fully determines the one-step-ahead counterfactual conception. However, the discussion from \citep[Section 7]{dance2025counterfactual}, where the authors mention setting the model class, mostly revolves around a trade-off between interpretability of the model and adaptivity to complex data. Critically, there is no mention to the question of counterfactual choices. All in all, their method is a mass-transportation alternative to bijective SCMs, which aims at providing more robust estimations of causal effects. Crucially, it does not intend to allow analysts to select a counterfactual conception compatible with a given SCM. This is a fundamental distinction to our work, also shared by the thread of research we detail hereafter.


\subsection{Estimation of causal effects via conditional generative models}

Recent papers have proposed, given an observational distribution and a directed graph, to estimate causal queries across the causality ladder by using conditional generative models \citep{xia2022neural,chao2023interventional,rahman2024modular,rahman2024conditional}. The general principle consists in learning, for each $i \in [d]$, a generative model for $P_{i|\pa(i)}$, and then to leverage these modular models in conjunction along the topological order to draw samples from causal distributions consistent with the given CGM. Our approach based on normalizations follows the same principle. On the basis of a known CGM, one approximates with a neural network the conditional generator $\psi^\C_i$ (\cref{sec:implementation}) for each $i \in [d]$, to then compute various causal effects (\cref{sec:causal_estimation}). However, our general objective as well as our practical implementation differ from these contributions at a crucial regard.

Similar to \cite{dance2025counterfactual} and the structural approach, none of them tackle the question of \emph{choosing} a notion of singular causation compatible with a given CGM. They all propose to replace the causal mechanism $(f_i,P_{0,i})$ by some generative model, like a generative adversarial network \citep{xia2022neural}, a diffusion model \cite{chao2023interventional}, a deep causal generative models \citep{rahman2024modular}, or an arbitrary conditional generative model \citep{rahman2024conditional}. This provides a variant to an SCM, which determines a whole ladder of causation and may be more efficient than the structural framework to estimate causal effects. Critically, this approach does not allow to run across many counterfactual conceptions and the authors do not discuss the relation between the model's construction and the determination of counterfactuals. This seems consistent with the fact that, even when considering counterfactual effects, these papers mostly focus on \emph{general}-causation queries.\footnote{Note that they have several advantages \emph{at the general-causation level} compared to our approach, such as handling latent confounders (non-Markovian models) and more complex data.} In our case, the collection of conditional generative models $(\psi^\C_i)^d_{i=1}$ plays a fundamentally different role. It is meant to correspond to the CGM---not to an SCM. Notably, compared to the generative models proposed above, it is uniquely determined by the CGM. With this collection alone, one can estimate all effects identifiable in the CGM. Then, by specifying the distribution of an input random process, one can choose any compatible counterfactual distribution.

To conclude, we caution against misunderstanding expressions like \say{counterfactual inference} and \say{counterfactual identifiability}. One cannot deduce singular-level counterfactuals from observations and interventions alone. A proper counterfactual effect is not identifiable by a CGM. Actually, more complex models build on top of a CGM may determine all counterfactual laws. But the identifiability of a causal model from data differs from the identifiability of a law from a causal model. In particular, \emph{being able to draw samples from a counterfactual distribution radically does not mean being able to draw samples from a deliberately chosen counterfactual distribution}. Many of the causal approaches that we discussed (structure-based, cocycle-based, generative-model-based), which extend a given CGM, do not constrain the specification of the counterfactuals, leaving it unaccounted. To our knowledge, our framework is the first attempt to allow analysts to conveniently choose admissible counterfactuals, by disentangling general causation from singular causation in the construction of causal models.

\section{Conclusion}

We proposed a framework for modeling and implementing counterfactual assumptions compatible with a given causal graphical model and proved its formal equivalence to the structural framework. From a pedagogical viewpoint, this equivalence highlights the specific role of singular causation within the counterfactual layer of the causality ladder. To summarize, singular causation formally corresponds to the specification of functional relationships in Pearl's causal models and to the specification of process measures in their canonical representations. Critically, it can never be falsified and thereby rests on a choice that must be made explicit and legitimated. From a modeling viewpoint, our framework permits domain experts to intelligibly formalize their choices of counterfactual assumptions. We introduced normalizations to disentangle observational and graphical knowledge from counterfactual conceptions, which allows to parametrize counterfactuals in a more convenient way than structural causal models. From a practical viewpoint, it enables causal-inference analysts to withdraw counterfactual choices from the inference step and to test effortlessly many counterfactual conceptions, as illustrated by our numerical experiments. To our knowledge, this is the first approach well separating general causation (possibly estimable with data) from singular causation (unfalsifiable) in causal modeling. Moreover, developing this framework led us to contribute to the theory and practice of distributional regression and joint probability distributions.

We emphasize that the goal of this paper is not to replace the structural approach or other causal paradigms. Each type of causal models shows advantages and drawbacks depending of the context. Notably, SCM are better tailored to frame causal-discovery assumptions and to express intelligible cause-effect numerical relationships. Also, our contribution is simply a primer on canonical representations, which deserves future developments. Essentially, we hope that our work will encourage changes of practice in the causality literature---regardless of the employed causal framework. Concretely, we call researchers and practitioners to systematically clarify if their studies concern general causation or singular causation. In case it concerns singular causation, we ask them to clearly specify the counterfactual conception and to explain this choice. In particular, in the structural approach, this requires motivating with conceptual arguments the form of the structural equations.

This work also opens further research directions. A natural extension would be to address the non-Markovian setting. While Markovianity is frequently assumed for convenience in the causality literature, and was particularly useful to keep this paper as simple as possible, it is generally deemed too stringent in real-world scenarios. Finally, we believe that the distributional regression function we propose can have promising applications beyond causal reasoning.


\acks{The author thanks Marianne Clausel for her valuable feedback. The author also thanks Iain Henderson and Mathis Deronzier for helpful discussions on the measurability of random processes. This work was funded by the project CAUSALI-T-AI (ANR-23-PEIA-0007) of the French National Research Agency. }


\newpage

\appendix

\section{Proofs}\label{sec:proofs}

\subsection{Proofs of \cref{sec:setup}}

\begin{proof}[\Cref{prop:cgm_obs}]
Since $\G$ is acyclic, it induces a topological order on $[d]$. We can suppose without loss of generality that $[d]$ is sorted according to this order, so that $\pa(i) \subseteq [i-1]$ for every $i \in [d]$. Then, by artificially extending the second arguments of the kernels, we can write $K_i(\bullet|v_{\pa(i)}) = K'_i(\bullet|v_{[i-1]})$ for a probability kernel $K'_i$ from $\V_{[i-1]}$ to $\V_i$. As such, $P$ is defined as the product of $(K'_i)^d_{i=1}$. According to the Ionescu-Tulcea theorem \citep[Theorem 8.24]{kallenberg1997foundations}, such a product of nested probability kernels produces a unique probability measure. Finally, the equality $P_{i | \pa(i)}(\bullet | v_{\pa(i)}) = K_i(\bullet | v_{\pa(i)})$ follows by disintegration.
\end{proof}

\begin{proof}[\Cref{prop:cgm_exists}]
Define the graph $\G$ given by $\pa(1) = \emptyset$ and $\pa(i) = [i-1]$ for $i \geq 2$. Then, remark that
\begin{align*}
\mathrm{d}P(v_1,\ldots,v_d) &= \prod^d_{i=1} \mathrm{d} P_{i | [i-1]}(v_i | v_{[i-1]})\ \text{by successive disintegrations,}\\
&= \prod^d_{i=1} \mathrm{d} P_{i | \pa(i)}(v_i | v_{\pa(i)})\ \text{by definition of $\G$}.
\end{align*}
Therefore, choosing $K_i := P_{i | \pa(i)}$ defines $\C := (\V, \G, \K)$ such that $\Obs{C}(\C)=P$.
\end{proof}

\begin{proof}[\Cref{lem:markov}]
First, we show the existence of a solution. Because $\G$ is a DAG, it induces a topological order on $[d]$. Up to a permutation, we can assume without loss of generality that $[d]$ is sorted according to this topological order. Then, consider any random vector $(U_i)^d_{i=1}$ with law $P_0$, and define each $V_i$ recursively as $V_i := f_i(V_{\pa(i)},U_i)$. By construction $(V,U)$ is a solution of $\M$. Hereafter, we address each property in order.

\emph{Property n°1.} Remind that $V_j := f_j(V_{\pa(j)},U_j)$ for every $j \in [d]$ and note that by induction $V_j = g_j(U_{\an(j)})$ for some measurable function $g_j$. This shows that $V_i \independent U_{\nan(i)}$ by mutual independence of $(U_j)^d_{j=1}$  Moreover, $V_{\ndesc(i)}$ can be expressed as a measurable map of $U_{[d] \setminus \{i\}}$. It then follows from $U_i \independent U_{[d] \setminus \{i\}}$ that $U_i \independent V_{\ndesc(i)}$.

\emph{Property n°2.} It is a direct consequence of $V_i = f_i(V_{\pa(i)},U_i)$ and $U_i \independent V_{\pa(i)}$.

\emph{Property n°3.} The set $\ndesc(i)$ can be written as $\ndesc(i) = \pa(i) \sqcup \left( \ndesc(i) \setminus \pa(i) \right)$. Therefore, up to a permutation of the components, $V_{\ndesc(i)} = (V_{\pa(i)},V_{\ndesc(i) \setminus \pa(i)})$. Moreover, by induction, one can write $V_{\ndesc(i) \setminus \pa(i)} = \varphi_i(U_{H_i})$ for some measurable function $\varphi_i$, where $H_i \subseteq \nan(i)$. Therefore, for every $v_{\pa(i)} \in \V_{\pa(i)}$,
\begin{align*}
    &\law{(V_i, V_{\ndesc(i)}) | V_{\pa(i)}=v_{\pa(i)}} = \law{f_i(v_{\pa(i)},U_i), (v_{\pa(i)},\varphi_i(U_{H_i})) | V_{\pa(i)}=v_{\pa(i)}} \ \text{by expressions,}\\
    &= \law{f_i(v_{\pa(i)},U_i), (v_{\pa(i)},\varphi_i(U_{H_i}))} \ \text{by property n°1,}\\
    &= \law{f_i(v_{\pa(i)},U_i)} \otimes \law{(v_{\pa(i)},\varphi_i(U_{H_i}))} \ \text{since $i \notin H_i$,}\\
    &= \law{f_i(v_{\pa(i)},U_i) | V_{\pa(i)}=v_{\pa(i)}} \otimes \law{(v_{\pa(i)},\varphi_i(U_{H_i})) | V_{\pa(i)}=v_{\pa(i)}} \ \text{by property n°1,}\\
    &= \law{V_i | V_{\pa(i)}=v_{\pa(i)}} \otimes \law{ V_{\ndesc(i)} | V_{\pa(i)}=v_{\pa(i)}} \ \text{by expressions.}
\end{align*}
This concludes the proof.
\end{proof}

\begin{proof}[\Cref{prop:scm_obs}]
According to the proof of \cref{lem:markov}, there exists a collection of measurable maps $(g_i)^d_{i=1}$ such that $V_i = g_i(U_{\an(i)})$ for every $i \in [d]$ and every solution $(V,U)$. Consequently, the measurable map $g : \U \to \V, u \mapsto (g_i(u_{\an(i)}))^d_{i=1}$ is such that $V = g(U)$ for every solution $(V,U)$. Therefore, $\operatorname{Obs}(\M) := \mathbb{L}(V) = g \# \mathbb{L}(U) = g \# P_0$ for every solution $(V,U)$.

We call $g$ the solution map of $\M$. This notion will play a role in the proof of \cref{prop:block_markov}.
\end{proof}

\begin{proof}[\Cref{prop:scm_cgm}]
Let $\C_\M$ be defined as above. Since for every $i \in [d]$ the output space of $f_i$ is $\V_i$, the kernel $K_i$ is defined on $\B(\V_i) \times \V_{\pa(i)}$. Therefore, $\C_\M \in \CGM_{\V}$. We now prove each item in order.

\emph{Property 1.} We write $P := \Obs{M}(\M)$, and suppose without loss of generality that the variables are in topological order so that $[i-1] = \pa(i) \sqcup H_i$, where $H_i \subseteq \ndesc(i)$. Then, $P$ factorizes as:
\begin{align*}
\mathrm{d}P(v_1,\ldots,v_d) &= \mathrm{d} \P(V_1 = v_1, \ldots, V_d = v_d)\ \text{by definition of $P$,}\\
&= \prod^d_{i=1} \mathrm{d} \P(V_i = v_i | V_{[i-1]}=v_{[i-1]})\ \text{by the extended Bayes' formula,}\\
&= \prod^d_{i=1} \mathrm{d} \P(V_i = v_i | V_{\pa(i)}=v_{\pa(i)}, V_{H_i}=v_{H_i})\ \text{by definition of $H_i$,}\\
&= \prod^d_{i=1} \mathrm{d} \P(V_i = v_i | V_{\pa(i)}=v_{\pa(i)})\ \text{by \cref{lem:markov},}\\
&= \prod^d_{i=1} \mathrm{d} \P(f_i(v_{\pa(i)},U_i) = v_i)\ \text{by \cref{lem:markov},}\\
&= \prod^d_{i=1} \mathrm{d} K_i(v_i | v_{\pa(i)})\ \text{by definition of $\K$}.
\end{align*}   
Therefore, $P = \Obs{C}(\C_\M)$.

\emph{Property 2.} Let $\Phi \in \I(\V)$. The definition of the entailed graphical model along with the definition of an intervention on a structural causal model imply that $\M^\Phi$ entails $\C_{\M^\Phi} \in \CGM_{\V}$ with kernels $\K^\Phi$ given by $K^\Phi_i = \phi_i \# K_i$ for every $i \in [d]$. This means that $\C_{\M^\Phi} = \C^\Phi_\M$. Therefore,
\begin{align*}
    \Int{M}(\M,\Phi) &= \Obs{M}(\M^\Phi)\ \text{by definition,}\\
    &= \Obs{C}(\C_{\M^\Phi})\ \text{by the first property,}\\
    &= \Obs{C}(\C^\Phi_\M)\ \text{using the point made above,}\\
    &= \Int{C}(\C_\M, \Phi)\ \text{by definition.}
\end{align*}
This concludes the proof.
\end{proof}

\begin{proof}[\Cref{prop:scm_exists}]
\emph{Property n°1.} Let $\C := (\V,\G,\K) \in \CGM_{\V}$ have parental function $\pa$ and write $P := \Obs{C}(\C)$. Recall that for every $i \in [d]$ and every $v_{\pa(i)} \in \V_{\pa(i)}$, $G^P_{i | \pa(i)}(\bullet | v_{\pa(i)}) \# \Unif([0,1]) = K_i(\bullet | v_{\pa(i)})$. Then, $\M := (\V,\G,\U,P_0,f)$ with $\U_i = [0,1]$, $P_{0,i} = \operatorname{Unif}([0,1])$ and $f_i := G^P_{i | \pa(i)}$ is such that $\C_\M = \C$. Therefore, $\SCM_{\V,\C} \neq \emptyset$. We point out that the joint measurability of $G^P_{i | \pa(i)}(\bullet | \bullet)$, guaranteed by \citep[Theorem 2.1]{carlier2016vector}, is crucial for the SCM to be correctly defined. Note also that while we choose the conditional quantile function to provide a concrete example, there generally exist infinitely many functional representations of a same probability kernel $K_i$ \citep[Proposition 8.20]{kallenberg1997foundations}.

\emph{Property n°2.} This is a consequence of property n°1 and the existence of a causal graphical model fitting any observational distribution (\cref{prop:cgm_exists}).
\end{proof}

\begin{proof}[\Cref{prop:int_is_graph}]
We write $P := \Obs{M}(\M_1) = \Obs{M}(\M_2)$, and $\K^{(1)}, \K^{(2)}$ for the causal kernels of respectively $\C_{\M_1}$ and $\C_{\M_2}$. According to the first item of \cref{prop:scm_cgm}, $P = \Obs{C}(\C_{\M_1}) = \Obs{C}(\C_{\M_2})$. It then follows from \cref{prop:cgm_obs} applied with $\C_{\M_1}$ and $\C_{\M_2}$ that $P_{i | \pa(i)} = K^{(1)}_i$ and $P_{i | \pa(i)} = K^{(2)}_i$ for every $i \in [d]$. Therefore, $\K^{(1)} = \K^{(2)}$. Because the graph of $\C_{\M_1}$ and $\C_{\M_2}$ are the same by assumption, we have $\C_{\M_1} = \C_{\M_2}$. Finally, it follows from the second item of \cref{prop:scm_cgm} that $\M_1$ and $\M_2$ are interventionally equivalent.
\end{proof}

\begin{proof}[\Cref{prop:block_markov}]
Same as \cref{lem:markov}, but concatenating the solutions along the alternative worlds.
\end{proof}

\begin{proof}[\Cref{prop:counterfactual_law}]
We write $g^{\Phi_w}$ for the solution map of $\M^{\Phi_w}$ for every $w \in [W]$, as defined in the proof of \cref{prop:scm_obs}. It follows that every cross-world solution $(V^{\Phi_w})^W_{w=1}$ equals $(g^{\Phi_w}(U))^W_{w=1}$. This means that $\L\left( (V^{\Phi_w})^W_{w=1} \right) = (g^{\Phi_w})^W_{w=1} \# P_0$ for every cross-world solution. Let us write $Q_{\Phi_1,\ldots,\Phi_W} := \Ctf{M}(\M,(\Phi_w)^W_{w=1})$. Since, by definition, $Q_{\Phi_1,\ldots,\Phi_W} = \L\left( (V^{\Phi_w})^W_{w=1} \right)$, we have by projection that $Q_{\Phi_w} := \L(V^{\Phi_w}) := \Int{M}(\M,\Phi_w)$ for every $w \in [W]$. Therefore, $Q_{\Phi_1,\ldots,\Phi_W} \in \Joint{\{\Phi_w\}^W_{w=1}}{\V}{\left(\Int{M}(\M,\Phi_w)\right)^W_{w=1}}$.
\end{proof}

\begin{proof}[\Cref{prop:hierarchy}]
If $\Ctf{M}(\M_1,(\Phi_w)^W_{w=1}) = \Ctf{M}(\M_2,(\Phi_w)^W_{w=1})$, then, by marginalization, $\Int{M}(\M_1,\Phi_w) = \Int{M}(\M_2,\Phi_w)$ for each $w \in [W]$, according to \cref{prop:counterfactual_law}. This proves the first point. If $\Int{M}(\M_1,\Phi) = \Int{M}(\M_2,\Phi)$ for every $\Phi \in \I(\V)$, then, in particular, $\Int{M}(\M_1,\Phi_\emptyset) = \Int{M}(\M_2,\Phi_\emptyset)$, where $\Phi_\emptyset = \times^d_{i=1} \Id$. Additionally, note that $\Int{M}(\M,\Phi_\emptyset) = \Obs{M}(\M)$ for every $\M \in \SCM_{\V}$. This proves the second point.
\end{proof}

\begin{proof}[\Cref{prop:scm_vs_cgm}]
The introductory illustration from \cref{sec:intro_example} furnishes examples.
\end{proof}

\subsection{Proof of \cref{sec:ctf_models}}

\begin{proof}[\Cref{prop:int_margins}]
Let us write $Q := \Ctf{A}(\A, (\Phi_w)^W_{w=1})$. Then, for every $w \in [W]$,
\begin{align*}
    \mathrm{d}Q_{\Phi_w}(v^{(w)}) &= \prod^d_{i=1} \mathrm{d} \left( \phi_{w,i} \# S^{(i)}_{v^{(w)}_{\pa(i)}} \right) \left(v^{(w)}_i \right)\ \text{by projection on $w,$}\\
    &= \prod^d_{i=1} \mathrm{d} \left( \phi_{w,i} \# K_i\left(\bullet|v^{(w)}_{\pa(i)}\right) \right) \left(v^{(w)}_i \right)\ \text{by definition of $S^{(i)}$,}\\
    &= \prod^d_{i=1} \mathrm{d} \left( K^{\Phi_w}_i\left(\bullet|v^{(w)}_{\pa(i)}\right) \right) \left(v^{(w)}_i \right)\ \text{by definition of $\K^{\Phi_w}$}.
\end{align*}
Therefore, $Q_{\Phi_w} = \Int{C}(\C,\Phi_w) = \Int{A}(\A,\Phi_w)$ by definitions.
\end{proof}

\begin{proof}[\Cref{prop:entailed_a}]
Let $i \in [d]$ and $v_{\pa(i)} \in \V_{\pa(i)}$. Then,
\begin{align*}
    S^{(i)}_{v_{\pa(i)}} &= \law{f_i(v_{\pa(i)},U_i)}\ \text{by definition,}\\
    &= K_i(\bullet|v_{\pa(i)})\ \text{since $\M \in \SCM_{\V,\C}$}.
\end{align*}
This shows that $\A_\M \in \CR_{\V,\C}$. Next, let $(\Phi_w)^W_{w=1} \in \I(\V)^W$, where $W \geq 1$. Write $(V^{\Phi_w})^W_{w=1}$ for a cross-world solution of $\M$ relatively to $(\Phi_w)^W_{w=1}$. Then,
\begin{align*}
&\mathrm{d}\P \left( (V^{\Phi_w})^W_{w=1}=(v^{(w)})^W_{w=1} \right)\\ &= \prod^d_{i=1} \mathrm{d}\P \left( (V^{\Phi_w}_i)^W_{w=1}=(v^{(w)}_i)^W_{w=1} \mid (V^{\Phi_w}_{\pa(i)})^W_{w=1}=(v^{(w)}_{\pa(i)})^W_{w=1} \right)\ \text{by \cref{prop:block_markov},}\\
& = \prod^d_{i=1} \mathrm{d}\P \left( \left((\phi_{w,i} \circ f_i)(v^{(w)}_{\pa(i)}, U_i) \right)^W_{w=1}=(v^{(w)}_{\pa(i)})^W_{w=1} \right)\ \text{by \cref{prop:block_markov},}\\
&= \prod^d_{i=1} \mathrm{d}\P \left( \left( \underset{w \in [W]}{\times} \phi_{w,i} \right) \# S^{(i)}_{v^{(1)}_{\pa(i)},\ldots,v^{(W)}_{\pa(i)}} =(v^{(w)}_{\pa(i)})^W_{w=1} \right)\ \text{by definition of $S^{(i)}$}.
\end{align*}
The left-hand side corresponds to $\Ctf{M}(\M,(\Phi_w)^W_{w=1})$ and the right-hand side corresponds to $\Ctf{A}(\A_\M,(\Phi_w)^W_{w=1})$ by definitions. This concludes the proof.
\end{proof}

\begin{proof}[\Cref{thm:representation}]
Let $\C := (\V, \G, \K) \in \CGM_{\V}$ and $\A := (\C,\S) \in \CR_{\V,\C}$. Recall that $\S$ is a collection of one-step-ahead counterfactual distributions $(S^{(i)})^d_{i=1}$ adapted to $\C$. Since one-step-ahead counterfactual distributions are composable by definition, for every $i \in [d]$ there exists a \emph{measurable} stochastic process $Z_i = (Z_{i,v_{\pa(i)}})_{v_{\pa(i)} \in \V_{\pa(i)}}  : \Omega \to {\V_i}^{\V_{\pa(i)}}$ with law $S^{(i)}$. We define $\U := \times^d_{i=1} \U_i$, $P_0 := \otimes^d_{i=1} P_{0,i}$, and $f := (f_i)^d_{i=1}$ as follows. For every $i \in [d]$, $\U_i := \Omega$, $P_{0,i} := \P$ and $f_i : \V_{\pa(i)} \times \U_i \to \V_i, (v_{\pa(i)}, u_i) \mapsto Z_{i,v_{\pa(i)}}(u_i)$ for every $i \in [d]$. Crucially, the measurability of $f_i$ comes from the measurability of $Z_i$ By construction, $\M := (\V,\G,\U,P_0,f)$ is such that $\A_\M = \A$.
\end{proof}

\begin{proof}[\Cref{prop:concise}]
Define $\M_1$ as in the proof of \cref{prop:scm_exists}: letting $P := \Obs{C}(\C)$, $f^{(1)}_i := G^P_{i | \pa(i)}$ and $P^{(1)}_{0,i} := \operatorname{Unif}([0,1])$ for every $i \in [d]$. Then, define $\M_2$ via a change of exogenous variable. Concretely, let $f^{(2)}_i := G^P_{i | \pa(i)} \circ (\Id \times F^\Norm)$ and $P^{(2)}_{0,i} := \Norm$ for every $i \in [d]$. As such, $\M_1 \neq \M_2$.

Hereafter, we compare their entailed counterfactual models. To do so, its suffices to compare their entailed one-step-ahead counterfactual processes. For $i \in [d]$, $\M_1$ entails
\[
\left( G^P_{i | \pa(i)}(\bullet | v_{\pa(i)}) \right)_{v_{\pa(i)} \in \V_{\pa(i)}} \# \operatorname{Unif}([0,1])
\]
while $\M_2$ entails
\begin{multline*}
\left(  G^P_{i | \pa(i)}(\bullet | v_{\pa(i)}) \circ F^\Norm \right)_{v_{\pa(i)} \in \V_{\pa(i)}} \# \mathcal{N}(0,1) =\\ \left(  G^P_{i | \pa(i)}(v_{\pa(i)} | \bullet)\right)_{v_{\pa(i)} \in \V_{\pa(i)}} \# ( F^\Norm \# \mathcal{N}(0,1))
= \left(  G^P_{i | \pa(i)}(v_{\pa(i)} | \bullet)\right)_{v_{\pa(i)} \in \V_{\pa(i)}} \# \operatorname{Unif}([0,1]).
\end{multline*}
Consequently, $\A_{\M_1} = \A_{\M_2}$.
\end{proof}

\begin{proof}[\Cref{thm:ctf_eq}]
Let us start by the direct implication: if $\A_{\M_1} = \A_{\M_2}$ then for every $W \geq 1$ and every $(\Phi_w)^W_{w=1} \in \I(\V)^W$, $\Ctf{M}(\M_1,(\Phi_w)^W_{w=1}) = \Ctf{M}(\M_2,(\Phi_w)^W_{w=1})$, since $\Ctf{M}(\M_1,(\Phi_w)^W_{w=1}) = \Ctf{A}(\A_{\M_1},(\Phi_w)^W_{w=1})$ and $\Ctf{M}(\M_2,(\Phi_w)^W_{w=1}) = \Ctf{A}(\A_{\M_2},(\Phi_w)^W_{w=1})$ according to \cref{prop:entailed_a}. This means that $\M_1, \M_2$ are counterfactually equivalent.

Conversely, suppose that $\M_1, \M_2$ are counterfactually equivalent. We write $\M_1 := (\V,\G,\U^{(1)},P^{(1)}_0,f^{(1)})$ and $\M_2 := (\V,\G,\U^{(2)},P^{(2)}_0,f^{(2)})$, with solutions and $(V^{(1)},U^{(1)})$ and $(V^{(2)},U^{(2)})$. To show that $\A_{\M_1} = \A_{\M_2}$, it suffices to show that for each $i \in [d]$ the one-step-ahead counterfactual processes $Z^{(1)}_i := (f^{(1)}(v_{\pa(i)},U^{(1)}_i))_{v_{\pa(i)} \in \V_{\pa(i)}}$ and $Z^{(2)}_i := (f^{(2)}(v_{\pa(i)},U^{(2)}_i))_{v_{\pa(i)} \in \V_{\pa(i)}}$ have the same distribution. Let $i \in [d]$, $W \geq 1$, and $(v^{(w)}_{\pa(i)})^W_{w=1} \in \V_{\pa(i)}^W$ be fixed. Then, define $\Phi_w := \doint(\pa(i),v^{(w)}_{\pa(i)})$ for every $w \in [W]$. It corresponds to doing in $W$ parallel worlds a different atomic intervention on the parents of the variable $i$. Since $\M_1, \M_2$ are counterfactually equivalent, the associated corresponding solutions $(V^{(1),\Phi_w})^W_{w=1}$ and $(V^{(2),\Phi_w})^W_{w=1}$ have the same distribution. Therefore, $(V^{(1),\Phi_w}_i)^W_{w=1}$ and $(V^{(2),\Phi_w}_i)^W_{w=1}$ have the same distribution in particular. Moreover, $(V^{(1),\Phi_w}_i)^W_{w=1} = \left( f^{(1)}(v^{(w)}_{\pa(i)},U^{(1)}_i) \right)^W_{w=1} = (Z^{(1)}_{i,v^{(w)}_{\pa(i)}})^W_{w=1}$ and $(V^{(2),\Phi_w}_i)^W_{w=1} = \left( f^{(2)}(v^{(w)}_{\pa(i)},U^{(2)}_i) \right)^W_{w=1}  = (Z^{(2)}_{i,v^{(w)}_{\pa(i)}})^W_{w=1}$ by definition of $\left({\Phi_w}\right)^W_{w=1}$. This shows that the finite projections $(Z^{(1)}_{i,v^{(1)}_{\pa(i)}}, \ldots, Z^{(1)}_{i,v^{(W)}_{\pa(i)}})$ and $(Z^{(2)}_{i,v^{(1)}_{\pa(i)}}, \ldots, Z^{(1)}_{i,v^{(W)}_{\pa(i)}})$ are equal in law for any $W \geq 1$ and any $(v^{(w)}_{\pa(i)})^W_{w=1} \in \V_{\pa(i)}^W$. Consequently, $Z^{(1)}_i$ and $Z^{(2)}_i$ have the same distribution.
\end{proof}

\begin{proof}[\Cref{prop:unnormalize}]
For every $v_{\pa(i)} \in \V_{\pa(i)}$, the function $\psi^\C_i(\bullet|v_{\pa(i)})$ is defined as the non-decreasing transport map from $\Norm$ to $K_i(\bullet|v_{\pa(i)})$. Formally, $\psi^\C_i(\bullet|v_{\pa(i)}) = G_i(\bullet|v_{\pa(i)}) \circ F^\Norm$, where $G_i$ is the conditional quantile function associated to the probability kernel $K_i$. The measurability of $G_i(\bullet|v_{\pa(i)})$ follows from monotonicity, while the one of $G_i(e|\bullet)$ follows from the measurability of $K_i(\bullet|\bullet)$ in its second arguments \citep{fissler2022measurability,de2023conditional}. The joint measurability of $G_i$ requires further arguments. It can be seen as a special case of \citep[Theorem 2.1]{carlier2016vector}, which tackles the more general case of multivariate quantile functions. Then, $\psi^\C_i$ is measurable by composition.

Let us turn to $\Psi^C_i := \left( \underset{v_{\pa(i)} \in \V_{\pa(i)}}{\times} \psi^\C_i(\bullet|v_{\pa(i)}) \right)$. It is a (possibly uncountable) product of measurable functions, which is always measurable. This concludes the proof.
\end{proof}

\begin{proof}[\Cref{lem:norm_comp}]
Let $Q \in \Q^*(\X,\Y)$: there exists a \emph{measurable} random process $Y := (Y_x)_{x \in \X} \sim Q$. By push-forward, the random process $\left( \underset{x \in \X}{\times} \psi(\bullet|x) \right)\left( Y \right)$ represents the law $\left( \underset{x \in \X}{\times} \right) \# Q$. To conclude, we show that it is measurable.

Note that $\left( \underset{x \in \X}{\times} \psi(\bullet|x) \right)\left( Y \right) = \left( \psi(Y_x|x) \right)_{x \in \X}$. Since measurability is stable by composition, $(x,\omega) \mapsto \psi(Y_x(\omega)|x)$ is measurable by measurability of the process $Y$ and the function $\psi$. 
\end{proof}

\begin{proof}[\Cref{prop:norm}]
\emph{Item n°1.} First, we prove that $\Joint{\V_{\pa(i)}}{\V_i}{K_i(\bullet|\V_{\pa(i)})} = \Upsilon^{\C}_i\left( \Norma{\V_{\pa(i)}} \right)$. Note that there is no composability involved. The \say{$\supseteq$} sense directly follows from the definition of $\Psi^{\C}_i$. The \say{$\subseteq$} sense relies on copula arguments. It requires to show that, for every process measure $S^{(i)}$ with preassigned marginals, there exists a process measure $N^{(i)}$ with normal marginals such that $\Psi^{\C}_i \# N^{(i)} = S^{(i)}$. Let us write $P := \Obs{C}(\C)$, so that $K_i = P_{i \mid \pa(i)}$ for every $i \in [d]$. According to the infinite extension of Sklar's theorem \citep[Theorem 1]{benth2022copula}, there exists a process measure
\[
C^{(i)} \in \Joint{\V_{\pa(i)}}{[0,1]}{(\operatorname{Unif}([0,1]))_{v_{\pa(i)} \in \V_{\pa(i)}}}
\]
such that $\left( \underset{v_{\pa(i)} \in \V_{\pa(i)}}{\times} G^P_{i \mid \pa(i)}(\bullet|v_{\pa(i)}) \right) \# C^{(i)} = S^{(i)}$. For notational convenience we define $G^\times_i = \left( \underset{v_{\pa(i)} \in \V_{\pa(i)}}{\times} G^P_{i \mid \pa(i)}(\bullet|v_{\pa(i)}) \right)$ in the rest of the proof. We recall that $G^P_{i \mid \pa(i)}(\bullet|v_{\pa(i)})$ is the unique non-decreasing transport map from $\Unif{[0,1]}$ to $P_{i \mid \pa(i)}(\bullet | v_{\pa(i)})$ for every $v_{\pa(i)} \in \V_{\pa(i)}$. Since $F^{\Norm} \# \Norm = \operatorname{Unif}([0,1])$ and $F^{\Norm}$ is bijective such that ${F^{\Norm}}^{-1} = G^{\Norm}$, $N^{(i)} := \left( \underset{v_{\pa(i)} \in \V_{\pa(i)}}{\times} G^{\Norm} \right) \# C^{(i)}$ belongs to $\Norma{\V_{\pa(i)}}$. Moreover, this process measure meets $G^\times_i \circ \left( \underset{v_{\pa(i)} \in \V_{\pa(i)}}{\times} F^{\Norm} \right) \#  N^{(i)} = S^{(i)}$ by successive transports. To conclude, recall that $F^\Norm$ is increasing so that that by uniqueness of the non-decreasing transport map between two distributions $G^P_{i \mid \pa(i)}(\bullet|v_{\pa(i)}) \circ F^\Norm = \psi^{\C}_i(\bullet|v_{\pa(i)})$ for every $v_{\pa(i)} \in \V_{\pa(i)}$. Therefore, $G^\times_i \circ \left( \underset{v_{\pa(i)} \in \V_{\pa(i)}}{\times} F^{\Norm} \right) = \Psi^{\C}_i$, which means that $\Psi^{\C}_i \# N^{(i)} = S^{(i)}$.

To conclude this first part of the proof, note that the inclusion $\Jointc{\V_{\pa(i)}}{\V_i}{K_i(\bullet|\V_{\pa(i)})} \subseteq \Joint{\V_{\pa(i)}}{\V_i}{K_i(\bullet|\V_{\pa(i)})}$ is trivial. Moreover, the inclusion $\Upsilon^{\C}_i\left( \Normac{\V_{\pa(i)}} \right) \subseteq \Jointc{\V_{\pa(i)}}{\V_i}{K_i(\bullet|\V_{\pa(i)})}$ rests on \cref{lem:norm_comp}.

\emph{Item n°2}. If $\psi^{\C}_i(\bullet|v_{\pa(i)})$ is injective for every $v_{\pa(i)} \in \V_{\pa(i)}$, then $\Psi^{\C}_i$ is injective. On this basis, we first deal with non-necessarily composable process measures. Let $S^{(i)} \in \Joint{\V_{\pa(i)}}{\V_i}{K_i(\bullet|\V_{\pa(i)})}$, and let $N^{(i)},{N'}^{(i)} \in \Norma{\V_{\pa(i)}}$ such that $\Psi^{\C}_i \# N^{(i)} = \Psi^{\C}_i \# {N'}^{(i)} = S^{(i)}$. Applying the operator ${\Psi^{\C}_i}^{-1} \#$ both sides gives $N^{(i)} = {N'}^{(i)}$. Therefore, $\Upsilon^{\C}_i$ is injective from $\Norma{\V_{\pa(i)}}$ to $\Joint{\V_{\pa(i)}}{\V_i}{K_i(\bullet|\V_{\pa(i)})}$. Bijectivity follows from item n°1, which ensures surjectivity.

We turn to composable process measures. From item n°1, we know that for every $S^{(i)} \in \Jointc{\V_{\pa(i)}}{\V_i}{K_i(\bullet|\V_{\pa(i)})}$ there exists an $N^{(i)} \in \Norma{\V_{\pa(i)}}$ such that $\Psi^{\C}_i \# N^{(i)} = S^{(i)}$. Applying the operator ${\Psi^{\C}_i}^{-1} \#$ both sides gives $N^{(i)} = {\Psi^{\C}_i}^{-1} \# S^{(i)}$. The fact that ${\Psi^{\C}_i}^{-1} = \left( \underset{v_{\pa(i)} \in \V_{\pa(i)}}{\times} \left(\psi^{\C}_i\right)^{-1}(\bullet|v_{\pa(i)}) \right)$ and that $S^{(i)}$ is composable implies that $N^{(i)}$ is composable by \cref{lem:norm_comp}. It also entails uniqueness of the antecedent $N^{(i)}$. Therefore, $\Upsilon^{\C}_i$ is injective from $\Normac{\V_{\pa(i)}}$ to $\Jointc{\V_{\pa(i)}}{\V_i}{K_i(\bullet|\V_{\pa(i)})}$.
\end{proof}

\begin{proof}[\Cref{prop:scm_to_ctf}]
We recall that $S^{(i)} = \left( f_i(v_{\pa(i)},\bullet) \right)_{v_{\pa(i)} \in \V_{\pa(i)}} \# P_{0,i}$. Then (i) directly follows from the fact that $f_i(v_{\pa(i)},\bullet)$ is injective for every $v_{\pa(i)} \in \V_{\pa(i)}$.

Let us turn to (ii). If $f_i$ is \emph{non-decreasing} in its exogenous variable, then let $\tau_i$ be the unique \emph{non-decreasing} transport map from $\Norm$ to $P_{0,i}$. Otherwise, let $\tau_i$ be the unique \emph{non-increasing} transport map from $\mathcal{N}(0,1)$ to $P_{0,i}$. Whatever the situation, the function $f_i(v_{\pa(i)},\tau_i(\bullet))$ is non-decreasing for every $v_{\pa(i)} \in \V_{\pa(i)}$. Therefore, by uniqueness of the non-decreasing transport map from $\mathcal{N}(0,1)$ to $K_i(\bullet|v_{\pa(i)})$, $f_i(v_{\pa(i)},\tau_i(\bullet)) = \psi^\C_i(\bullet|v_{\pa(i)})$ for every $v_{\pa(i)} \in \V_{\pa(i)}$. Moreover, $S^{(i)} = \left( f_i(v_{\pa(i)},\tau_i(\bullet))\right)_{v_{\pa(i)} \in \V_{\pa(i)}} \# \Norm$ by successive transports. Then, by definition of $N^{\uparrow}$, this can be reframed as $S^{(i)} = \left( \underset{v_{\pa(i)} \in \V_{\pa(i)}}{\times} f_i(v_{\pa(i)},\tau_i(\bullet)) \right) \# N^{\uparrow} = \left( \underset{v_{\pa(i)} \in \V_{\pa(i)}}{\times} \psi^\C_i(\bullet|v_{\pa(i)}) \right) \# N^{\uparrow}$. This concludes the proof, as $S^{(i)} = \Psi^\C_i \# N^{\uparrow}$.
\end{proof}

\begin{proof}[\Cref{prop:not_anm}]
If $\sigma^2$ is constant, say $\sigma^2(x)=c^2$, then the ANM corresponding to the assignment $Y = m(X) + U_{\mathtt{y}}$, with $U_{\mathtt{y}} \sim \mathcal{N}(0,c^2)$, fits $\C$.

Let us suppose now that $\sigma^2$ is not constant. We reason \emph{ad absurdum}, by assuming that for $(X,Y) \sim P_{\mathtt{x,y}}$, there exist a measurable function $h_{\mathtt{y}} : \R \to \R$ and a random variable $U_{\mathtt{y}} \independent X$ such that $Y = h_{\mathtt{y}}(X) + U_{\mathtt{y}}$. Applying $\E\left[(\bullet - \E[\bullet | X=x])^2|X=x\right]$ (that is, the conditional variance) both sides of this equality gives $\sigma^2(x) = 0 + \E\left[(U_{\mathtt{y}} - \E[U_{\mathtt{y}}])^2\right]$. This contradicts the fact that $\sigma^2$ is not constant.
\end{proof}

\subsection{Proofs of \cref{sec:regression}}

\begin{proof}[\Cref{prop:posedness}]
First of all, let us show that, for every $\psi \in \digamma$
\begin{equation}\label{eq:characterization}
    \E\left[ D \left( \psi(\bullet | X) \# P_{\mathtt{e}}, P_{\mathtt{y} | \mathtt{x}}(\bullet | X) \right) \right] = 0 \iff \psi^*(\bullet|x) \# P_{\mathtt{e}} = P_{\mathtt{y} | \mathtt{x}}(\bullet | x) \text{ for $P_{\mathtt{x}}$-almost-every $x \in \R^d$.}
\end{equation}
By positivity of $\E$ and $D$, $\E\left[ D \left( \psi(\bullet | X) \# P_{\mathtt{e}}, P_{\mathtt{y} | \mathtt{x}}(\bullet | X) \right) \right] = 0 \iff D \left( \psi(\bullet | X) \# P_{\mathtt{e}}, P_{\mathtt{y} | \mathtt{x}}(\bullet | X) \right) = 0$ almost surely. According to the assumption on $D$, this happens just in case $\psi(\bullet | X) \# P_{\mathtt{e}} = P_{\mathtt{y} | \mathtt{x}}(\bullet | X)$ almost surely. By push-forward this is equivalent to $\psi(\bullet | x) \# P_{\mathtt{e}} = P_{\mathtt{y} | \mathtt{x}}(\bullet | x)$ for $P_{\mathtt{x}}$-almost-every $x \in \R^d$.

\emph{Property n°1.} According to \cref{eq:characterization}, $\E\left[ D \left( \tilde{\psi}(\bullet | X) \# P_{\mathtt{e}}, P_{\mathtt{y} | \mathtt{x}}(\bullet | X) \right) \right] = 0$. Moreover, by positivity of $D$ and $\E$, $\E\left[ D \left( \psi(\bullet | X) \# P_{\mathtt{e}}, P_{\mathtt{y} | \mathtt{x}}(\bullet | X) \right) \right] \geq 0$ for every $\psi \in \digamma$. Therefore,
\[
\min_{\psi \in \digamma} \E\left[ D \left( \psi(\bullet | X) \# P_{\mathtt{e}}, P_{\mathtt{y} | \mathtt{x}}(\bullet | X) \right) \right] = \E\left[D \left( \tilde{\psi}(\bullet | X) \# P_{\mathtt{e}}, P_{\mathtt{y} | \mathtt{x}}(\bullet | X) \right)\right] = 0.
\]
This means that any solution $\psi^*$ of \cref{eq:generic_learning} satisfies $E\left[D \left( \psi^*(\bullet | X) \# P_{\mathtt{e}}, P_{\mathtt{y} | \mathtt{x}}(\bullet | X) \right)\right] = 0$. Applying again \cref{eq:characterization} yields $\psi^*(\bullet|x) \# P_{\mathtt{e}} = P_{\mathtt{y} | \mathtt{x}}(\bullet | x)$ for $P_{\mathtt{x}}$-almost-every $x \in \R^d$.

\emph{Property n°2.} According to item n°1, any solution $\psi^*$ of \cref{eq:generic_learning} satisfies $\psi^*(\bullet|x) \# P_{\mathtt{e}} = P_{\mathtt{y} | \mathtt{x}}(\bullet | x)$ for $P_{\mathtt{x}}$-almost-every $x \in \R^d$. If additionally, $\psi^*(\bullet|x)$ is non-decreasing for every $x \in \R^d$, then $\psi^*(\bullet|x)$ is the unique non-decreasing transport map from $P_{\mathtt{e}}$ to $P_{\mathtt{y} | \mathtt{x}}(\bullet | x)$ for $P_{\mathtt{x}}$-almost-every $x \in \R^d$. This concludes the proof.
\end{proof}

\subsection{Proofs of \cref{sec:comparison}}

\begin{proof}[\Cref{prop:deterministic_process}]
We prove each proposition in order. The first one serves to prove the second one.

\emph{Proposition n°1.} We start by the \say{if} implication. The equality $\gamma = (\Id,T) \# P$ means that $\gamma$ is deterministic from left to right while the equality $\gamma = (T^{-1},\Id) \# P'$ means that $\gamma$ is deterministic from right to left. Therefore, $\gamma$ is deterministic by definition.

We turn to the \say{only if} implication. By assumption, there exist $T : \Y \to \Y'$ and $T : \Y' \to \Y$ such that $\gamma = (\Id,T) \# P = (T',\Id) \# P'$. Before all, write $H := \{(y,y') \in \Y \times \Y' \mid T(y)=y'\}$ and notice that $(\Id,T)^{-1}(H) = \{y \in \Y \mid T(y)=T(y)\} = \Y$. Consequently, $\gamma(H) = \left( (\Id,T) \# P \right) (H) = P\left( (\Id,T)^{-1}(H) \right) = P(\Y) = 1$. Similarly, we can show that $\gamma(H')=1$ where $H' := \{(y,y') \in \Y \times \Y' \mid y=T'(y'))\}$. To sum-up, $\gamma(H)=\gamma(H')=1$. Then, we show that $P(T' \circ T = \Id) = 1$ and $P'(T \circ T' = \Id) = 1$. Firstly, $P(\{y \in \Y \mid (T' \circ T)(y)=y \}) = \gamma(\{(y,y') \in \Y \times \Y' \mid (T' \circ T)(y)=y \})$. Then, $\gamma(\{(y,y') \in \Y \times \Y' \mid (T' \circ T)(y)=y \}) = \gamma(\{(y,y') \in \Y \times \Y' \mid (T' \circ T)(y)=y \} \cap H)$ since $\gamma(H)=1$. Moreover, by the definitions of $H$ and $H'$, $\gamma(\{(y,y') \in \Y \times \Y' \mid (T' \circ T)(y)=y \} \cap H) = \gamma(\{(y,y') \in \Y \times \Y' \mid T'(y')=y \} \cap H) = \gamma(H' \cap H) = 1$. Consequently, $P(\{y \in \Y \mid (T' \circ T)(y)=y \}) = 1$, which means that $T = {T'}^{-1}$ $P$-almost everywhere. Following the same strategy, one can show that $P'(\{ y' \in \Y' \mid (T \circ T')(y')=y'\}) = 1$. This finally entails that $\gamma = (T',\Id) \# P' = (T^{-1},\Id) \# P'$ and $\gamma = (\Id,T) \# P = (\Id, {T'}^{-1}) \# P$.

\emph{Proposition n°2.} We start by the \say{if} implication. By assumption, there exists a collection of measurable maps $(s_{x,x'})_{x,x' \in \X}$ such that $S_{x,x'} = (\Id, s_{x,x'}) \# S_x$ for every $x,x' \in \X$. Before all, we show properties of $(s_{x,x'})_{x,x' \in \X}$. Let $x,x' \in \X$ and notice that the coupling $S_{x',x}$ has two expressions: $S_{x',x} = (\Id, s_{x',x}) \# S_{x'}$ by assumption and $S_{x',x} = (s_{x,x'}, \Id) \# S_x$ by permutation of $S_{x,x'}$. It then follows from the above proposition that $s_{x,x'}$ is invertible so that $s^{-1}_{x,x'} = s_{x',x}$ (up to sets of measure zero). Next, we fix a $z \in \X$ and define $P_0 := S_z$ and $b_x := s_{z,x}$ for every $x \in \X$. The rest of the proof amounts to showing that $(b_x)_{x \in \X} \# P_0 = S$. Since $S \in \Q(\X,\Y)$, there exists a random process $Y : \Omega \to \Y^\X$ such that $Y \# \P = S$ \citep[Theorem 8.23]{kallenberg1997foundations}. We point out that there is no guarantee that $Y_x$ is injective for every $x \in \X$. Let $x \in \X$ and observe that $S_{z,x} = (\Id, b_x) \# P_0 = (Y_z, Y_x) \# \P$. Consequently, according to the proof of proposition n°1, $Y_x = b_x \circ Y_z$ ($\P$-almost surely) for every $x \in \X$. To conclude, $S = (Y_x)_{x \in \X} \# \P = (b_x \circ Y_z)_{x \in \X} \# \P = (b_x)_{x \in \X} \# (Y_z \# \P) = (b_x)_{x \in \X} \# P_0$.

We turn to the \say{only if} implication. By assumption, there exists a collection of injective measurable maps $(b_x)_{x \in \X}$ and a probability $P_0$ such that $(b_x)_{x \in \X} \# P_0 = S$. Let $x,x' \in \X$ and consider the two following implications. First, $b_x \# P_0 = S_x$ and $P_0 = b^{-1}_x \# S_x$ by invertibility. Second, by projection, $S_{x,x'} = (b_x,b_{x'}) \# P_0$. Combining the two leads to $S_{x,x'} = (\Id,b_{x'} \circ b^{-1}_x) \# P_0$, which means that $S_{x,x'}$ is deterministic.
\end{proof}

\begin{proof}[\Cref{prop:composition}]
Let us start with item n°1. Note that (i) implies that $s_{x, x'}$ is invertible such that $s^{-1}_{x, x'} = s_{x', x}$ for every $x,x' \in \X$. Fix an arbitrary $z \in \X$ and define $b_x := s_{z, x}$ for every $x \in \X$. This enables us to write $s_{x, x'} = b_{x'} \circ b^{-1}_x$ and $(\Id, s_{x, x'}) \# P^{(x)} := (b_x, b_{x'}) \# P^{(z)}$ for every $x,x' \in \X$. By (ii), the marginals of $(\Id, s_{x, x'}) \# P^{(x)}$ are $P^{(x)}$ and $P^{(x')}$. To conclude, define the process measure $S := {(b_x)_{x \in \X}} \# P^{(z)}$: by construction it fits the marginals $(P^{(x)})_{x \in \X}$ and is such that $S_{x,x'} = (\Id, s_{x, x'}) \# P^{(x)}$. According to \cref{prop:deterministic_process}, $S$ is deterministic.

We turn to proving item n°2. By assumption, $S$ can be expressed as $S = (b_x)_{x \in \X} \# P_0$ where $b_x$ is injective for every $x \in \X$. Then, define $s_{x, x'} := b_{x'} \circ b^{-1}_x$ for every $x,x' \in \X$. Note that (a) holds by construction, while (b) is a direct consequence of ${b_{x'}} \# P_0 = P^{(x')}$ and ${b^{-1}_x} \# P^{(x)} = P_0$ for every $x,x' \in \X$. Moreover, by definition, $S_{x, x'} = (b_x \times b_{x'}) \# P_0$, and, by assumption, $P_0 = b^{-1}_x \circ P^{(x)}$, for every $x,x' \in \X$. Therefore, $S_{x, x'} = (\Id, b_{x'} \circ b^{-1}_x) \# P^{(x)} = (\Id, s_{x,x'}) \# P^{(x)}$, which means that (c) holds.
\end{proof}

\section{Constrained monotonic neural networks}\label{sec:monotonic_networks}

In this section, we summarize the key principles of \cite{runje2023constrained} to design partially monotonic neural networks with light requirements and universal approximation guarantees of continuous monotonic functions. They combine two ingredients: dense monotonic linear layers and special activation function. 

We start with the linear layers. Let $d_1,d_2 \geq 1$ be integers. For any $M \in \R^{d_2 \times d_1}$ and any $\iota \in \{-1,0,1\}^{d_1}$, define $\abs{M}_\iota$ as
\[
    (\abs{M}_\iota)_{i,j} := \begin{cases}
        \abs{M_{i,j}} & \text{if } \iota_j = 1\\
        -\abs{M_{i,j}} & \text{if } \iota_j = -1\\
        M_{i,j} & \text{otherwise}.
    \end{cases}
\]
Then, let $M \in \R^{d_2 \times d_1}$, $o \in \R^{d_2}$, and $\iota \in \{-1,0,1\}^{d_1}$. the \emph{constrained linear layer} with monotonicity indicator $\iota$ and weights $(M,o)$ is given by $z \mapsto \abs{M}_\iota \cdot z + o$. Each output component is forced to be partially non-decreasing (respectively, non-increasing) in every input $j$ such that $\iota_j = 1$ (respectively, $\iota_j=-1$) layer.

We turn the the activation functions. Classical activation function (ReLU, ELU, sigmoid,\ldots) are non-decreasing. Thereby, composing a monotonic linear layer by a classical activation function preserves the monotonicity. It is actually the oldest and simplest way to achieve monotonicity in neural networks \citep{archer1993application}. However, constructing feed-forward neural networks as such generally fails to well approximate any monotonic function. Because classical activation functions are \emph{convex non-decreasing}, the obtained neural networks cannot fit nonconvex functions. To remedy to this limitation, the authors proposed alternative versions of activation functions. For any zero-centered, non-decreasing, convex, and lower bounded activation function $\breve{\rho} : \R^{d_2} \to \R^{d_2}$, define for every $z \in \R^{d_2}$
\begin{align*}
    &\widehat{\rho}(z) := - \breve{\rho}(-z),
    &\tilde{\rho}(z) := \begin{cases}
        \breve{\rho}(z+1)-\breve{\rho}(1) & \text{if } z<0\\
        \breve{\rho}(z-1)+\breve{\rho}(1) & \text{otherwise.}\\
    \end{cases}
\end{align*}
Note that $\widehat{\rho}$ and $\tilde{\rho}$ are, respectively, concave upper-bounded and bounded counterparts of $\breve{\rho}$. Crucially, the three versions are all non-decreasing. Then, for a split parameter $\eta := (\breve{\eta}, \widehat{\eta}, \tilde{\eta}) \in \N^3$ such that $\breve{\eta} + \widehat{\eta} + \tilde{\eta} = d_2$, the combined activation function $\rho^\eta$ is defined for every $z \in \R^{d_2}$ and every $i \in [d_2]$ as
\[
\rho^\eta(z)_i := \begin{cases}
    \breve{\rho}(z_i) & \text{if } i \leq \breve{\eta}\\
    \widehat{\rho}(z_i) & \text{if } \breve{\eta} < i \leq \breve{\eta} + \widehat{\eta}\\
    \tilde{\rho}(z_i) & \text{otherwise}.
\end{cases}
\]

One can build a neural network with partial monotonicity constraints with the following recipe. Consider only monotonic constrained linear layers as dense layers: encode the desired monotonicity in the indicator $\iota$ of the \emph{input} layer and choose indicators filled with only ones in the successive layers. Take $\rho^\eta$ as activation functions (for some $\rho$ and $\eta$) in every layers. We follow this strategy in practice. Interestingly, this approach does not add any trainable parameter to the network and does not modify the optimization problem.

\section{Infinite quantile regression}\label{sec:quantile_regression}

As mentioned in \cref{sec:regression}, the distributional regression problem \cref{eq:generic_learning} can solve infinite quantile regression: it suffices to take $P_{\mathtt{e}} := \Unif{[0,1]}$. This section includes a short simulation to illustrate this point.

\begin{figure}[tb]
    \centering
    \includegraphics[width=0.94\linewidth]{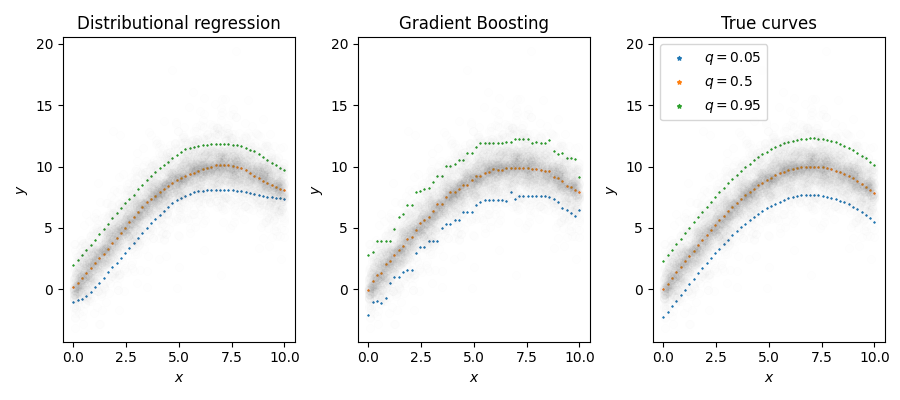}
    \caption{Quantile curves from three regression models.}
    \label{fig:toy_quantiles}
\end{figure}

We consider the same dataset as in the toy example from \cref{sec:exp}. On a given training sample of size 5000, we train a distributional-regression model with $L=4$ and $R=64$, during 40 epochs with a batch size of 64 and a learning rate of $10^{-4}$. Additionally, we conduct quantile regression with gradient boosting models for the quantiles $q \in \{0.05, 0.5, 0.95\}$. Then, on a test sample of size 5000, we evaluate the trained models with the mean-pinball loss of degree $q$ for every $q$ and compare them against the true conditional quantiles. For evaluation, we repeat the training on 10 independent data sets. \Cref{tab:pinball} reports results. \Cref{fig:toy_quantiles} represents the obtained regression curve at a single repetition.

\begin{table}[]
    \centering
    \begin{tabular}{|c||l|l|l|}
    \hline
      & 0.05 & 0.5 & 0.95\\
        \hline
     DR & $0.3890 \pm 0.0007$ & $3.040 \pm 0.0050$ & $6.3152 \pm 0.0148$ \\
     GB & $0.1679 \pm 0.0004$ & $0.5067 0.0050$ & $0.1687 \pm 0.0005$ \\
     Truth & $0.1664$ & $0.5022$ & $0.1646$ \\
     \hline
    \end{tabular}
    \caption{Mean values and standard deviations of the mean pinball loss for every model and every quantile, computed over 10 replications of the training process.}
    \label{tab:pinball}
\end{table}

Our model is the least accurate, with an error that increases with the quantile. This is consistent with the observation from \cref{sec:exp}: the distributional regression model produces outliers at high $\tty$ values. We leave the analysis of this phenomenon for future research. Additionally, note that training a distributional regression model, which amounts to learning a conditional generative model, is more computationally involved than training standard quantile regression models. Therefore, at this stage, we recommend practitioners to employ standard regression models when only a restricted number of conditional moments (the conditional expectation, conditional quantiles) are of interest. Distributional regression becomes convenient when one needs to make a potentially large number of queries on $P_{\tty|\ttx}$.

\bibliography{references}

\end{document}